\documentclass[runningheads, dvipsnames]{llncs}
\usepackage{graphicx}

\usepackage{tikz}
\usepackage{comment}
\usepackage{amsmath,amssymb} 

\usepackage{moresize}
\usepackage{booktabs}
\usepackage[utf8]{inputenc} 
\usepackage[T1]{fontenc}    
\usepackage{url}            
\usepackage{amsfonts}       
\usepackage{nicefrac}       
\usepackage{bm}
\usepackage{enumitem}
\usepackage{multirow}
\usepackage{multicol}
\usepackage{wrapfig}
\usepackage{float}
\usepackage{subcaption}
\usepackage{caption}
\usepackage{enumitem}
\usepackage{xcolor}
\usepackage{cite}
\usepackage{ragged2e}

\newcommand{\Maml}{\textsc{Maml}}
\newcommand{\Proto}{\textsc{ProtoNet}}
\newcommand{\MC}{\textsc{MC}}
\newcommand{\IMP}{\textsc{IMP}}

\usepackage{hyperref}
\usepackage[capitalise]{cleveref}

\usepackage{amsmath,amsfonts,bm}

\def\eqref#1{equation~\ref{#1}}

\def\1{\bm{1}}

\newcommand{\test}{\mathcal{D_{\mathrm{test}}}}

\def\rvc{{\mathbf{c}}}

\def\rvq{{\mathbf{q}}}

\def\rvs{{\mathbf{s}}}

\def\rvw{{\mathbf{w}}}
\def\rvx{{\mathbf{x}}}

\def\rmW{{\mathbf{W}}}
\def\rmX{{\mathbf{X}}}

\DeclareMathAlphabet{\mathsfit}{\encodingdefault}{\sfdefault}{m}{sl}
\SetMathAlphabet{\mathsfit}{bold}{\encodingdefault}{\sfdefault}{bx}{n}

\def\gC{{\mathcal{C}}}

\def\gL{{\mathcal{L}}}

\def\gN{{\mathcal{N}}}

\def\sR{{\mathbb{R}}}

\def\sV{{\mathbb{V}}}

\def\sX{{\mathbb{X}}}
\def\sY{{\mathbb{Y}}}

\newcommand{\E}{\mathbb{E}}

\newcommand{\softmax}{\mathrm{softmax}}

\DeclareMathOperator*{\argmin}{arg\,min}

\newcommand{\Lcal}[0]{\mathcal{L}}
\newcommand{\Scal}[0]{\mathcal{S}}
\newcommand{\Mcal}[0]{\mathcal{M}}

\newcommand{\ie}[0]{\emph{i.e.~}}

\usepackage[accsupp]{axessibility}  

\begin{document}
\pagestyle{headings}
\mainmatter
\def\ECCVSubNumber{6144}  

\title{Improving Few-Shot Learning through Multi-task Representation Learning Theory}


\titlerunning{Improving FSL through MTR Theory}
%

\author{Quentin Bouniot\inst{1,2}\orcidID{0000-0002-0982-372X}, Ievgen Redko\inst{2}\orcidID{0000-0002-3860-5502}, Romaric Audigier\inst{1}\orcidID{0000-0002-4757-2052}, Angélique Loesch\inst{1}\orcidID{0000-0001-5427-3010}, Amaury Habrard\inst{2,3}\orcidID{0000-0003-3038-9347}}
\authorrunning{Q. Bouniot et al.}
%
\institute{Université Paris-Saclay, CEA, List, F-91120, Palaiseau, France \\
\email{\{firstname.lastname\}@cea.fr}
\and
Université de Lyon, UJM-Saint-Etienne, CNRS, IOGS,  \\
Laboratoire Hubert Curien UMR 5516, F-42023, Saint-Etienne, France \\
\email{\{firstname.lastname\}@univ-st-etienne.fr} \and
Institut Universitaire de France (IUF)
}
\maketitle

\linepenalty=1000

\allowdisplaybreaks

\begin{abstract}
In this paper, we consider the framework of multi-task representation (MTR) learning where the goal is to use source tasks to learn a representation that reduces the sample complexity of solving a target task. We start by reviewing recent advances in MTR theory and show that they can provide novel insights for popular meta-learning algorithms when analyzed within this framework. In particular, we highlight a fundamental difference between gradient-based and metric-based algorithms in practice and put forward a theoretical analysis to explain it. Finally, we use the derived insights to improve the performance of meta-learning methods via a new spectral-based regularization term and confirm its efficiency through experimental studies on few-shot classification benchmarks. To the best of our knowledge, this is the first contribution that puts the most recent learning bounds of MTR theory into practice for the task of few-shot classification.

\end{abstract}

\section{Introduction}

Even though many machine learning methods now enjoy a solid theoretical justification, some more recent advances in the field are still in their preliminary state which requires the hypotheses put forward by the theoretical studies to be implemented and verified in practice. One such notable example is the success of \textit{meta-learning}, also called \textit{learning to learn} (LTL), methods where the goal is to produce a model on data coming from a set of (meta-train) source tasks to use it as a starting point for learning successfully a new previously unseen (meta-test) target task. The success of many meta-learning approaches is directly related to their capacity of learning a good representation \cite{Raghu2020Rapid} from a set of tasks making it closely related to multi-task representation learning (MTR).
For this latter, several theoretical studies \cite{baxterModelInductiveBias2000,pentinaPACBayesianBoundLifelong,maurerBenefitMultitaskRepresentation2016,AmitM18,Yin2020Meta-Learning}
provided probabilistic learning bounds that require the amount of data in the meta-train source task \textit{and} the number of meta-train tasks to tend to infinity for it to be efficient. While capturing the underlying general intuition, these bounds do not suggest that all the source data is useful in such learning setup due to the additive relationship between the two terms mentioned above and thus, for instance, cannot explain the empirical success of MTR in few-shot classification (FSC) task. To tackle this drawback, two very recent studies \cite{duFewShotLearningLearning2020,tripuraneniProvableMetaLearningLinear2020} aimed at finding deterministic assumptions that lead to faster learning rates allowing MTR algorithms to benefit from all the source data. Contrary to probabilistic bounds that have been used to derive novel learning strategies for meta-learning algorithms \cite{AmitM18,Yin2020Meta-Learning}, there has been no attempt to verify the validity of the assumptions leading to the fastest known learning rates in practice or to enforce them through an appropriate optimization procedure.

In this paper, we aim to use the recent advances in MTR theory \cite{tripuraneniProvableMetaLearningLinear2020,duFewShotLearningLearning2020} to explore the inner workings of these popular meta-learning methods. Our rationale for such an approach stems from a recent work \cite{pmlr-v139-wang21ad} proving that the optimization problem behind the majority of meta-learning algorithms can be written as an MTR problem.
Thus, we believe that looking at meta-learning algorithms through the recent MTR theory lens, could provide us a better understanding for the capacity to work well in the few-shot regime.
In particular, we take a closer look at two families of meta-learning algorithms, notably: gradient-based algorithms \cite{raviOPTIMIZATIONMODELFEWSHOT2017,nicholFirstOrderMetaLearningAlgorithms2018,liMetaSGDLearningLearn2017, NEURIPS2018_66808e32,leeMetaLearningDifferentiableConvex2019,bertinettoMetalearningDifferentiableClosedform2019,ParkO19, NEURIPS2020_8989e07f, Raghu2020Rapid} including \Maml\ \cite{finnModelAgnosticMetaLearningFast2017} and metric-based algorithms \cite{kochSiameseNeuralNetworks,vinyalsMatchingNetworksOne2016,snellPrototypicalNetworksFewshot2017,sungLearningCompareRelation2018,pmlr-v97-allen19b,liFindingTaskRelevantFeatures2019, simonAdaptiveSubspacesFewShot2020} with its most prominent example given by \Proto\ \cite{snellPrototypicalNetworksFewshot2017}. \\
Our main contributions are then two-fold: 
\begin{enumerate}[topsep=0pt, partopsep=0pt]
    \item We empirically show that tracking the validity of assumptions on optimal predictors used in \cite{tripuraneniProvableMetaLearningLinear2020,duFewShotLearningLearning2020} reveals a striking difference between the behavior of gradient-based and metric-based methods in how they learn their optimal feature representations. We provide elements of theoretical analysis that explain this behavior and explain the implications of it in practice. Our work is thus complementary to Wang et al. \cite{pmlr-v139-wang21ad} and connects MTR, FSC and Meta-Learning from both theoretical and empirical points of view.
    \item Following the most recent advances in the MTR field that leads to faster learning rates, we show that theoretical assumptions mentioned above can be forced through simple yet effective learning constraints which improve performance of the considered algorithms for FSC baselines: gradient- and metric-based methods using episodic training, as well as non-episodic algorithm such as Multi-Task Learning (MTL \cite{pmlr-v139-wang21ad}).
\end{enumerate}
The rest of the paper is organized as follows. We introduce the MTR problem and the considered meta-learning algorithms in \cref{sec:prelim}. In \cref{sec:understand}, we investigate and explain how they behave in practice. We further show that one can force meta-learning algorithms to satisfy such assumptions through adding an appropriate spectral regularization term to their objective function.  
In \cref{sec:expe}, we provide an experimental evaluation of several state-of-the-art meta-learning and MTR methods. 
We conclude and outline the future research perspectives in \cref{sec:conclusion}.

\section{Preliminary Knowledge}
\label{sec:prelim}

\subsection{Multi-Task Representation Learning Setup}
Given a set of $T$ source tasks observed through finite size samples of size $n_1$ grouped into matrices $\rmX_t = (\rvx_{t,1}, \dots, \rvx_{t,n_1}) \in \sR^{n_1\times d}$ and vectors of outputs $y_t = (y_{t,1}, \dots, y_{t,n_1}) \in \sR^{n_1},\ \forall t \in [[T]]:= \{1, \dots, T\}$ generated by their respective distributions $\mu_{t}$, the goal of MTR is to learn a shared representation $\phi$ belonging to a certain class of functions $\Phi := \{\phi\ |\ \phi: \sX \rightarrow \sV,\ \sX \subseteq \sR^d, \ \sV \subseteq \sR^k\}$, generally represented as (deep) neural networks, and linear predictors $\rvw_t \in \sR^{k},\ \forall t \in [[T]]$ grouped in a matrix $\rmW \in \sR^{T\times k}$. More formally, this is done by solving the following optimization problem:
\begin{equation}
    \hat{\phi}, \widehat{\rmW} \!=\! \argmin_{\phi \in \Phi, \rmW \in \sR^{T\times k}} \frac{1}{Tn_1}\sum_{t=1}^T\sum_{i=1}^{n_1} \ell(y_{t,i},\langle \rvw_t, \phi(\rvx_{t,i})\rangle),
\end{equation}
where $\ell:\sY \times \sY \rightarrow \sR_+$, with $\sY \subseteq \sR$, is a loss function. Once such a representation is learned, we want to apply it to a new previously unseen target task observed through a pair $(\rmX_{T+1} \in \sR^{n_2\times d}, y_{T+1} \in \sR^{n_2})$ containing $n_2$ samples generated by the distribution $\mu_{T+1}$. We expect that a linear classifier $\rvw$ learned on top of the obtained representation leads to a low true risk over the whole distribution $\mu_{T+1}$. For this, we first use $\hat{\phi}$ to solve the following problem:
\begin{equation}
    \hat{\rvw}_{T+1} = \argmin_{\rvw \in \sR^{k}} \frac{1}{n_2}\sum_{i=1}^{n_2} \ell(y_{T+1,i},\langle \rvw, \hat{\phi}(\rvx_{T+1,i})\rangle).
\end{equation}
Then, we define the true target risk of the learned linear classifier $\hat{\rvw}_{T+1}$ as:
$\gL(\hat{\phi}, \hat{\rvw}_{T+1}) = \mathop{\E}_{(\rvx,y)\sim \mu_{T+1}}[\ell(y,\langle \hat{\rvw}_{T+1}, \hat{\phi}(\rvx)\rangle)]$
and want it to be as close as possible to the ideal true risk $\gL(\phi^*, \rvw_{T+1}^*)$ where $\rvw_{T+1}^*$ and $\phi^*$ satisfy:
\begin{equation}
        \forall t \in [[T+1]] \text{  and  } (\rvx,y) \sim \mu_t, \quad y = \langle \rvw_t^*, \phi^*(\rvx)\rangle + \varepsilon, \quad \varepsilon \sim \gN(0,\sigma^2).
    \label{eq:data_gen_model}
\end{equation}

\noindent Equivalently, most of the works found in the literature seek to upper-bound the \textit{excess risk} defined as $\text{ER}(\hat{\phi}, \hat{\rvw}_{T+1}) := \gL(\hat{\phi}, \hat{\rvw}_{T+1}) - \gL(\phi^*, \rvw_{T+1}^*)$. 

\subsection{Learning Bounds and Assumptions}
First studies in the context of MTR relied on the probabilistic assumption \cite{baxterModelInductiveBias2000,pentinaPACBayesianBoundLifelong,maurerBenefitMultitaskRepresentation2016,AmitM18,Yin2020Meta-Learning} stating that meta-train and meta-test tasks distributions are all sampled i.i.d. from the same random distribution. 
A natural improvement to this bound was then proposed by \cite{duFewShotLearningLearning2020} and \cite{tripuraneniProvableMetaLearningLinear2020} that obtained the bounds on the excess risk behaving as 
\begin{equation}
    \text{ER}(\hat{\phi}, \hat{\rvw}_{T+1}) \leq O\left(\frac{C(\Phi)}{n_1T}+\frac{k}{n_2}\right),
\end{equation}
where $C(\Phi)$ is a measure of the complexity of $\Phi$.
Both these results show that all the source and target samples are useful in minimizing the excess risk. Thus, in the FSC regime where target data is scarce, all source data helps to learn well. From a set of assumptions made by the authors in both of these works, we note the following two:

\textbf{Assumption 1: Diversity of the source tasks}\phantom{.} The matrix of optimal predictors $\rmW^*$ should cover all the directions in $\sR^k$ evenly. More formally, this can be stated as
        $\kappa(\rmW^*) := \frac{\sigma_1(\rmW^*)}{\sigma_k(\rmW^*)} = O(1),$

where $\sigma_i(\cdot)$ denotes the $i\textsuperscript{th}$ singular value of $\rmW^*$. As pointed out by the authors, such an assumption can be seen as a measure of diversity between the source tasks that are expected to be complementary to each other to provide a useful representation for a previously unseen target task. In the following, we will refer to $\kappa(\rmW)$ as the \emph{condition number} for matrix $\rmW$. 

\textbf{Assumption 2: Consistency of the classification margin}\ The norm of the optimal predictors $\rvw^*$ should not increase with the number of tasks seen during meta-training\footnote{While not stated separately, this assumption is used in \cite{duFewShotLearningLearning2020} to derive the final result on p.5 after the discussion of Assumption 4.3.}. This assumption says that the classification margin of linear predictors should remain constant thus avoiding over- or under-specialization to the seen tasks. 

An intuition behind these assumptions and a detailed review can be found in the Appendix. While being highly insightful, the authors did not provide any experimental evidence suggesting that verifying these assumptions in practice helps to learn more efficiently in the considered learning setting. 

\subsection{Meta-Learning Algorithms}
Meta-learning algorithms considered below learn an optimal representation sequentially via the so-called episodic training strategy introduced by~\cite{vinyalsMatchingNetworksOne2016}, instead of jointly minimizing the training error on a set of source tasks as done in MTR. Episodic training mimics the training process at the task scale with each task data being decomposed into a training set \emph{(support set $S$)} and a testing set \emph{(query set $Q$)}. 
Recently, \cite{chenCloserLookTraining2020} showed that the episodic training setup used in meta-learning leads to a generalization bounds of $O(\frac{1}{\sqrt{T}})$. This bound is independent of the task sample size $n_1$, which could explain the success of this training strategy for FSC in the asymptotic limit. 
However, unlike the results obtained by \cite{duFewShotLearningLearning2020} studied in this paper, the lack of dependence on $n_1$ makes such a result uninsightful in practice as we are in a finite-sample size setting. This bound does not give information on other parameters to leverage when the task number cannot increase. 
We now present two major families of meta-learning approaches below. 

\noindent \textbf{Metric-based methods}\phantom{.} These methods learn an embedding space in which feature vectors can be compared using a similarity function (usually a $L_2$ distance or cosine similarity) \cite{kochSiameseNeuralNetworks,vinyalsMatchingNetworksOne2016,snellPrototypicalNetworksFewshot2017,sungLearningCompareRelation2018,pmlr-v97-allen19b,liFindingTaskRelevantFeatures2019, simonAdaptiveSubspacesFewShot2020}. They typically use a form of contrastive loss as their objective function, similarly to Neighborhood Component Analysis (NCA)~\cite{NIPS2004_NCA_goldberger} or Triplet Loss~\cite{hofferDeepMetricLearning2015}. In this paper, we focus our analysis on the popular Prototypical Networks~\cite{snellPrototypicalNetworksFewshot2017} (\Proto) that computes prototypes as the mean vector of support points belonging to the same class: 
$\rvc_i = \frac{1}{|S_i|} \sum_{\rvs \in S_i} \phi(\rvs),$
with $S_i$ the subset of support points belonging to class $i$.
\Proto\ minimizes the negative log-probability of the true class $i$ computed as the softmax over distances to prototypes $\rvc_i$: 
\begin{equation}
    \Lcal_{proto}(S,Q,\phi):= \E_{\rvq \sim Q} \left[ - \log \frac{\exp(-d(\phi(\rvq), \rvc_i))}{\sum_{j} \exp{(- d(\phi(\rvq), \rvc_j))}} \right]
\end{equation}
with $d$ being a distance function used to measure similarity between points in the embedding space.

\noindent \textbf{Gradient-based methods}\phantom{.} These methods learn through end-to-end or two-step optimization  \cite{raviOPTIMIZATIONMODELFEWSHOT2017,nicholFirstOrderMetaLearningAlgorithms2018,liMetaSGDLearningLearn2017, NEURIPS2018_66808e32,leeMetaLearningDifferentiableConvex2019,bertinettoMetalearningDifferentiableClosedform2019,ParkO19, NEURIPS2020_8989e07f, Raghu2020Rapid} where given a new task, the goal is to learn a model from the task's training data specifically adapted for this task.
\Maml~\cite{finnModelAgnosticMetaLearningFast2017} updates its parameters $\theta$ using an end-to-end optimization process to find the best initialization such that a new task can be learned quickly, \ie with few examples. More formally, given the loss $\ell_t$ for each task $t \in [[T]]$, \Maml\ minimizes the expected task loss after an \emph{inner loop} or \emph{adaptation} phase, computed by a few steps of gradient descent initialized at the model's current parameters:
\begin{equation}
    \Lcal_{\Maml}(\theta) := \E_{t \sim \eta} [\ell_t(\theta - \alpha \nabla \ell_t(\theta))],
\end{equation}
with $\eta$ the distribution of the meta-training tasks and $\alpha$ the learning rate for the adaptation phase. For simplicity, we take a single step of gradient update in this equation.

In what follows, we establish our theoretical analysis for the popular methods \Proto\ and \Maml. We add their improved variations respectively called Infinite Mixture Prototypes~\cite{pmlr-v97-allen19b} (\IMP) and Meta-Curvature~\cite{ParkO19} (\MC) in the experiments to validate our findings.

\section{Understanding Meta-learning Algorithms through MTR Theory}\label{sec:understand}

\subsection{Link between MTR and Meta-learning}

Recently, \cite{pmlr-v139-wang21ad} has shown that meta-learning algorithms that only optimize the last layer in the inner-loop, solve the same underlying optimization procedure as multi-task learning. In particular, their contributions have the following implications:
\begin{enumerate}
    \item For \emph{Metric-based algorithms}, the majority of methods can be seen as MTR problems. This is true, in particular, for \Proto\ and \IMP\ algorithms considered in this work.
    \item In the case of \emph{Gradient-based algorithms}, such methods as ANIL~\cite{Raghu2020Rapid} and MetaOptNet~\cite{leeMetaLearningDifferentiableConvex2019} that do not update the embeddings during the inner-loop, can be also seen as multi-task learning. However, \Maml\ and \MC\ in principal do update the embeddings even though there exists strong evidence suggesting that the changes in the weights during their inner-loop are mainly affecting the last layer \cite{Raghu2020Rapid}. Consequently, we follow \cite{pmlr-v139-wang21ad} and use this assumption to analyze \Maml\ and \MC\ in MTR framework as well.
    \item In practice, \cite{pmlr-v139-wang21ad} showed that the mismatch between the multi-task and the actual episodic training setup leads to a negligible difference.
\end{enumerate}

\noindent In the following section, we start by empirically verifying that
the behavior of meta-learning methods reveals very distinct features when looked at through the prism of the considered MTR theoretical assumptions. 

\subsection{What happens in practice?}\label{sec:practice}
To verify whether theoretical results from MTR setting are also insightful for episodic training used by popular meta-learning algorithms, we first investigate the natural behavior of \Maml\ and \Proto\ when solving FSC tasks on the popular \emph{miniImageNet}~\cite{raviOPTIMIZATIONMODELFEWSHOT2017} and \emph{tieredImageNet}~\cite{renMetaLearningSemiSupervisedFewShot2018} datasets. The full experimental setup is detailed in \cref{sec:expe_setup} and in the Appendix. Additional experiments for \emph{Omniglot}~\cite{lakeHumanlevelConceptLearning2015} benchmark dataset portraying the same behavior are also postponed to the Appendix.

To verify Assumption 1 from MTR theory, we want to compute singular values of $\rmW$ during the meta-training stage and to follow their evolution. In practice, as $T$ is typically quite large, we propose a more computationally efficient solution that is to calculate the condition number only for the last batch of $N$ predictors (with $N \ll T$) grouped in the matrix $\rmW_N \in \sR^{N\times k}$ that capture the latest dynamics in the learning process. We further note that $\sigma_i(\rmW_N \rmW_N^{\top}) = \sigma_i^2(\rmW_N),\ \forall i \in [[N]]$ implying that we can calculate the SVD of $\rmW_N \rmW_N^{\top}$ (or $\rmW_N^{\top} \rmW_N$ for $k \leq N$) and retrieve the singular values from it afterwards. 
We now want to verify whether $\rvw_t$ cover all directions in the embedding space and track the evolution of the ratio of singular values $\kappa(\rmW_N)$ during training. 
For the first assumption to be satisfied, we expect $\kappa(\rmW_N)$ to decrease gradually during the training thus improving the generalization capacity of the learned predictors and preparing them for the target task. To verify the second assumption, the norm of the linear predictors should not increase with the number of tasks seen during training, \textit{i.e.}, $\|\rvw\|_2 = O(1)$ or, equivalently, $\|\rmW\|_F^2 = O(T)$ and $\|\rmW_N\|_F = O(1)$.

\begin{figure*}[!t]
    \centering
    \includegraphics[width=0.32\linewidth]{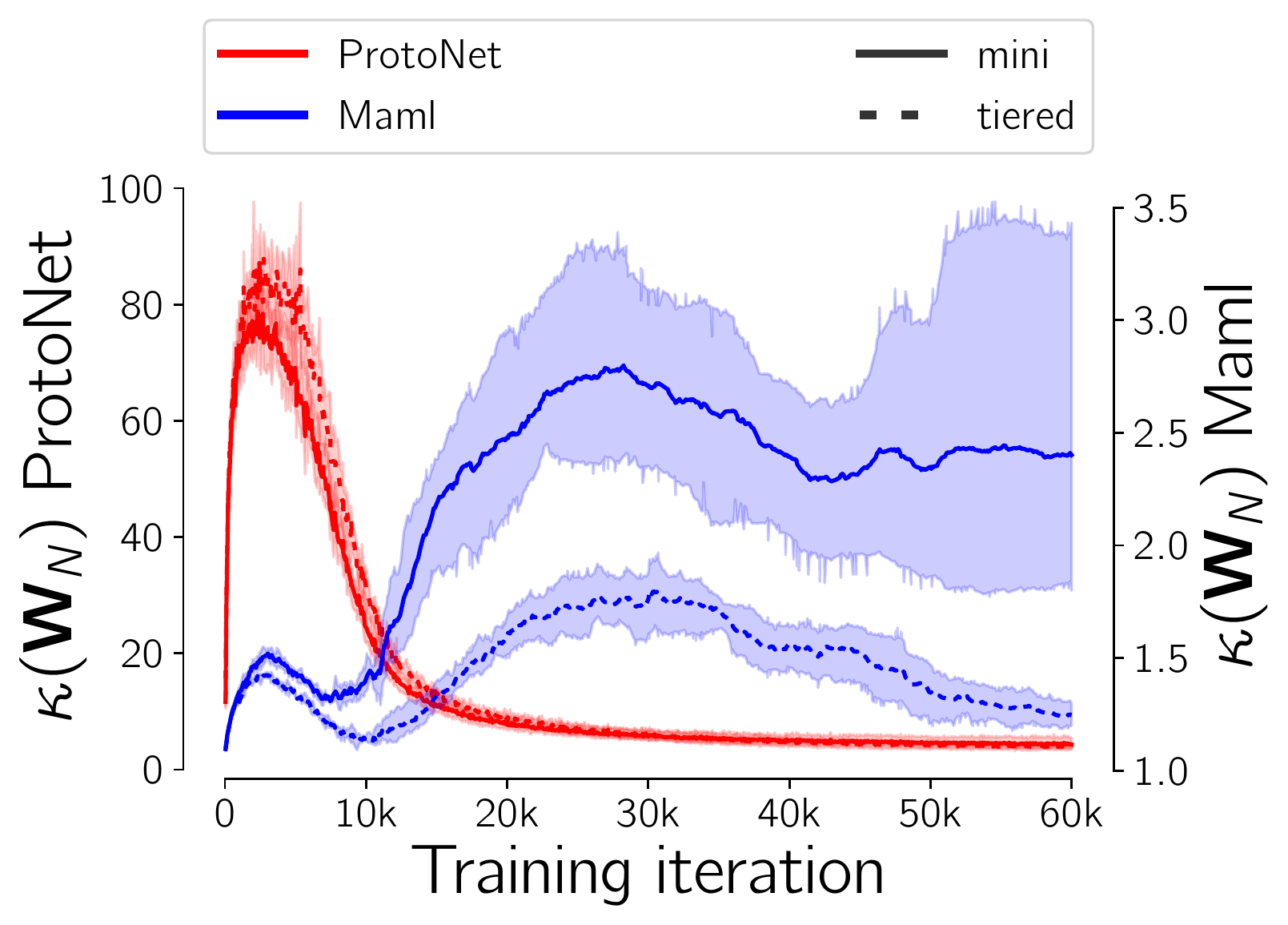}
    \hspace{0.9em}
    \includegraphics[width=0.32\linewidth]{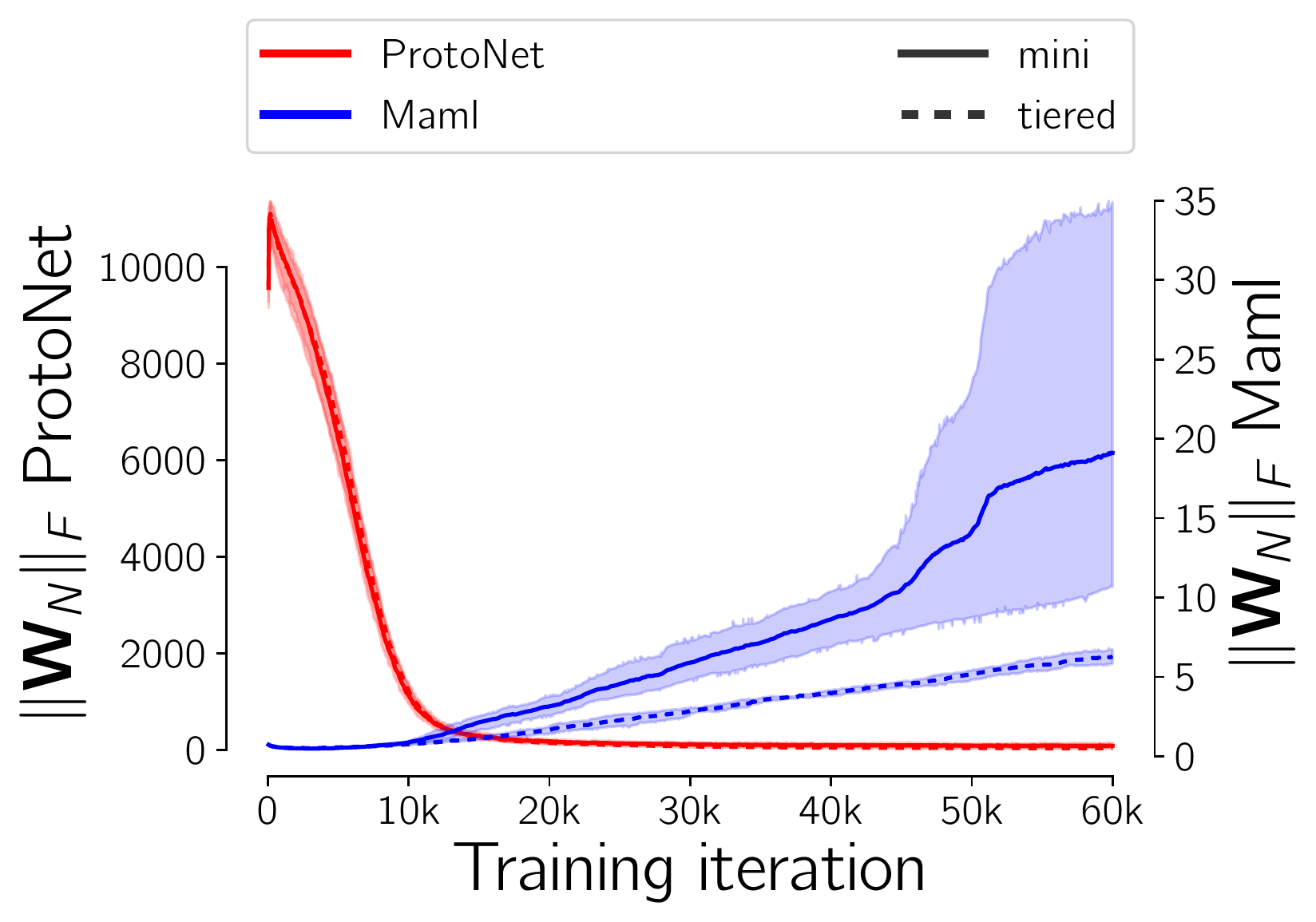}
    \hspace{0.9em}
    \includegraphics[width=0.28\linewidth]{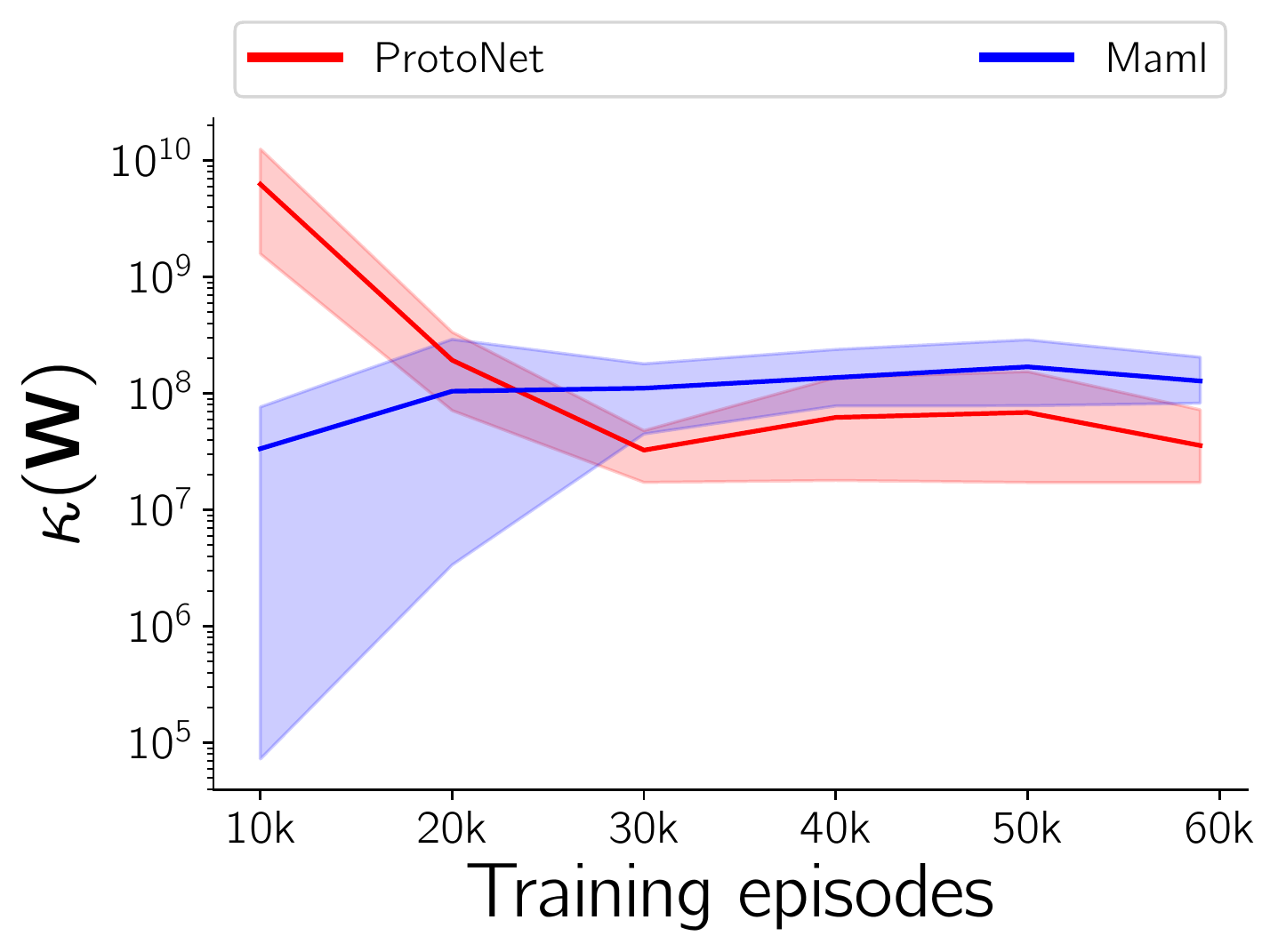}
    \\
    \caption{{Evolution of $\kappa(\rmW_N)$, $\|\rmW_N\|_F$ and $\kappa(\rmW)$ (\emph{in log scale}) during the training of \Proto\ (\emph{red, left axes}) and \Maml\ (\emph{blue, right axes}) on miniImageNet (\emph{mini, solid lines}) and tieredImageNet (\emph{tiered, dashed lines}) with 5-way 1-shot episodes. 
    }}
    \label{fig:training_curves}
\end{figure*}

For gradient-based methods, linear predictors are directly the weights of the last layer of the model. Indeed, for each task, the model learns a batch of linear predictors and we can directly take the weights for $\rmW_N$.
Meanwhile, metric-based methods do not use linear predictors but compute a similarity between features. In the case of \Proto, the similarity is computed with respect to class prototypes that are mean features of the instances of each class. For the Euclidean distance, \emph{this is equivalent to a linear model} with the prototype of a class acting as the linear predictor of this class \cite{snellPrototypicalNetworksFewshot2017}. This means that we can apply our analysis directly to the \emph{prototypes} computed by \Proto. In this case, the matrix $\rmW^*$ will be the matrix of the optimal prototypes and we can then take the \emph{prototypes} computed for each task as our matrix $\rmW_N$.

From \cref{fig:training_curves}, we can see that for \Maml\ (\emph{blue}), both $\|\rmW_N\|_F$ (\emph{left}) and $\kappa(\rmW_N)$ (\emph{middle}) increase with the number of tasks seen during training, whereas \Proto\ (\emph{red}) naturally learns prototypes with a good coverage of the embedding space, and minimizes their norm.
Since we compute the singular values of the last $N$ predictors in $\kappa(\rmW_N)$, we can only compare the \emph{overall behavior} throughout training between methods. For the sake of completeness, we also compute $\kappa(\rmW)$ (\emph{right}) at different stages in the training. To do so, we fix the encoder $\phi_T$ learned after $T$ episodes and recalculate the linear predictors of the $T$ past training episodes with this fixed encoder. We can see that $\kappa(\rmW)$ of \Proto\ also decreases during training and reach a lower final value than $\kappa(\rmW)$ of \Maml. This confirms that the dynamics of $\kappa(\rmW_N)$ and $\kappa(\rmW)$ are similar whereas the values $\kappa(\rmW_N)$ between methods should not be directly compared.
The behavior of $\kappa(\rmW)$ also validate our finding that \Proto\ learns to cover the embedding space with prototypes.
This behavior is rather peculiar as neither of the two methods explicitly controls the theoretical quantities of interest, and still, \Proto\ manages to do it implicitly. 

\subsection{The case of Meta-Learning algorithms}
\label{sec:grad_vs_metric}

The differences observed above for the two methods call for a deeper analysis of their behavior. 

\noindent \textbf{\Proto}\phantom{.} We start by first explaining why \Proto\ learns prototypes that cover the embedding space efficiently. This result is given by the following theorem (cf. Appendix for the proof):
\begin{theorem}\label{th:norm_proto}\emph{(Normalized \Proto)} \\
If $\forall i\ \|\rvc_i\| = 1$, then $\forall \hat{\phi} \in \argmin_{\phi} \Lcal_{proto}(S,Q,\phi)$, the matrix of the optimal prototypes $\rmW^*$ is well-conditioned, \ie $\kappa(\rmW^*) = O(1)$.
\end{theorem}

This theorem explains the empirical behavior of \Proto\ in FSC task: the minimization of its objective function naturally minimizes the condition number when the norm of the prototypes is low. 

In particular, it implies that norm minimization seems to initiate the minimization of the condition number seen afterwards due to the contrastive nature of the loss function minimized by \Proto. We confirm this latter implication through experiments in \cref{sec:expe} showing that norm minimization is enough for considered metric-based methods to obtain the well-behaved condition number and that minimizing both seems redundant. 

\textbf{\Maml}\phantom{.} Unfortunately, the analysis of \Maml\ in the most general case is notoriously harder, as even expressing its loss function and gradients in the case of an overparametrized linear regression model with only 2 parameters requires using a symbolic toolbox for derivations \cite{sha_maml_2021}. 
To this end, we resort to the linear regression model considered in this latter paper and defined as follows. We assume for all $t \in [[T]]$ that the task parameters $\boldsymbol{\theta}_t$ are normally distributed with $\boldsymbol{\theta}_t \sim \mathcal{N}(\bm{0}_d,\bm{I}_d)$, the inputs $\rvx_t \sim \mathcal{N}(\bm{0}_d,\bm{I}_d)$ and the output $y_t \sim \mathcal{N}(\langle\boldsymbol{\theta}_t, \rvx_t\rangle, 1)$. For each $t$, we consider the following learning model and its associated square loss:
\begin{equation}
\hat{y}_t = \langle \rvw_t, \rvx_t\rangle, \quad \ell_t = \mathbb{E}_{p(\rvx_t,y_t|\boldsymbol{\theta}_t)} (y_t-\langle \rvw_t, \rvx_t\rangle)^2.   
\label{eq:setup_maml}
\end{equation}
We can now state the following result.
\begin{proposition}\label{prop:maml_kappa}
Let $\forall t \in [[T]]$, $\boldsymbol{\theta}_t \sim \mathcal{N}(\bm{0}_d,\bm{I}_d)$, $\rvx_t \sim \mathcal{N}(\bm{0}_d,\bm{I}_d)$ and $y_t \sim \mathcal{N}(\langle \boldsymbol{\theta}_t, \rvx_t\rangle, 1)$. Consider the learning model from \cref{eq:setup_maml}, let $\boldsymbol{\Theta}_i := [\boldsymbol{\theta}_i, \boldsymbol{\theta}_{i+1}]^T$, and denote by $\widehat{\rmW}_2^i$ the matrix of last two predictors learned by \Maml\ at iteration $i$ starting from $\widehat{\rvw}_0 = \bm{0}_d$. Then, we have that: \\
\begin{equation}
    \forall i, \quad \kappa(\widehat{\rmW}_2^{i+1})\geq \kappa(\widehat{\rmW}_2^i),\ \text{ if } \sigma_\text{min}(\boldsymbol{\Theta}_i) = 0.
\end{equation}
\end{proposition}

This proposition provides an explanation of why \Maml\ may tend to increase the ratio of singular values during the iterations. Indeed, the condition when this happens indicates that the optimal predictors forming matrix $\boldsymbol{\Theta}_i$ are linearly dependent implying that its smallest singular values becomes equal to 0. While this is not expected to be the case for all iterations, we note, however, that in FSC task the draws from the dataset are in general not i.i.d. and thus may correspond to co-linear optimal predictors. In every such case, the condition number is expected to remain non-decreasing, as illustrated in \cref{fig:training_curves} (\emph{left}) where for \Maml, contrary to \Proto, $\kappa(\rmW_N)$ exhibits plateaus but also intervals where it is increasing. 

This highlights a major difference between the two approaches: \Maml\ does not specifically seek to diversify the learned predictors, while \Proto\ does.
\cref{prop:maml_kappa} gives a hint to the essential difference between the methods studied. On the one hand, \Proto\ constructs its classifiers directly from the data of the current task and they are independent of the other tasks. On the other hand, for \Maml, the weights of the classifiers are reused between tasks and only slightly adapted to be specific to each task. This limits the generalization capabilities of the linear predictors learned by \Maml\ since they are based on predictors from previous tasks.

\subsection{Enforcing the assumptions}\label{sec:toy_problem}

\textbf{Why should we force the assumptions ?}\phantom{.} From the results obtain by \cite{duFewShotLearningLearning2020}, and with the same assumptions, we can easily make appear $\kappa(\rmW^*)$ to obtain a more explicit bound:
\begin{proposition}\label{prop:new_bound}
If $\forall t \in [[T]], \|\rvw_t^*\| = O(1)$ and $\kappa(\rmW^*) = O(1)$, and $\rvw_{T+1}$ follows a distribution $\nu$ such that $\| \E_{\rvw \sim \nu}[\rvw\rvw^\top]\| \leq O \left(\frac{1}{k} \right)$, then
\begin{equation}
    \text{ER}(\hat{\phi}, \hat{\rvw}_{T+1}) \leq O\left(\frac{C(\Phi)}{n_1T}\cdot \kappa(\rmW^*) +\frac{k}{n_2}\right).
\end{equation}
\end{proposition}

\cref{prop:new_bound} suggests that the terms $\|\rvw_t^*\|$ and $\kappa(\rmW^*)$ underlying the assumptions directly impact the tightness of the established bound on the excess risk. The full proof can be found in the Appendix.

\noindent \textbf{Can we force the assumptions ?}\phantom{.} According to the previous result, satisfying the assumptions from MTR theory is expected to come in hand with better performance. However all the terms involved refer to optimal predictors, that we cannot act upon. Thus, we aim to answer the following question: \\
\indent \textit{Given $\rmW^*$ such that $\kappa(\rmW^*)\gg 1$, can we learn $\widehat{\rmW}$ with $\kappa(\widehat{\rmW})\approx 1$ while solving the underlying classification problems equally well?} \\
\indent While obtaining such a result for any distribution seems to be very hard in the considered learning setup, we provide a constructive proof for the existence of a distribution for which the answer to the above-mentioned question is positive in the case of two tasks. The latter restriction comes out of the necessity to analytically calculate the singular values of $\rmW$ but we expect our example to generalize to more general setups and a larger number of tasks as well.
\begin{proposition}
Let $T=2$, $\sX \subseteq \sR^d$ be the input space and $\sY = \{-1,1\}$ be the output space. Then, there exist distributions $\mu_1$ and $\mu_2$ over $\sX \times \sY$, representations $\widehat{\phi}\neq \phi^*$ and matrices of predictors $\widehat{\rmW} \neq \rmW^*$ that satisfy the data generating model (\cref{eq:data_gen_model}) with $\kappa(\widehat{\rmW})\approx 1$ and $\kappa(\rmW^*) \gg 1$.
\label{prop:example}
\end{proposition}

\noindent See Appendix for full proof and illustration. The established results show that even when $\rmW^*$ does not satisfy Assumptions 1-2 in the $\phi^*$ space, it may still be possible to learn $\widehat{\phi}$ such that the optimal predictors do satisfy them.

\noindent \textbf{How to force the assumptions ?}\phantom{.}  This can be done either by considering the constrained problem (\cref{eq:meta_train_hat}) or by using a more common strategy that consists in adding $\kappa(\rmW)$ and $\|\rmW\|_F^2$ as regularization terms (\cref{eq:meta_train_reg}):
\begin{align}
\begin{split}
    \displaystyle\widehat{\phi}, \widehat{\rmW} = \argmin_{\phi, \rmW} \frac{1}{Tn_1}\sum_{t=1}^T\sum_{i=1}^{n_1} \ell(y_{t,i},\langle \rvw_t, \phi(\rvx_{t,i})\rangle) \text{ s.t. } \kappa(\rmW) = O(1), \|\rvw_t\| = 1\,,
    \label{eq:meta_train_hat} \\
\end{split} \\
\begin{split}
    \displaystyle\widehat{\phi}, \widehat{\rmW} = \argmin_{\phi, \rmW} \frac{1}{Tn_1}\sum_{t=1}^T\sum_{i=1}^{n_1} \ell(y_{t,i},\langle \rvw_t, \phi(\rvx_{t,i})\rangle) + \lambda_1 \kappa(\rmW) + \lambda_2 \|\rmW\|_F^2\,.
    \label{eq:meta_train_reg}
\end{split}
\end{align}

To the best of our knowledge, such regularization terms based on insights from the advances in MTR theory have never been used in the literature before. We refer the reader to the Appendix for more details about them.

\subsection{Positioning with respect to Previous Work}\label{sec:related_work}

\textbf{Understanding meta-learning}\phantom{.} While a complete theory for meta-learning is still lacking, several recent works aim to shed light on phenomena commonly observed in meta-learning by evaluating different intuitive heuristics. For instance, \cite{Raghu2020Rapid} investigate whether \Maml\  
works well due to rapid learning with significant changes in the representations when deployed on target task, or due to feature reuse where the learned representation remains almost intact. 
In \cite{goldblumUnravelingMetaLearningUnderstanding2020}, the authors explain the success of meta-learning approaches by their capability to either cluster classes more tightly in feature space (task-specific adaptation approach), or to search for meta-parameters that lie close in weight space to many task-specific minima (full fine-tuning approach). Finally, the effect of the number of shots on the FSC 
accuracy was studied in \cite{CaoLF20} for \Proto\
. More recently, \cite{yeHowTrainYour2021a} studied the impact of the permutation of labels when training \Maml. 
Our paper investigates a new aspect of meta-learning that has never been studied before and, unlike \cite{Raghu2020Rapid}, \cite{CaoLF20}, \cite{goldblumUnravelingMetaLearningUnderstanding2020} and \cite{yeHowTrainYour2021a}, provides a more complete experimental evaluation with the two different approaches of meta-learning. \\
\textbf{Normalization}\phantom{.} Multiple methods in the literature introduce a normalization of their features either to measure cosine similarity instead of Euclidean distance \cite{gidarisDynamicFewShotVisual2018, qiLowShotLearningImprinted2018, chenCLOSERLOOKFEWSHOT2019} or because of the noticed improvement in their performance \cite{wangSimpleShotRevisitingNearestNeighbor2019, tian2020rethinking, pmlr-v139-wang21ad}. 
In this work, we proved in \cref{sec:grad_vs_metric} above that for \Proto\ prototypes normalization is enough to achieve a good coverage of the embedding space, and we empirically show in \cref{sec:metric} below that it leads to better performance. Since we only normalize the prototypes and not all the features, we do not measure cosine similarity. Moreover, with our \cref{th:norm_proto}, we give explanations through MTR theory regarding the link between the coverage of the representation space and performance. \\
\textbf{Common regularization strategies}\phantom{.}  
In general, we note that regularization in meta-learning (i) is applied to either the weights of the whole neural network \cite{balajiMetaRegDomainGeneralization,Yin2020Meta-Learning}, or (ii) the predictions \cite{jamalTaskAgnosticMetaLearning2019, goldblumUnravelingMetaLearningUnderstanding2020} or (iii) is introduced via a prior hypothesis biased regularized empirical risk minimization \cite{pentinaPACBayesianBoundLifelong,kuzborskij2017fast,denevi2018incremental,denevi2018learning,DeneviCGP19}.
Contrary to the first group of methods and weight decay approach \cite{kroghSimpleWeightDecay1992}, we do not regularize the whole weight matrix learned by the neural network but the linear predictors of its last layer. While weight decay is used to avoid overfitting by penalizing large magnitudes of weights, our goal is to keep the classification margin unchanged during the training to avoid over-/under-specialization to source tasks. Similarly, spectral normalization proposed by \cite{MiyatoKKY18} does not affect the condition number $\kappa$. Second, we regularize the singular values of the matrix of linear predictors obtained in the last batch of tasks instead of the predictions used by the methods of the second group (\textit{e.g.}, using the theoretic-information quantities in \cite{jamalTaskAgnosticMetaLearning2019}). Finally, the works of the last group are related to the online setting with convex loss functions only and do not specifically target the spectral properties of the learned predictors.

\section{Impact of enforcing theoretical assumptions}\label{sec:expe}

\subsection{Experimental Setup}\label{sec:expe_setup}

We consider on three benchmark datasets for FSC, namely: 
1) \textbf{Omniglot} \cite{lakeHumanlevelConceptLearning2015} consisting of $1{,}623$ classes with $20$ images/class of size $28 \times 28$; 2) \textbf{miniImageNet} \cite{raviOPTIMIZATIONMODELFEWSHOT2017} consisting of $100$ classes with $600$ images of size $84 \times 84$/class; 3) \textbf{tieredImageNet} \cite{renMetaLearningSemiSupervisedFewShot2018} consisting of $779{,}165$ images divided into $608$ classes.

For each dataset, we follow a common experimental protocol used in \cite{finnModelAgnosticMetaLearningFast2017,chenCLOSERLOOKFEWSHOT2019} and use a four-layer convolution backbone with 64 filters (C64) as done by \cite{chenCLOSERLOOKFEWSHOT2019}. We also provide experiments with the ResNet-12 architecture (R12) \cite{leeMetaLearningDifferentiableConvex2019} and we follow the recent practice to initialize the models with the weights pretrained on the entire meta-training set \cite{ye2020fewshot, rusu2018meta, qiao2018few}. For \Proto, we use the released code of \cite{ye2020fewshot} with a temperature of 1 instead of tuning it for all settings to be fair with other settings and methods. For \Maml, we follow the exact implementation details from \cite{yeHowTrainYour2021a} since the code was not available.
We measure the performance using the top-1 accuracy with $95\%$ confidence intervals, reproduce the experiments with 4 different random seeds and average the results over $2400$ test tasks. For all FSC experiments, unless explicitly stated, we use the regularization parameters $\lambda_1 = \lambda_2 = 1$ in the regularized problem (\cref{eq:meta_train_reg}). We refer the reader to the Appendix for all the hyperparameters used.\footnote{Code for the experiments is available at \url{https://github.com/CEA-LIST/MetaMTReg}.}

\begin{figure*}[t]
    \begin{minipage}{0.58\linewidth}
        \includegraphics[width=0.49\textwidth]{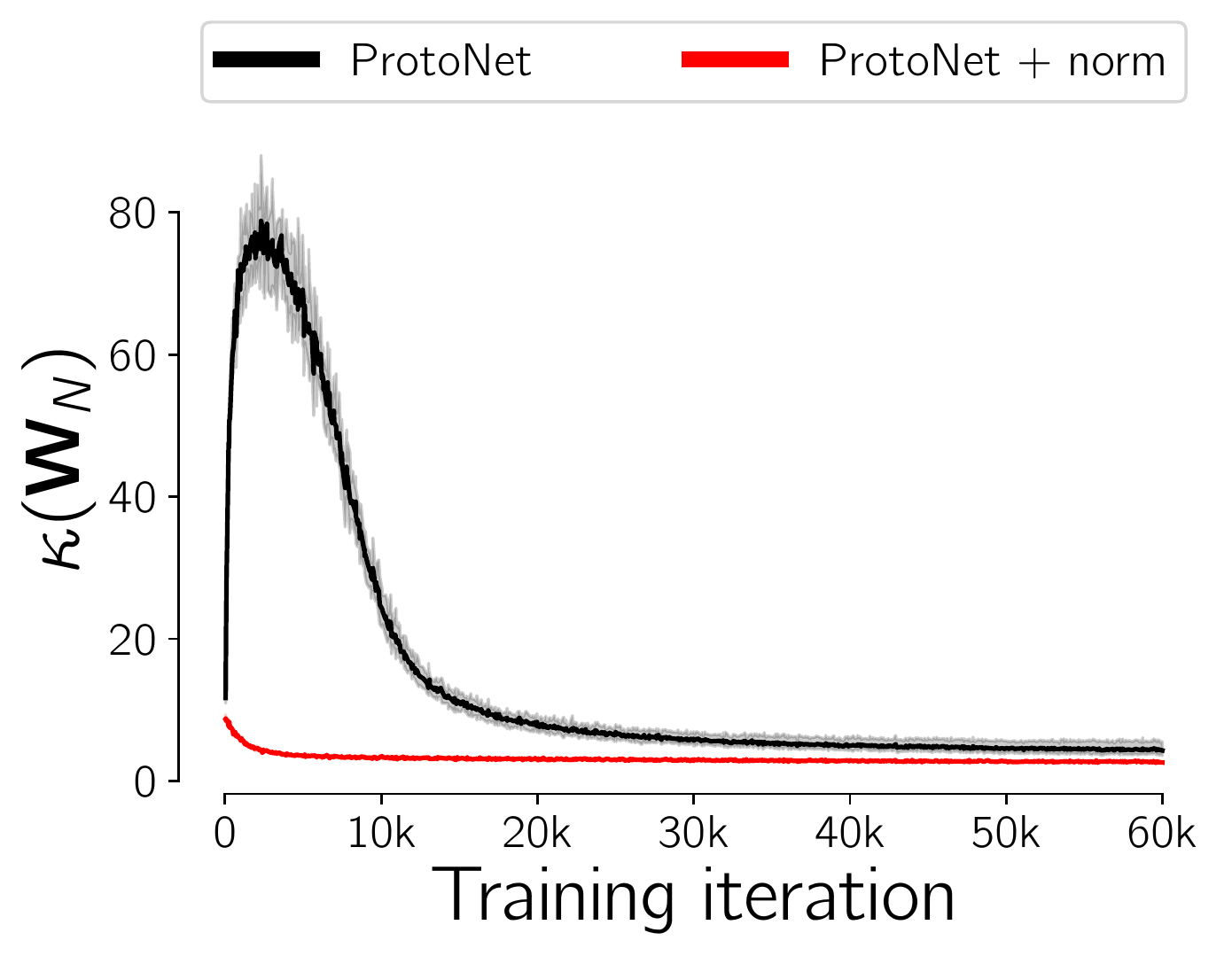}
        \includegraphics[width=0.49\textwidth]{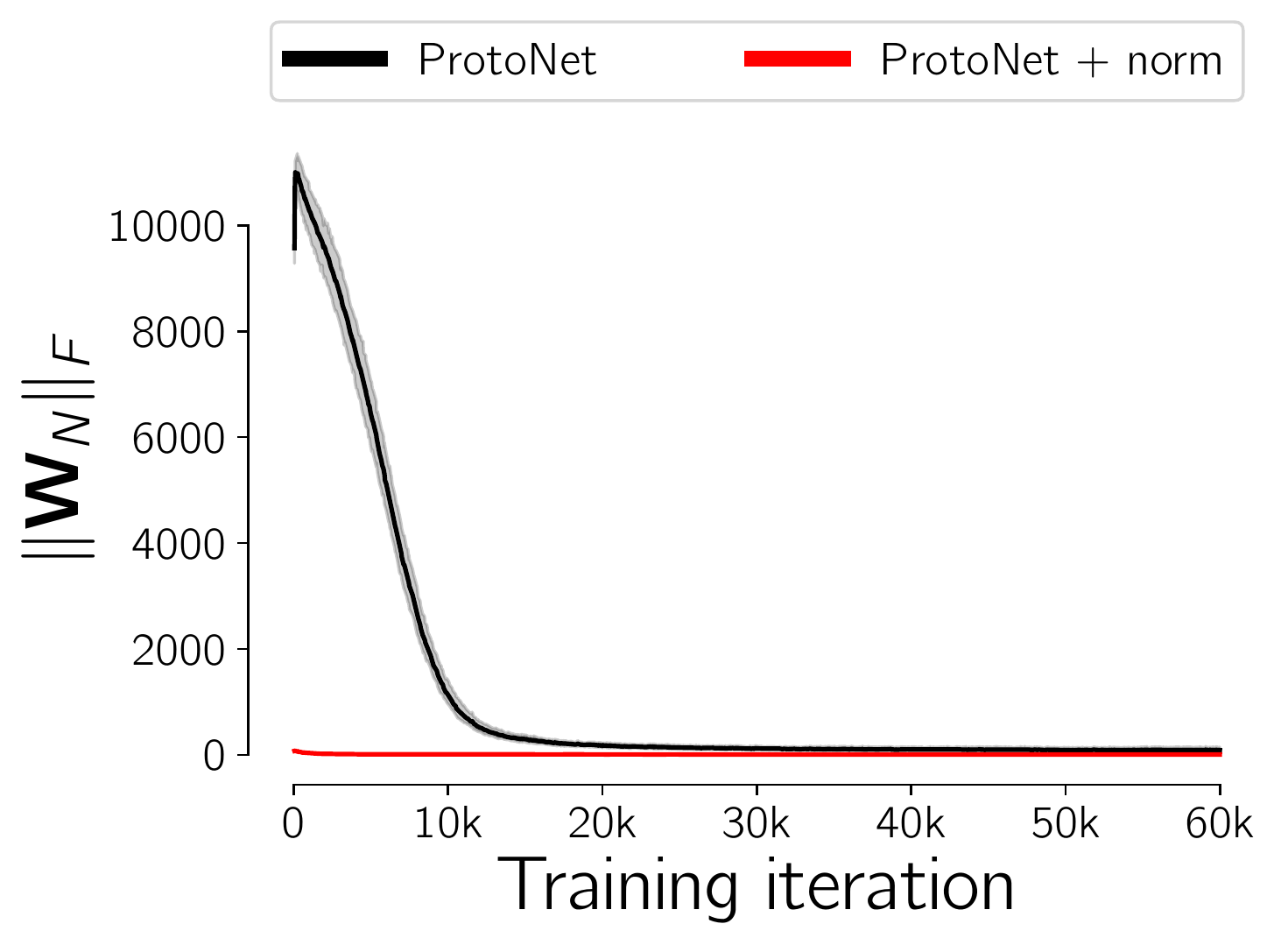} \\
        \includegraphics[width=0.49\textwidth]{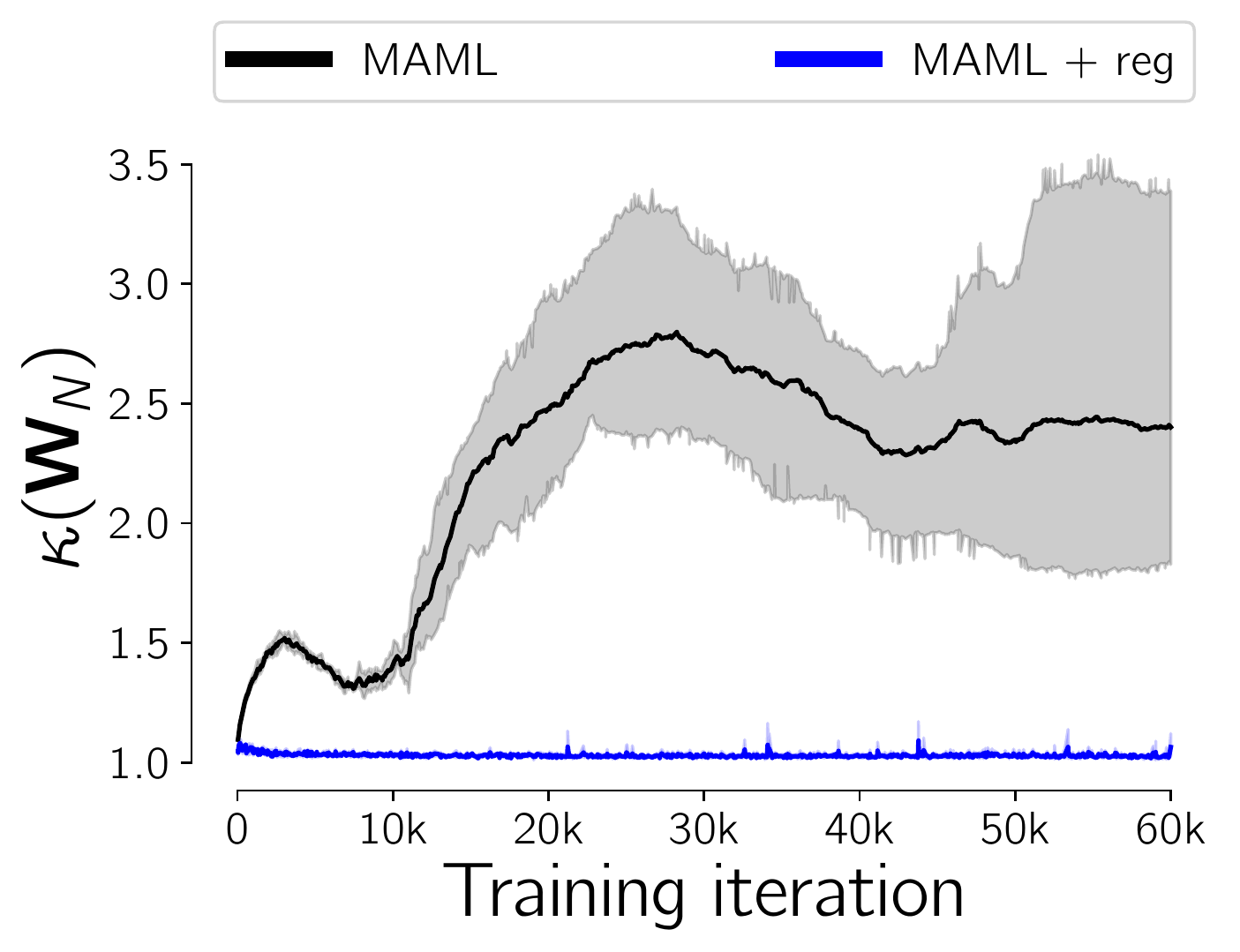}
        \includegraphics[width=0.49\textwidth]{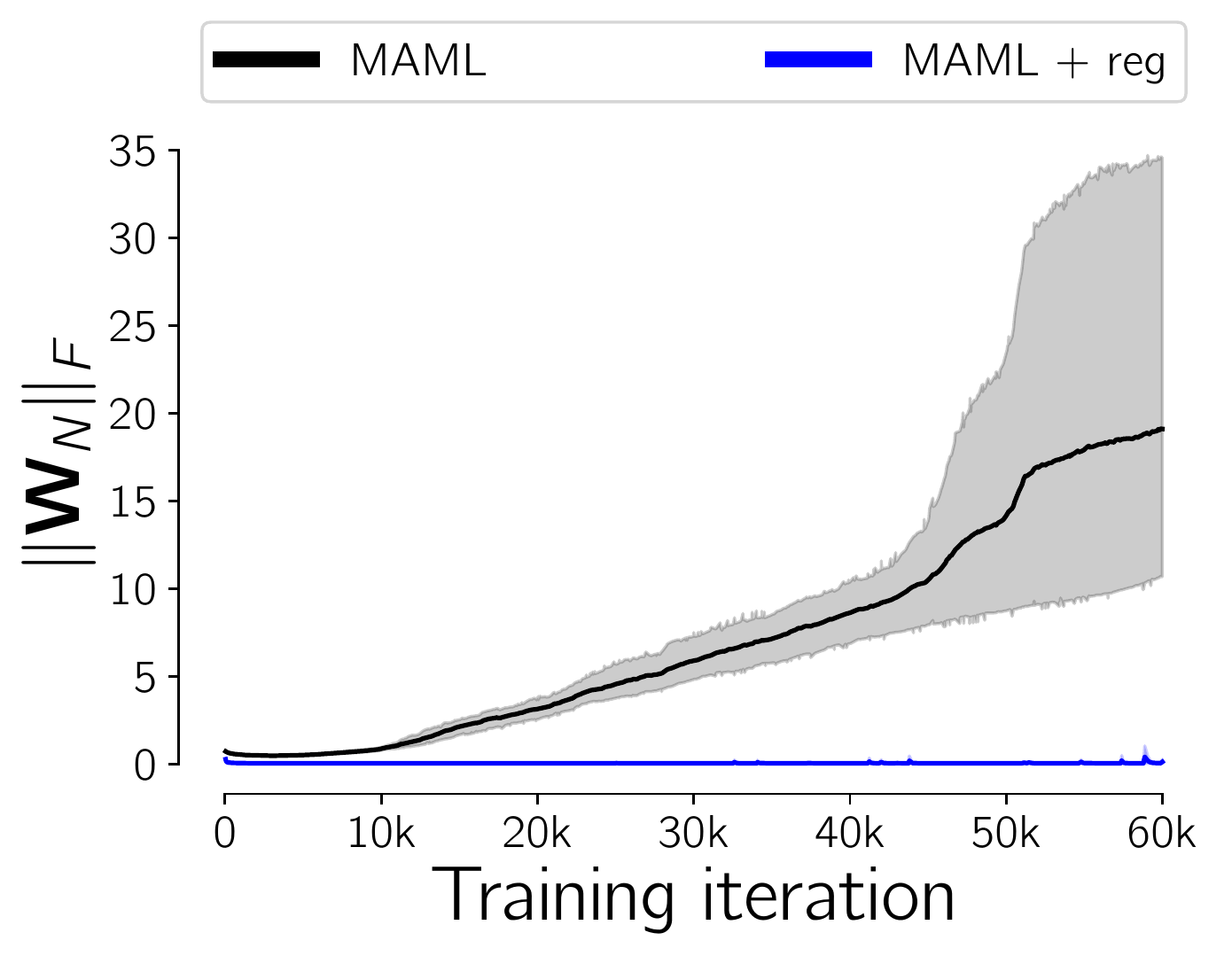}
    \end{minipage}
    \hspace{5pt}
    \begin{minipage}{0.39\linewidth}
        \includegraphics[width=\linewidth]{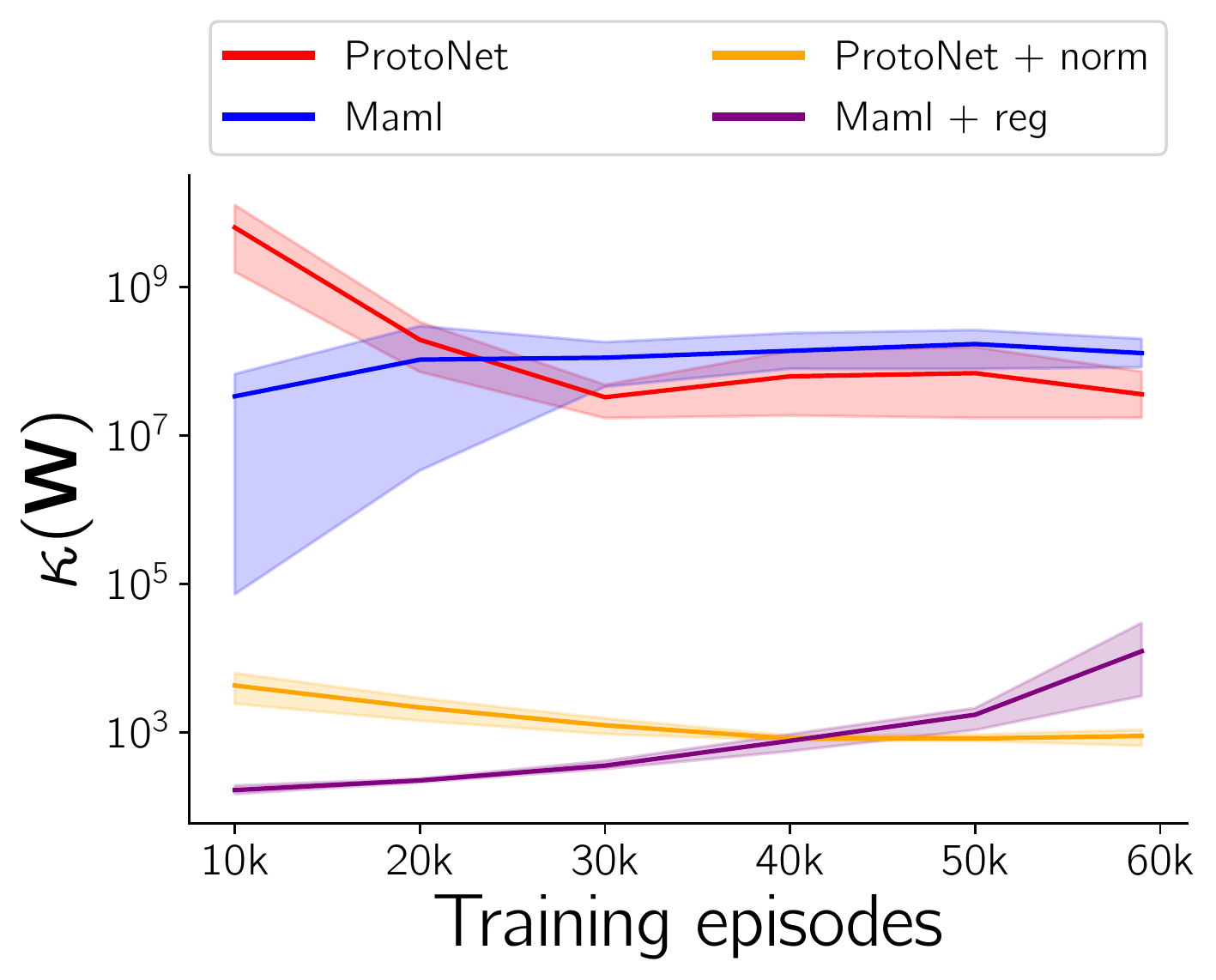}
    \end{minipage}
   \captionof{figure}{
    Evolution of $\kappa(\rmW_N)$ (\emph{left}), $\|\rmW_N\|_F$ (\emph{middle}) and $\kappa(\rmW)$ (\emph{right}) on 5-way 1-shot episodes from \emph{miniImageNet},
    for \Proto\ (\emph{top}), \Maml\ (\emph{bottom}) and their regularized or normalized counterparts. 
    }
    \label{fig:acc_curves}
\end{figure*}

\begin{figure*}[t]
    \centering
    \includegraphics[width=0.3\linewidth]{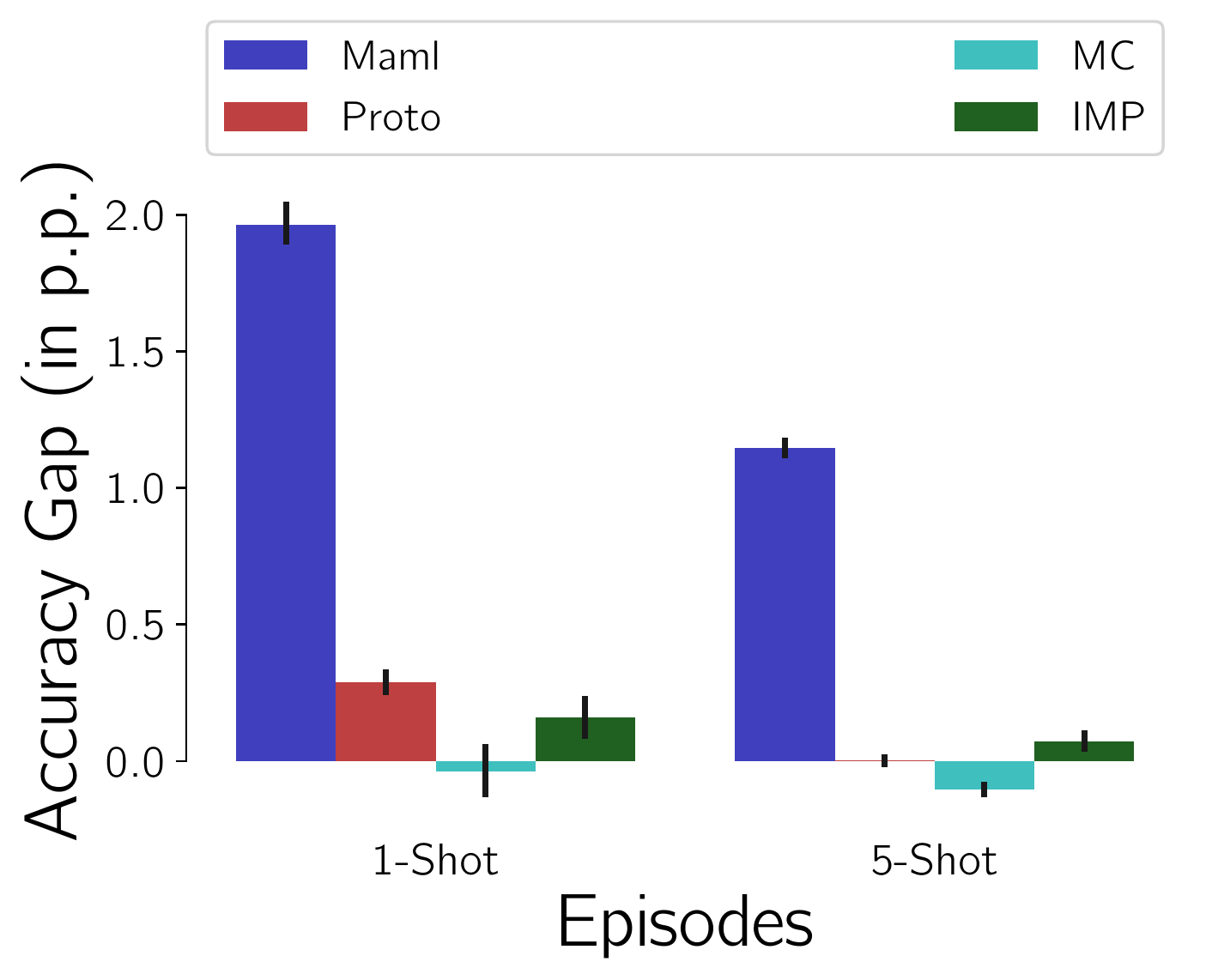}
    \includegraphics[width=0.3\linewidth]{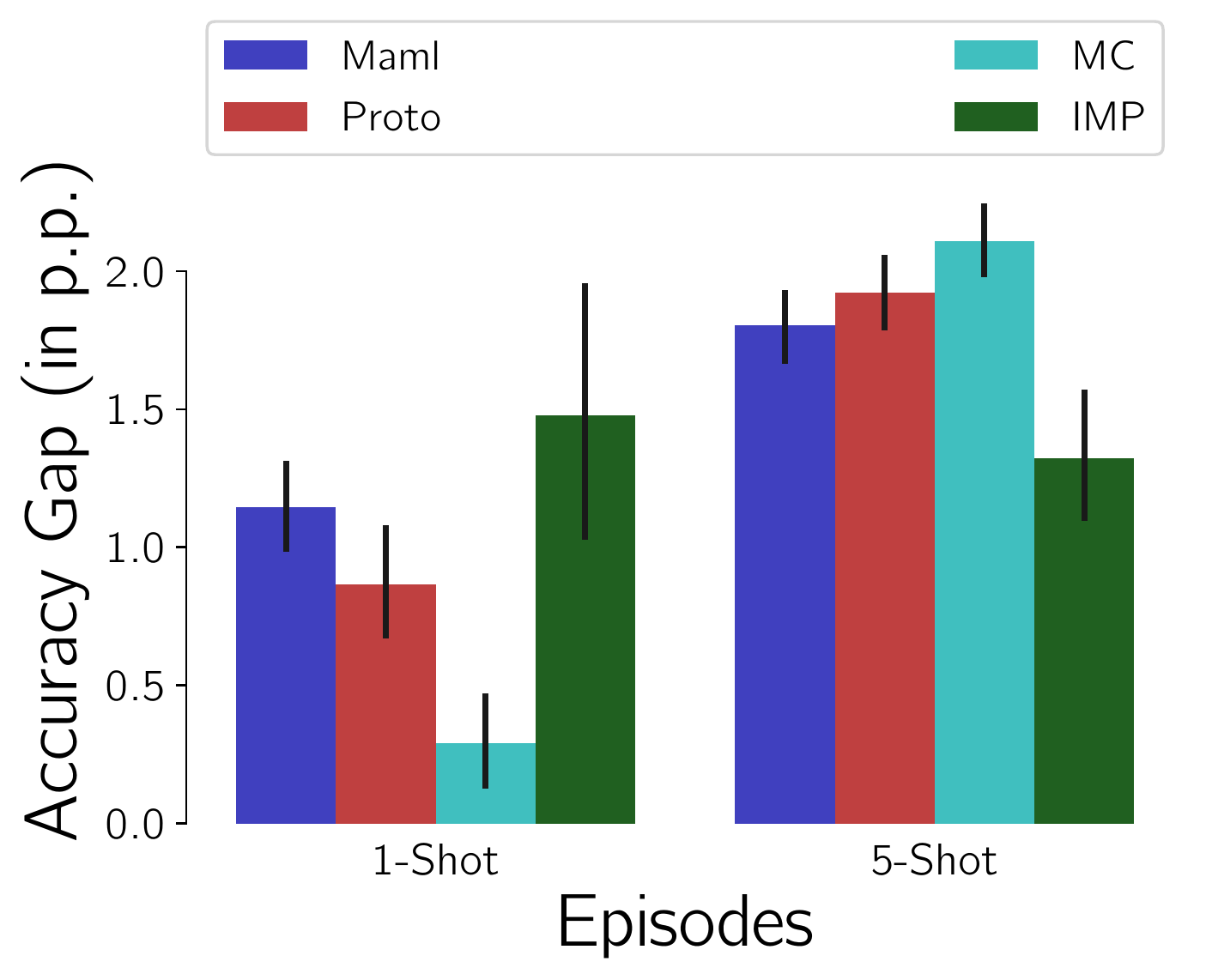}
    \includegraphics[width=0.3\linewidth]{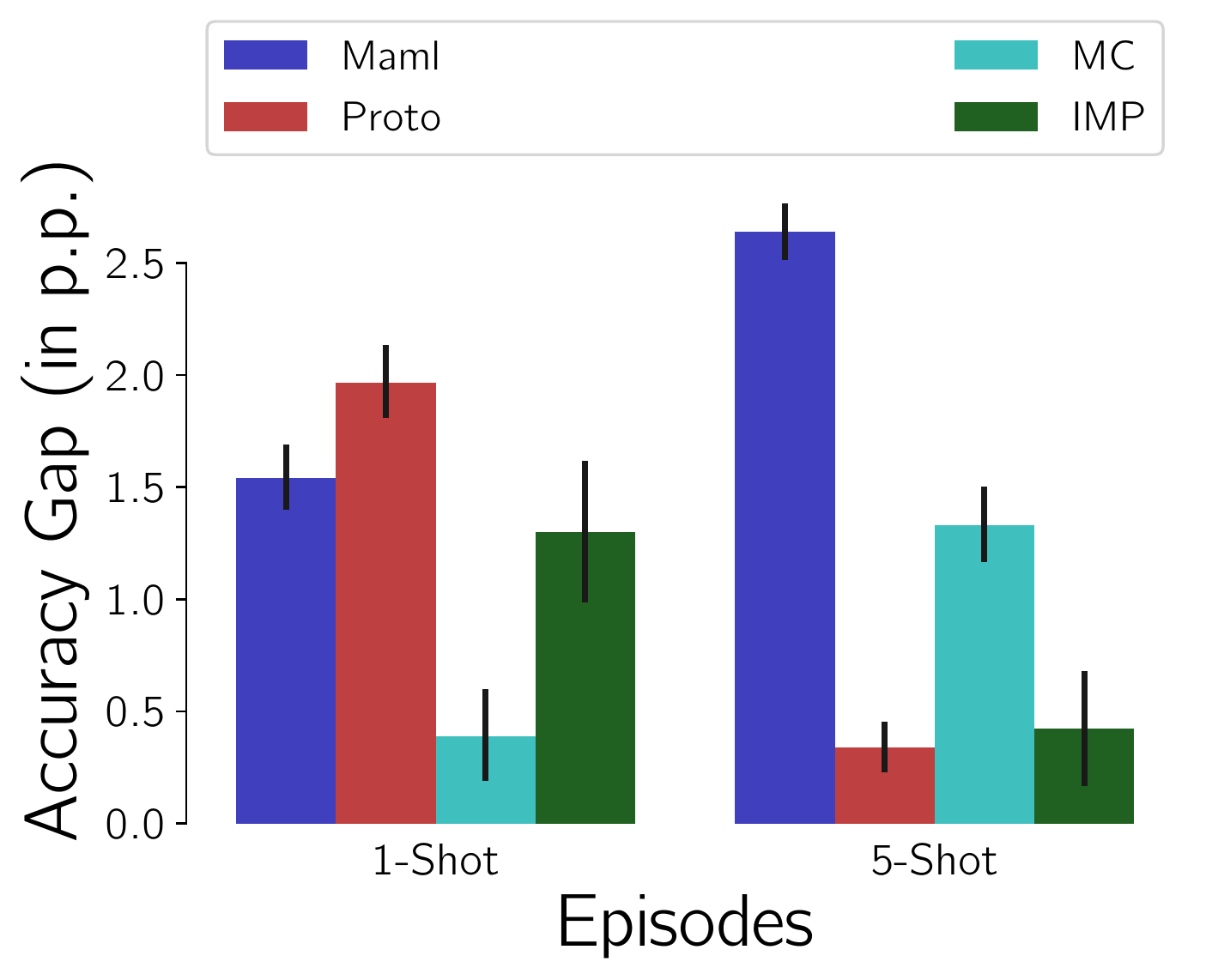}

    \caption{Accuracy gap (in p.p.) when adding the normalization of prototypes for \Proto\ (\emph{red}) and \IMP\ (\emph{green}), and both spectral and norm regularization for \Maml\ (\emph{blue}) and \MC\ (\emph{purple}) enforcing the theoretical assumptions on Omniglot (\emph{left}), miniImagenet (\emph{middle}) and tieredImagenet (\emph{right}) datasets. 
    }
    \label{fig:barplots}
\end{figure*}

\begin{table*}[!htb]
    \centering
\begin{minipage}[!t]{0.99\linewidth}
\centering
\subfloat[][\label{tab:sota_perfs}]{
\resizebox{0.9\linewidth}{!}{%
    \begin{tabular}{@{}llcccc@{}} \toprule
    & & \multicolumn{2}{c}{miniImageNet 5-way} & \multicolumn{2}{c}{tieredImageNet 5-way} \\
    \cmidrule(lr){3-4}
    \cmidrule(lr){5-6}
    Model & Arch. & 1-shot & 5-shot & 1-shot & 5-shot \\
    \midrule
    \multicolumn{6}{l}{Gradient-based algorithms} \\
    \midrule
    \Maml \cite{finnModelAgnosticMetaLearningFast2017} & C64 & 48.70 $\pm$ 1.84 & 63.11 $\pm$ 0.92 &  - & - \\
    \textsc{Anil} \cite{Raghu2020Rapid} & C64 & 48.0 $\pm$ 0.7 & 62.2 $\pm$ 0.5 & - & - \\
    Meta-SGD \cite{liMetaSGDLearningLearn2017} & C64 & \textbf{50.47 $\pm$ 1.87} & 64.03 $\pm$ 0.94 & - & - \\
    TADAM \cite{NEURIPS2018_66808e32} & R12 & 58.50 $\pm$ 0.30 & 76.70 $\pm$ 0.30 & - & - \\
    MC \cite{ParkO19} & R12 & 61.22 $\pm$ 0.10 & 75.92 $\pm$ 0.17 & \textbf{66.20 $\pm$ 0.10} & \textbf{82.21 $\pm$ 0.08} \\
    MetaOptNet \cite{leeMetaLearningDifferentiableConvex2019} & R12 & 62.64 $\pm$ 0.61 & 78.63 $\pm$ 0.46 & \underline{65.99 $\pm$ 0.72} & { \underline{81.56 $\pm$ 0.53}} \\
    MATE\cite{NEURIPS2020_8989e07f} & R12 & 62.08 $\pm$ 0.64 & 78.64 $\pm$ 0.46 & - & - \\
    \\
    \Maml\ (Ours) & C64 & 47.93 $\pm$ 0.83 & 64.47 $\pm$ 0.69 & 50.08 $\pm$ 0.91 & 67.5 $\pm$ 0.79 \\
    \Maml\ + reg. (Ours) & C64 & 49.15 $\pm$ 0.85 & \textbf{66.43 $\pm$ 0.69} & 51.5 $\pm$ 0.9 & 70.16 $\pm$ 0.76 \\
    \MC\ (Ours) & C64 & 49.28 $\pm$ 0.83 & 63.74 $\pm$ 0.69 & \underline{55.16 $\pm$ 0.94} & 71.95 $\pm$ 0.77 \\
    \MC\ + reg. (Ours) & C64 & \underline{49.64 $\pm$ 0.83} & \underline{65.67 $\pm$ 0.70} & \textbf{55.85 $\pm$ 0.94} & \textbf{73.34 $\pm$ 0.76} \\
    \Maml\ (Ours) & R12 & 63.52 $\pm$ 0.20 & 81.24 $\pm$ 0.14 & 63.96 $\pm$ 0.23 & \underline{81.79 $\pm$ 0.16} \\
    \Maml\ + reg. (Ours) & R12 & \textbf{64.04 $\pm$ 0.22} & \textbf{82.45 $\pm$ 0.14} & 64.32 $\pm$ 0.23 & \underline{81.28 $\pm$ 0.11} \\ 
    
    \midrule
    \multicolumn{6}{l}{Metric-based algorithms} \\
    \midrule
    \Proto\ \cite{snellPrototypicalNetworksFewshot2017} & C64 & 46.61 $\pm$ 0.78 & 65.77 $\pm$ 0.70 & -- & -- \\
    \IMP\ \cite{pmlr-v97-allen19b} & C64 & 49.6 $\pm$ 0.8 & \textbf{68.1 $\pm$ 0.8} & -- & -- \\
    SimpleShot \cite{wangSimpleShotRevisitingNearestNeighbor2019} & C64 & \underline{49.69 $\pm$ 0.19} & 66.92 $\pm$ 0.17 & 51.02 $\pm$ 0.20 & 68.98 $\pm$ 0.18 \\
    Relation Nets \cite{sungLearningCompareRelation2018} & C64 & \underline{50.44 $\pm$ 0.82} & 65.32 $\pm$ 0.70 & -- & -- \\
    SimpleShot \cite{wangSimpleShotRevisitingNearestNeighbor2019} & R18 & \textbf{62.85 $\pm$ 0.20} & \textbf{80.02 $\pm$ 0.14} & \textbf{69.09 $\pm$ 0.22} & \textbf{84.58 $\pm$ 0.16} \\
    CTM \cite{liFindingTaskRelevantFeatures2019} & R18 & \underline{62.05 $\pm$ 0.55} & 78.63 $\pm$ 0.06 & 64.78 $\pm$ 0.11 & 81.05 $\pm$ 0.52 \\
    DSN \cite{simonAdaptiveSubspacesFewShot2020} & R12 & \underline{62.64 $\pm$ 0.66} & 78.83 $\pm$ 0.45 & 66.22 $\pm$ 0.75 & 82.79 $\pm$ 0.48 \\
    \\
    \Proto\ (Ours) & C64 & 49.53 $\pm$ 0.41 & 65.1 $\pm$ 0.35 & 51.95 $\pm$ 0.45 & 71.61 $\pm$ 0.38 \\
    \Proto + norm. (Ours) & C64 & \underline{50.29 $\pm$ 0.41} & \underline{67.13 $\pm$ 0.34} & \textbf{54.05 $\pm$ 0.45} & \underline{71.84 $\pm$ 0.38} \\
    \IMP\ (Ours) & C64 & 48.85 $\pm$ 0.81 & 66.43 $\pm$ 0.71 & 52.16 $\pm$ 0.89 & 71.79 $\pm$ 0.75 \\
    \IMP\ + norm. (Ours) & C64 & \textbf{50.69 $\pm$ 0.8} & \underline{67.29 $\pm$ 0.68} & \underline{53.46 $\pm$ 0.89} & \textbf{72.38 $\pm$ 0.75} \\
    \Proto\ (Ours) & R12 & 59.25 $\pm$ 0.20 & 77.92 $\pm$ 0.14 & 41.39 $\pm$ 0.21 & 83.06 $\pm$ 0.16 \\
    \Proto + norm. (Ours) & R12 & \textbf{62.69 $\pm$ 0.20} & \textbf{80.95 $\pm$ 0.14} & \textbf{68.44 $\pm$ 0.23} & \textbf{84.20 $\pm$ 0.16} \\
 
    \bottomrule
    \end{tabular}%
    }
}
\end{minipage}
\begin{minipage}[!t]{0.99\linewidth}
\centering
\subfloat[][\label{tab:mtl_perfs}]{
\resizebox{0.90\linewidth}{!}{%
\begin{tabular}{@{}llcccc@{}} \toprule
    & & \multicolumn{2}{c}{miniImageNet 5-way} & \multicolumn{2}{c}{tieredImageNet 5-way} \\
    \cmidrule(lr){3-4}
    \cmidrule(lr){5-6}
    Model & Arch. & 1-shot & 5-shot & 1-shot & 5-shot \\
    \midrule
    MTL & R12 & 55.73 $\pm$ 0.18 & 76.27 $\pm$ 0.13 & 62.49 $\pm$ 0.21 & 81.31 $\pm$ 0.15 \\
    MTL + norm. & R12 & 59.49 $\pm$ 0.18 & \textbf{77.3 $\pm$ 0.13} & \textbf{66.66 $\pm$ 0.21} & \textbf{83.59 $\pm$ 0.14} \\
    MTL + reg. (Ours) & R12 & \textbf{61.12 $\pm$ 0.19} & \underline{76.62 $\pm$ 0.13} & \underline{66.28 $\pm$ 0.22} & 81.68 $\pm$ 0.15 \\

    \bottomrule
    \end{tabular}
    }
}
\end{minipage}
\caption{Performance comparison on FSC benchmarks. (a) FSC models. (b) MTL models \cite{pmlr-v139-wang21ad}. \emph{For a given architecture}, \textbf{bold} values are the highest accuracy and \underline{underlined values} are near-highest accuracies (less than 1-point lower).}
\label{fig:results_perfs}
\end{table*}

\subsection{Metric-based Methods}\label{sec:metric}

Theorem \ref{th:norm_proto} tells us that with normalized class prototypes that act as linear predictors, \Proto\ naturally decreases the condition number of their matrix. Furthermore, since the prototypes are directly the image features, adding a regularization term on the norm of the prototypes makes the model collapse to the trivial solution which maps all images to 0.
To this end, we choose to ensure the theoretical assumptions for metric-based methods (\Proto\ and \IMP) only with prototype normalization, by using the normalized prototypes $\tilde{\rvw} = \frac{\rvw}{\|\rvw\|}$.
According to Theorem~\ref{th:norm_proto}, the normalization of the prototypes makes the problem similar to the constrained problem given in \cref{eq:meta_train_hat}.

As can be seen in \cref{fig:acc_curves}, the normalization of the prototypes has the intended effect on the condition number of the matrix of predictors. Indeed, $\kappa(\rmW_N)$ (\emph{left}) stay constant and low during training, and we achieve a much lower $\kappa(\rmW)$ (\emph{right}) than without normalization.
From \cref{fig:barplots}, we note that normalizing the prototypes from the very beginning of the training process has an overall positive effect on the obtained performance, and this gain is statistically significant in most of the cases according to the \emph{Wilcoxon signed-rank} test (p < 0.05) \cite{wilcoxon, David1956NonparametricSF}.

In \cref{tab:sota_perfs}, we compare the performance obtained against state-of-the-art algorithms behaving similarly to Instance Embedding algorithms~\cite{ye2020fewshot} such as \Proto, depending on the architecture used. Even with a ResNet-12 architecture, the proposed normalization still improves the performance to reach competitive results with the state-of-the-art. On the miniImageNet 5-way 5-shot benchmark, our normalized \Proto\ achieves 80.95\%, better than DSN (78.83\%), CTM (78.63\%) and SimpleShot (80.02\%).
We refer the reader to the Appendix for more detailed training curves.

\subsection{Gradient-based Methods}\label{sec:gradient}
Gradient-based methods learn a batch of linear predictors for each task, and we can directly take them as $\rmW_N$ to compute its SVD. 
In the following experiments, we consider the regularized problem of \cref{eq:meta_train_reg} for \Maml\ as well as Meta-Curvature (\MC). 
As expected, the dynamics of $\|\rmW_N\|_F$ and $\kappa(\rmW_N)$ during the training of the regularized methods remain bounded and the effect of the regularization is confirmed with the lower value of $\kappa(\rmW)$ achieved (cf. \cref{fig:acc_curves}).

The impact of our regularization on the results is quantified in \cref{fig:barplots} where a statistically significant accuracy gain is achieved in most cases, according to the \emph{Wilcoxon signed-rank} test (p < 0.05) \cite{wilcoxon, David1956NonparametricSF}.
In \cref{tab:sota_perfs}, we compare the performance obtained to state-of-the-art gradient-based algorithms. We can see that our proposed regularization is globally improving the results, even with a bigger architecture such as ResNet-12 and with an additional pretraining. On the miniImageNet 5-way 5-shot benchmark, with our regularization \Maml\ achieves 82.45\%, better than TADAM (76.70\%), MetaOptNet (78.63\%) and MATE with MetaOptNet (78.64\%).
We include in the Appendix ablative studies on the effect of each term in our regularization scheme for gradient-based methods, and more detailed training curves.

\subsection{Multi-Task learning Methods}\label{sec:mtl}

We implement our regularization on a recent Multi-Task Learning (MTL) method\cite{pmlr-v139-wang21ad}, following the same experimental protocol.
The objective is to empirically validate our analysis on a method using the MTR framework.
As mentioned in Sec. \ref{sec:related_work}, the authors introduce feature normalization in their method, speculating that it improves coverage of the representation space \cite{pmlr-v119-wang20k}. Using their code, we reproduce their experiments on three different settings compared in \cref{tab:mtl_perfs}: the vanilla MTL, the MTL with feature normalization, and MTL with our proposed regularization on the condition number and the norm of the linear predictors. We use $\lambda_1 = 1$ in all the settings, and $\lambda_2 = 1$ in the 1-shot setting and $\lambda_2 = 0.01$ in the 5-shot settings. We include in the Appendix an ablative study on the effect of each term of the regularization. Our regularization, as well as the normalization, globally improve the performance over the non-normalized models. Notably, our regularization is the most effective when there is the less data which is well-aligned with the MTR theory in few-shot setting. We can also note that in most of the cases, the normalized models and the regularized ones achieve similar results, hinting that they may have a similar effect.
All of these results show that our analysis and our proposed regularization are also valid in the MTL framework. 

\section{Conclusion}\label{sec:conclusion}
In this paper, we studied the validity of the theoretical assumptions made in recent papers of Multi-Task Representation Learning theory when applied to popular metric- and gradient-based meta-learning algorithms. We found a striking difference in their behavior and provided both theoretical and experimental arguments explaining that metric-based methods satisfy the considered assumptions, while gradient-based
don't. We further used this as a starting point to implement a regularization strategy ensuring these assumptions and observed that it leads to faster learning and better generalization.

While this paper proposes an initial approach to bridging the gap between theory and practice for Meta-Learning, some questions remain open on the inner workings of these algorithms. In particular, being able to take better advantage of the particularities of the training tasks during meta-training could help improve the effectiveness of these
approaches. The similarity between the source and test tasks was not taken into account in this work, which is an additional assumption in the theory of \cite{duFewShotLearningLearning2020}. We provide a preliminary study using different datasets between the meta-training and meta-testing in the Appendix to foster future work on this topic.
Self-supervised meta-learning and multiple target tasks prediction are also important future perspectives for the application of meta-learning. 

\section{Acknowledgements}
This work was made possible by the use of the Factory-AI supercomputer, financially supported by the Ile-de-France Regional Council.

\bibliographystyle{splncs04}
\bibliography{references}

\clearpage

\appendix

\section{Appendix}

The supplementary material is organized as follows. Section~\ref{ax:MTR} provides an additional review on Multi-task Representation Learning Theory. Section~\ref{ax:proofsS3} provides the full proofs of the theoretical results discussed in the paper. Section~\ref{ax:details_reg} gives more details behind the proposed regularization terms. Section~\ref{ax:expe_setup} describes the full experimental setup with all the hyperparameters used. Section~\ref{ax:performancecomparisons} gives the detailed results used to derive Figure 3 in the paper. Section~\ref{ax:entropy} provides more experiments showing that further enforcing the condition number assumption for \Proto\ is unfavorable. In Section~\ref{ax:ablation}, we study the effect of each term in the proposed regularization.
Section~\ref{ax:cross-domain} provides a preliminary analysis in the out-of-domain setting.

\subsection{Review of Multi-task Representation Learning Theory} \label{ax:MTR}
We formulate the main results of the three main theoretical analyses of Multi-task Representation (MTR) Learning Theory provided in \cite{maurerBenefitMultitaskRepresentation2016,duFewShotLearningLearning2020,tripuraneniProvableMetaLearningLinear2020} in Table \ref{sota:meta_learning_theory} to give additional details for Sections 2.2 and 4.4 of our paper.

\renewcommand*{\arraystretch}{1.5}
\begin{table*}[tb]
    \caption{Overview of main theoretical contributions related to MTR learning with their assumptions, considered classes of representations and the obtained bounds on the excess risk. Here $\tilde{O}(\cdot)$ hides logarithmic factors.}
    \centering
    \resizebox{\textwidth}{!}{\begin{tabular}{l|l|c|c}
        \textbf{Paper} & \multicolumn{1}{c}{\textbf{Assumptions}} & $\Phi$ & \textbf{Bound}\\
        \cite{maurerBenefitMultitaskRepresentation2016} & \textbf{A1}. $\forall t \in [[T+1]],\ \mu_t \sim \eta$ & -- & $O\left(\frac{1}{\sqrt{n_1}}+\frac{1}{\sqrt{T}}\right)$\\\hline
        \multirow{6}{*}{\cite{duFewShotLearningLearning2020}} &
        \textbf{A2.0}. $\forall t$, $\|\rvw_t^*\| = \Theta(1)$ &  &  \\
        & \textbf{A2.1}. $\forall t$, $\Bar{\rvx}$ is $\rho^2$-subgaussian &  &  \\
        & \textbf{A2.2}. $\forall t \in [[T]], \exists c>0: \Sigma_t \succeq c\Sigma_{T+1}$ & \textbf{A2.1-2.4}, linear, $k\ll d$ & $O\left(\frac{kd}{cn_1T}+\frac{k}{n_2}\right)$\\
        & \textbf{A2.3}. $\frac{\sigma_1(\rmW^*)}{\sigma_k(\rmW^*)} = O(1)$ & \textbf{A2.3-2.5}, general, $k\ll d$ & $O\left(\frac{\gC(\Phi)}{n_1T}+\frac{k}{n_2}\right)$\\
         & \textbf{A2.4}. $\rvw_{T+1}^* \sim \mu_\rvw: ||\mathop{\E}_{\rvw\sim\mu_\rvw}[\rvw\rvw^T]||\leq O(\frac{1}{k})$ & \textbf{A2.1,2.5,2.6}, linear + $\ell_2$ regul., $k\gg d$ & $\sigma\Bar{R}\tilde{O}\left(\frac{\sqrt{\text{Tr}(\Sigma)}}{\sqrt{n_1T}}+\frac{\sqrt{||\Sigma||_2}}{\sqrt{n_2}}\right)$\\
        & \textbf{A2.5}. $\forall t$, $p_t = p, \Sigma_t = \Sigma$ & \textbf{A2.1,2.5,2.6,2.7}, two-layer NN (ReLUs+ $\ell_2$ regul.) & $\sigma\Bar{R}\tilde{O}\left(\frac{\sqrt{\text{Tr}(\Sigma)}}{\sqrt{n_1T}}+\frac{\sqrt{||\Sigma||_2}}{\sqrt{n_2}}\right)$\\
        & \textbf{A2.6}. Point-wise+unif. cov. convergence &  & \\
        & \textbf{A2.7}. Teacher network &  & \\
        \hline
        \multirow{4}{*}{\cite{tripuraneniProvableMetaLearningLinear2020}} & \textbf{A3.1}. $\forall t$, $\rvx\sim \mu_{\sX_t}$ is $\rho^2$-subgaussian & \multirow{4}{*}{\textbf{A1-4}, linear, $k\ll d$} & \multirow{4}{*}{$\tilde{O}\left(\frac{kd}{n_1T}+\frac{k}{n_2}\right)$} \\
        & \textbf{A3.2}. $\frac{\sigma_1(\rmW^*)}{\sigma_k(\rmW^*)} = O(1)$ and $\forall t, \|\rvw_t\| = \Theta(1)$ &  &  \\
        & \textbf{A3.3}. $\widehat{\rmW}$ learned using the Method of Moments &  &  \\
        & \textbf{A3.4}. $\rvw_{T+1}^*$ is learned using Linear Regression & & \\
        \hline
    \end{tabular}}
    \label{sota:meta_learning_theory}
\end{table*}

One may note that all the assumptions presented in this table can be roughly categorized into two groups. First one consists of the assumptions related to the data generating process (A1, A2.1, A2.4-7 and A3.1), technical assumptions required for the manipulated empirical quantities to be well-defined (A2.6) and assumptions specifying the learning setting (A3.3-4). We put them together as they are not directly linked to the quantities that we optimize over in order to solve the meta-learning problem. The second group of assumptions include A2.2 and A3.2: both defined as a measure of diversity between source tasks' predictors that are expected to cover all the directions of $\sR^k$ evenly. These assumptions is of primary interest as it involves the matrix of predictors optimized in Eq. 1 as thus one can attempt to force it in order for $\widehat{\rmW}$ to have the desired properties. 

Finally, we note that assumption A2.2 related to the covariance dominance can be seen as being at the intersection between the two groups. On the one hand, this assumption is related to the population covariance and thus is related to the data generating process that is supposed to be fixed. On the other hand, we can think about a pre-processing step that precedes the meta-train step of the algorithm and transforms the source and target tasks' data so that their sample covariance matrices satisfy A2.2. While presenting a potentially interesting research direction, it is not clear how this can be done in practice especially under a constraint of the largest value of $c$ required to minimize the bound. \cite{duFewShotLearningLearning2020} circumvent this problem by adding A2.5, stating that the task data marginal distributions are similar. 

\begin{figure*}[htb]
    \centering
    \begin{tabular}{c|c}
        \textbf{Violated assumptions} & \textbf{Satisfied assumptions}\\
        \includegraphics[trim= 50 50 50 45, clip, width = .49\linewidth]{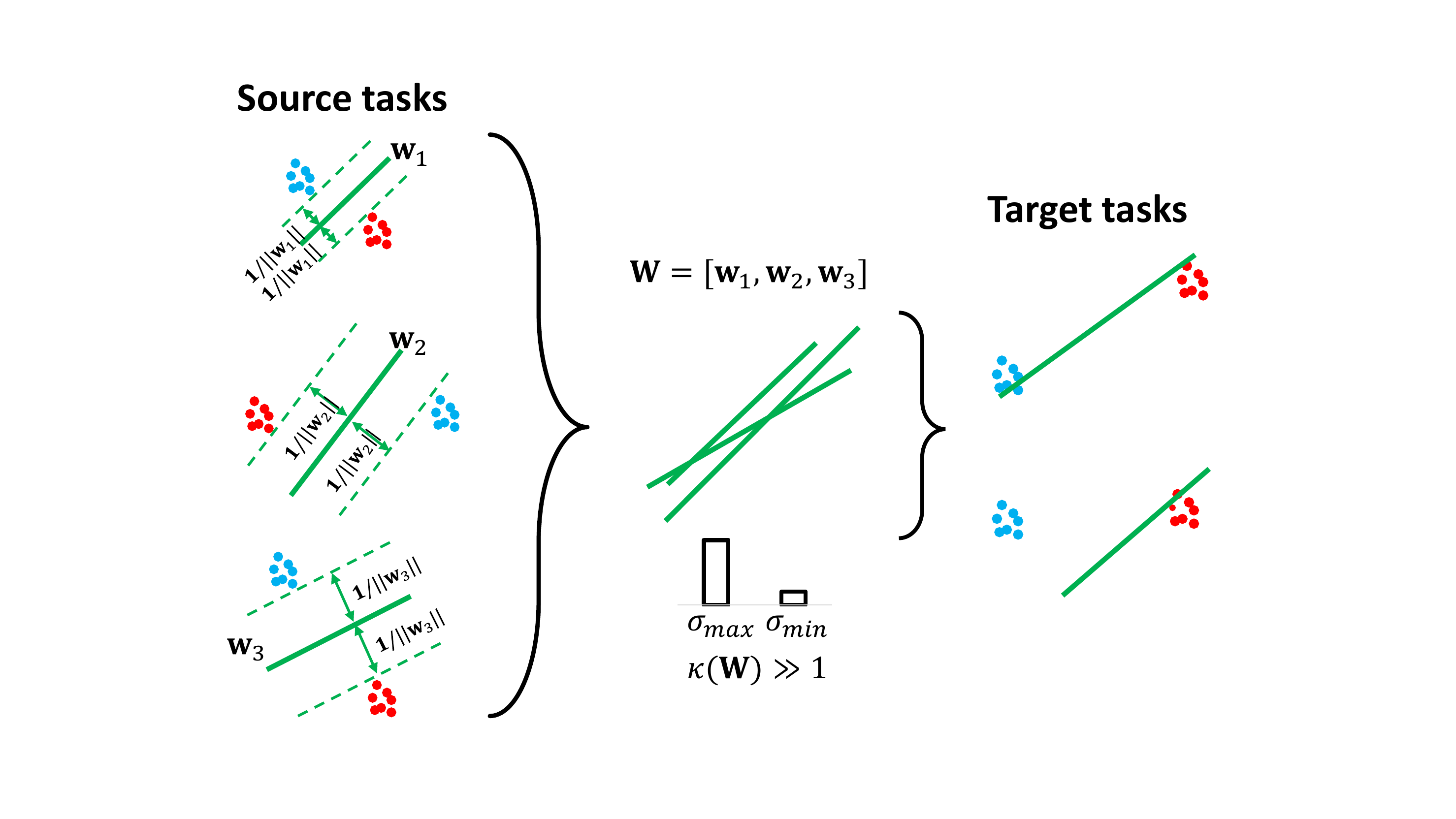}
        &
        \includegraphics[trim= 30 40 30 30, clip, width = .49\linewidth]{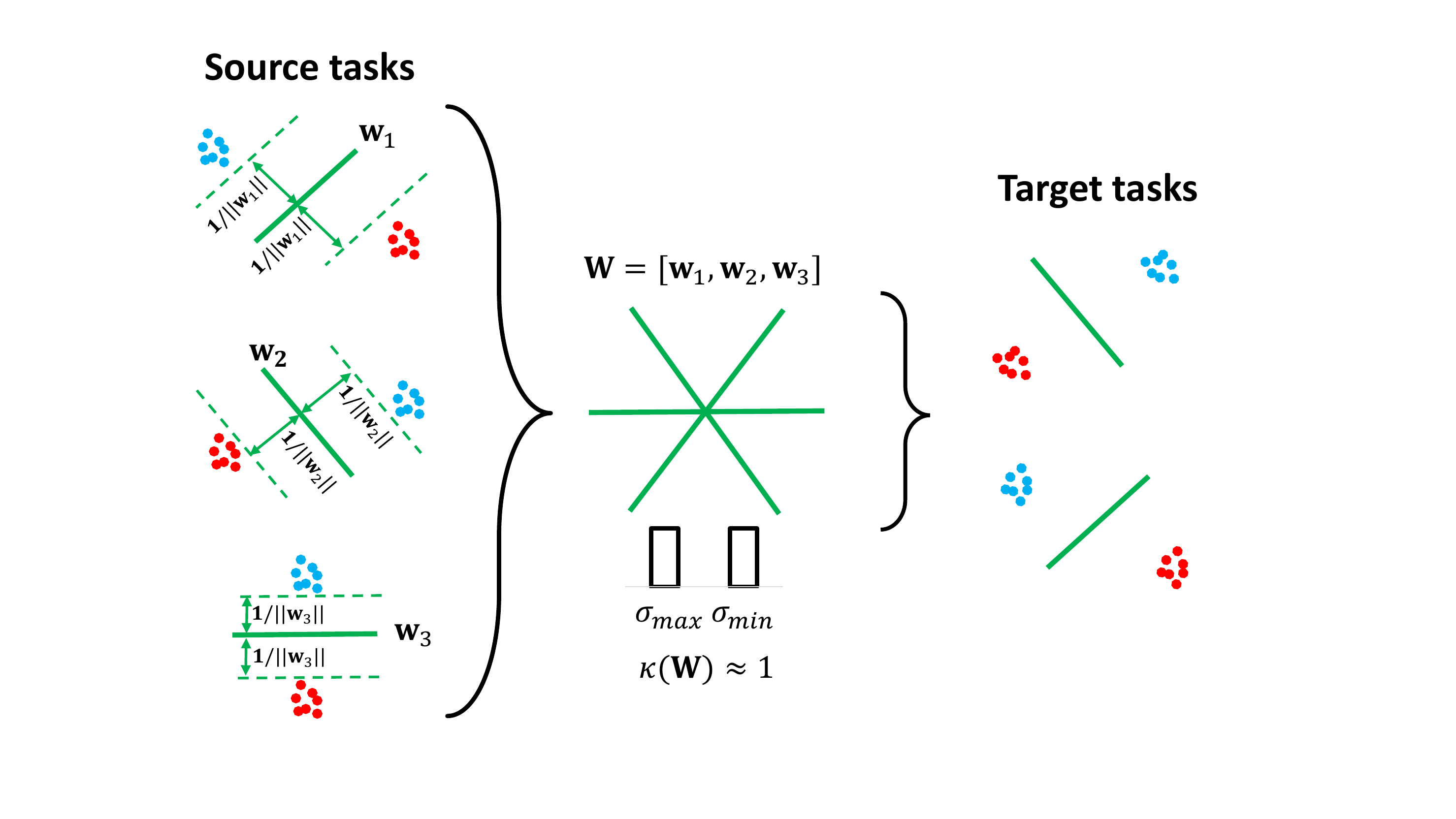}
    \end{tabular}
    \caption{Illustration of the intuition behind the assumptions derived from the MTR learning theory. \textbf{(left)} Lack of diversity and increasing norm of the linear predictors restrict them from being useful on the target task. \textbf{(right)} When the assumptions are satisfied, the linear predictors cover the embedding space evenly and their norm remains roughly constant on source tasks making them useful for a previously unseen task.}
    \label{fig:intuition0}
\end{figure*}

An intuition behind the main assumptions studied in this paper (Assumption 1 and 2 in this paper, and A2.0, A2.3, A3.2 in Table~\ref{sota:meta_learning_theory}) can be seen in Figure \ref{fig:intuition0}. When the assumptions do not hold, the linear predictors can be biased towards a single part of the space and over-specialized to the tasks. The representation learned will not generalize well to unseen tasks. If the assumptions are respected, the linear predictors are complementary and will not under- or over-specialize to the tasks seen. The representation learned can more easily adapt to the target tasks and achieve better generalization.

\subsection{Full proofs}
\label{ax:proofsS3}

\subsubsection{Proof of Theorem 1}
\begin{proof}

We start by recalling the prototypical loss $\Lcal_{proto}$ used during training of Prototypical Networks for a single episode with support set $S$ and query set $Q$:
\begin{align*}
\Lcal_{proto}(S,Q, \phi) &= \E_{(\rvq,i) \sim Q} \left[ - \log \frac{\exp(-d(\phi(\rvq), \rvc_{i}))}{\sum_{j} \exp{(- d(\phi(\rvq), \rvc_j))}} \right] \\
 &=  \underbrace{ \E_{(\rvq, i) \sim Q} \left[ d(\phi(\rvq), \rvc_{i}) \right] }_{(1)} \\
 & \quad + \underbrace{\E_{\rvq \sim Q} \log \sum_{j = 1}^{n} \exp{(-d(\phi(\rvq), \rvc_j))} }_{(2)}
\end{align*}
with $\rvc_i = \frac{1}{k} \sum_{\rvs \in S_i} \phi(\rvs)$ the prototype for class $i$, $S_i\subseteq S$ being the subset containing instances of $S$ labeled with class $i$.

\noindent
For \Proto, we consider the Euclidean distance between the representation of a query example $\phi(\rvq)$ and the prototype of a class $i$ $\rvc_i$:
\begin{align*}
-d(\phi(\rvq),\rvc_i) &= -\| \phi(\rvq) - \rvc_i \|^2_2 \\
 &= -\phi(\rvq)^\top \phi(\rvq) + 2 \rvc_i^\top \phi(\rvq) - \rvc^\top_i \rvc_i.
\end{align*}

\noindent Then, with respect to class $i$, the first term is constant and do not affect the softmax probabilities. The remaining terms are:
\begin{align*}
-d(\phi(\rvq), \rvc_i) &= 2 \rvc_i^\top \phi(\rvq) - \|\rvc_i\|^2_2 \\
 &= \frac{2}{|S_i|} \sum_{\rvs \in S_i } \phi(\rvs)^\top \phi(\rvq) - \|\rvc_i\|^2_2.
\end{align*}

\noindent
We can rewrite the first term in $\Lcal_{proto}$ as
\begin{align*}
    &\E_{(\rvq,i) \sim Q} \left[ d(\phi(\rvq), \rvc_{i}) \right ] \\
    &= - \E_{(\rvq,i) \sim Q} \left[ \frac{2}{|S_i|} \sum_{\rvs \in S_i} \phi(\rvs)^\top \phi(\rvq) - \|\rvc_{i}\|^2_2 \right ] \\
    &= - \E_{(\rvq,i) \sim Q} \left[ \frac{2}{|S_i|} \sum_{\rvs \in S_i} \phi(\rvs)^\top \phi(\rvq) \right ] \\
    &\quad + \E_{(\rvq,i) \sim Q} \left[ \|\rvc_{i}\|^2_2 \right],
\end{align*}

\noindent and the second term as
\begin{align*}
    & \E_{\rvq \sim Q} \left [ \log \sum_{j = 1}^{n} \exp{(-d(\phi(\rvq), \rvc_j))} \right ] \\
    &= \E_{\rvq \sim Q} \left [ \log \sum_{j = 1}^{n} \exp{(\frac{2}{|S_j|} \sum_{\rvs \in S_j} \phi(\rvs)^\top \phi(\rvq) - \|\rvc_j\|^2_2)} \right ] \\
    &= \E_{\rvq \sim Q} \left [ \log \sum_{j = 1}^{n} \exp{(2 \rvc_j^\top \phi(\rvq)-\|\rvc_j\|^2_2)} \right ] \\
    &= \E_{\rvq \sim Q} \left [ \log \left ( n \sum_{j = 1}^{n} \frac{1}{n} \left [ \exp{(2 \rvc_j^\top \phi(\rvq)-\|\rvc_j\|^2_2)} \right ] \right ) \right ] \\
    &= \E_{\rvq \sim Q} \left [ \log \sum_{j = 1}^{n} \frac{1}{n} \left [ \exp{(2 \rvc_j^\top \phi(\rvq)-\|\rvc_j\|^2_2)} \right ] + \log n \right ]. \\
\end{align*}

\noindent By dropping the constant part in the loss, we obtain:

\begin{align*}
    \Lcal_{proto}(S,Q,\phi) &= - \E_{(\rvq,i) \sim Q} \left[ \frac{2}{|S_i|} \sum_{\rvs \in S_i} \phi(\rvs)^\top \phi(\rvq) \right ] \\
    & + \E_{\rvq \sim Q} \left [ \log \sum_{j = 1}^{n} \frac{1}{n} \left [ \exp{(2 \rvc_j^\top \phi(\rvq))} \right ] \right ].
\end{align*}
Let us note $\Scal^d$ the hypersphere of dimension $d$, and $\Mcal(\Scal^d)$ the set of all possible Borel probability measures on $\Scal^d$. $\forall \mu \in \Mcal(\Scal^d), u \in \Scal^d$, we further define the continuous and Borel measurable function:
$$
U_\mu (u) := \int_{\Scal^d} \exp(2u^\top v) d\mu(v).
$$
Then, we can write the second term as 
\begin{align*}
    & \E_{\rvq \sim Q} \left [ \log \E_{\rvc \sim C \circ \phi^{-1}} \left [ \exp{(2 \phi(\rvc)^\top \phi(\rvq))} \right ] \right ] \\
    & = \E_{\rvq \sim Q} \left [ \log U_{C \circ \phi^{-1}} (\phi(\rvq)) \right ],
\end{align*}

\noindent where $C$ is the distribution of prototypes of $S$, \ie each data point in $C$ is the mean of all the points in $S$ that share the same label, and $C \circ \phi^{-1}$ is the probability measure of prototypes, \ie the pushforward measure of $C$ via $\phi$.

We now consider the following problem:
\begin{align}
    \min_{\mu \in \Mcal(\Scal^d)} \int_{\Scal^d} \log U_\mu(u) d\mu(u).
    \label{eq:wang_isola}
\end{align}
The unique minimizer of Eq.~\ref{eq:wang_isola} is the \emph{uniform distribution on $\Scal^d$}, as shown in~\cite{pmlr-v119-wang20k}. 
This means that learning with $\Lcal_{proto}$ leads to prototypes uniformly distributed in the embedding space. By considering $\rmW^*$ the matrix of the optimal prototypes for each task then $\rmW^*$ is \emph{well-conditioned}, \ie $\kappa(\rmW^*) = O(1)$.

\end{proof}

\subsubsection{Proof of Proposition 1}



\begin{proof}
We follow \cite{sha_maml_2021} and note that in the considered setup the gradient of the loss for each task is given by 
$$\frac{\partial \ell_t(\widehat{\rvw} - \alpha \nabla \ell_t(\boldsymbol{\theta}))}{\partial \widehat{\rvw}} \propto (1-\alpha)^2 (\widehat{\rvw}_t - \boldsymbol{\theta}_{t})$$
so that the meta-training update for a single gradient step becomes:
$$\widehat{\rvw}_{t} \leftarrow  \widehat{\rvw}_{t-1} - \beta (1-\alpha)^2 (\widehat{\rvw}_{t-1} - \boldsymbol{\theta}_{t}),$$
where $\beta$ is the meta-training update learning rate. 
Starting at $\widehat{\rvw}_0 = \bm{0}_d$, we have that 
\begin{align*}
    &\widehat{\rvw}_{1} = c\boldsymbol{\theta}_{1},\\
    &\widehat{\rvw}_{2} = c((c-1)\boldsymbol{\theta}_{1} +\boldsymbol{\theta}_{2}),\\
    & \dots\\
    &\widehat{\rvw}_n = c\sum_{i=1}^n \boldsymbol{\theta}_i(c-1)^{n-i},
\end{align*}
where $c :=\beta (1-\alpha)^2$. We can now define matrices $\widehat{\rmW}_2^{i}$ as follows:
\begin{align*}
    &\widehat{\rmW}_2^{1} = \begin{pmatrix}
                            c\boldsymbol{\theta}_{1},\\
                            c((c-1)\boldsymbol{\theta}_{1} +\boldsymbol{\theta}_{2})
                            \end{pmatrix}, \\
    &\widehat{\rmW}_2^{2} = \begin{pmatrix}
                            c((c-1)\boldsymbol{\theta}_{1} +\boldsymbol{\theta}_{2}),\\
                            c((c-1)^2\boldsymbol{\theta}_{1}+(c-1)\boldsymbol{\theta}_{2}+\boldsymbol{\theta}_3)
                            \end{pmatrix},\\
    & \dots \\
    &\widehat{\rmW}_2^{n} = \begin{pmatrix}
                            c\sum_{i=1}^{n} \boldsymbol{\theta}_i(c-1)^{n-i},\\
                            c\sum_{i=1}^{n+1} \boldsymbol{\theta}_i(c-1)^{n-i}
                            \end{pmatrix}.       
\end{align*}
We can note that for all $i>1$:
$$\widehat{\rmW}_2^{i+1} = (c-1)\widehat{\rmW}_2^{i} + c \boldsymbol{\Theta}_i.$$
Now, we can write:
\begin{align*}
    \kappa(\widehat{\rmW}_2^{i+1}) = \frac{\sigma_1(\widehat{\rmW}_2^{i+1})}{\sigma_2(\widehat{\rmW}_2^{i+1})} = \frac{\sigma_1((c-1)\widehat{\rmW}_2^{i} + c\boldsymbol{\Theta}_i)}{\sigma_2((c-1)\widehat{\rmW}_2^{i} + c\boldsymbol{\Theta}_i)}\\
    \geq \frac{\sigma_1((c-1)\widehat{\rmW}_2^{i}) - \sigma_2(c\boldsymbol{\Theta}_i)}{\sigma_2((c-1)\widehat{\rmW}_2^{i} + c\boldsymbol{\Theta}_i)} \\
    \geq \frac{\sigma_1((c-1)\widehat{\rmW}_2^{i}) - \sigma_2(c\boldsymbol{\Theta}_i)}{\sigma_2((c-1)\widehat{\rmW}_2^{i}) + \sigma_2(c\boldsymbol{\Theta}_i)}\\
    \geq \kappa(\widehat{\rmW}_2^{i}).
\end{align*}
where the second and third lines follow from the inequalities for singular values $\sigma_1(A+B) \leq \sigma_1(A) + \sigma_2(B)$ and $\sigma_i(A+B) \geq \sigma_i(A) - \sigma_{\text{min}}(B)$ and the desired result is obtained by setting $\sigma_\text{min}(\boldsymbol{\theta}_i) = 0.$
\end{proof}

\subsubsection{Proof of Proposition 2}



\begin{proof}
Du et al. \cite{duFewShotLearningLearning2020} assume that $\sigma_k(\rmW^*) \gtrsim \frac{T}{k}$ (Assumption 4.3 in their work). However, since we also have $\|\rvw_t^*\| = O(1)$, it is equivalent to $\frac{\sigma_1(\rmW^*)}{\sigma_k(\rmW^*)} = O(1)$. \\
We have $\sigma_1(\rmW^*) \gtrsim \sigma_k(\rmW^*) \gtrsim \frac{T}{k}$ and then $\frac{\sigma_1(\rmW^*)}{T \cdot \sigma_k(\rmW^*)} = \frac{1}{T} \cdot \kappa(\rmW^*) \gtrsim \frac{1}{k \cdot \sigma_k(\rmW^*)}$ which we use in their proof of Theorem 5.1 instead of $\frac{1}{T} \gtrsim \frac{1}{k \cdot \sigma_k(\rmW^*)}$ to obtain the desired result.
\end{proof}

\subsubsection{Proof of Proposition 3}



\begin{figure}[!t]
    \centering
    \includegraphics[width = .85\linewidth]{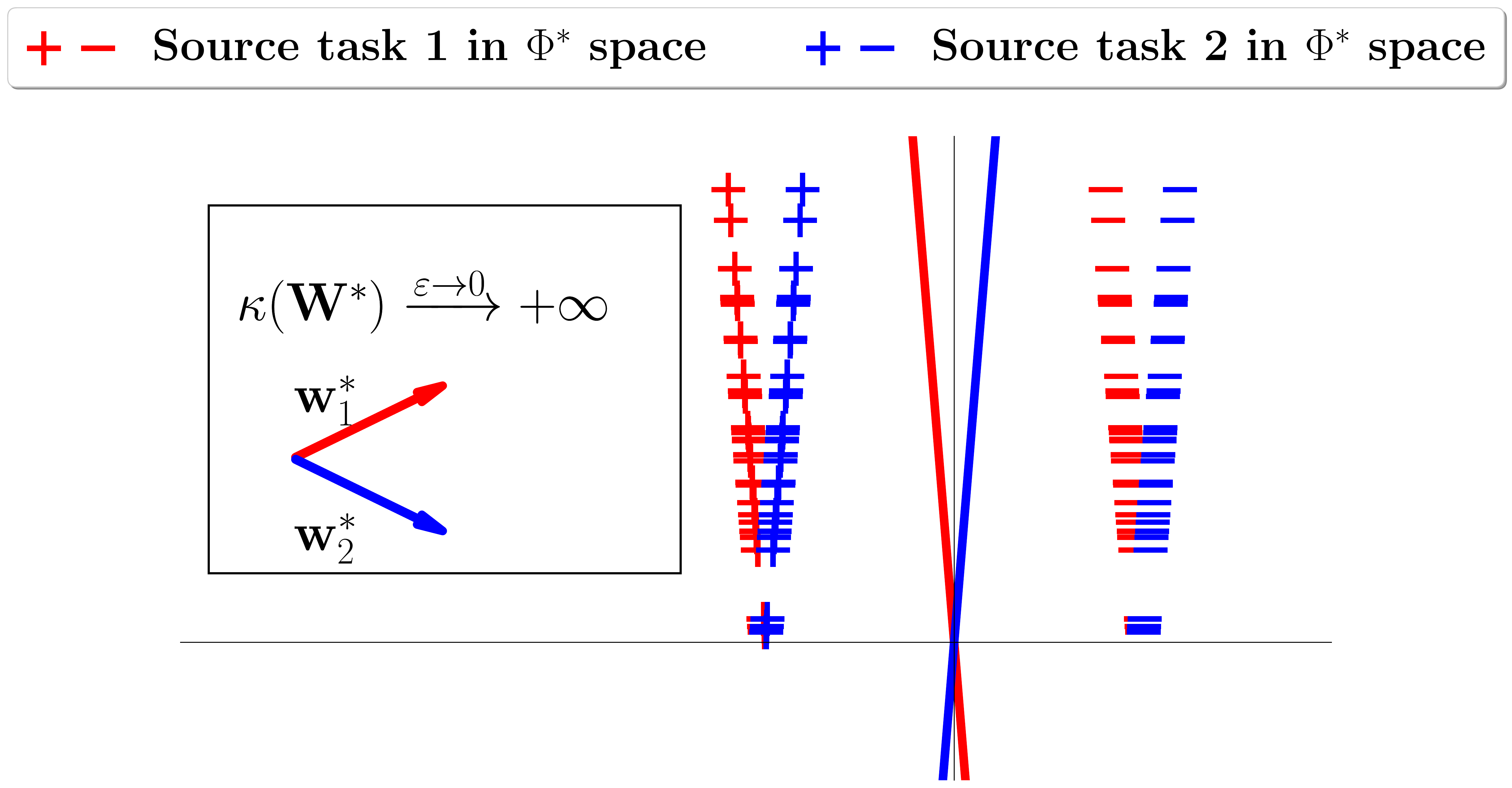}
    \includegraphics[width = .85\linewidth]{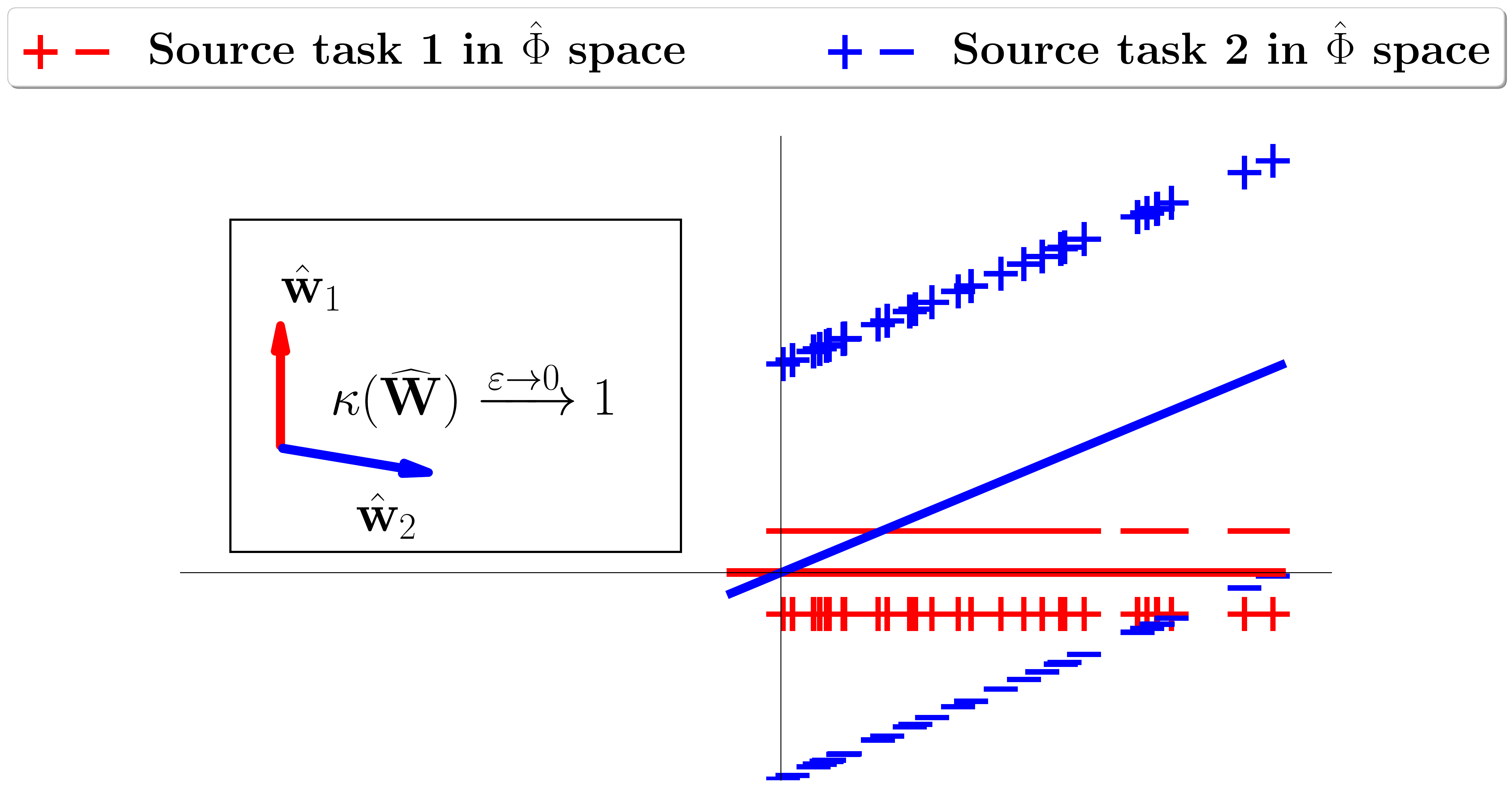}
    \caption{Visualization of the distributions used in the constructive example for the proof of Proposition \ref{prop:example}, with $\epsilon = 0.02$. In this example, $\kappa(\widehat{\rmW})$ is closer to 1 than $\kappa(\rmW^*)$. It shows that we can search for a representation $\widehat{\phi}$ such that optimal predictors in this space are fulfilling the assumptions, while solving the underlying problem equally well.}
    \label{fig:intuition}
\end{figure}

\begin{proof}

Let us define two uniform distributions $\mu_1$ and $\mu_2$ parametrized by a scalar $\varepsilon>0$ satisfying the data generating process from Eq.~\ref{eq:data_gen_model}:
\begin{enumerate}
    \item $\mu_1$ is uniform over $\{1-k\varepsilon,k,1, \underbrace{\dots}_{d-3}\}\times\{1\} \cup \{1+k\varepsilon,k,-1, \underbrace{\dots}_{d-3}\}\times\{-1\}$;
    \item $\mu_2$ is uniform over $\{1+k\varepsilon,k,\frac{k-1}{\varepsilon},  \underbrace{\dots}_{d-3}\}\times\{1\} \cup \{-1+k\varepsilon,k,\frac{1+k}{\varepsilon}, \underbrace{\dots}_{d-3}\}\times\{-1\}$.
\end{enumerate}
where last $d-3$ coordinates of the generated instances are arbitrary numbers.
We now define the optimal representation and two optimal predictors for each distribution as the solution to the MTR problem over the two data generating distributions and $\Phi = \{\phi |\ \phi(\rvx) = \boldsymbol{\Phi}^T\rvx,\ \boldsymbol{\Phi}\in \sR^{d\times 2}\}$:
\begin{align}
    \phi^*, \rmW^* = \argmin_{\phi \in \Phi, \rmW \in \sR^{2\times 2}}\sum_{i=1}^2\underset{(\rvx,y)\sim\mu_i}{\E} \ell(y,\langle \rvw_i, \phi(\rvx)\rangle),
    \label{eq:meta_train_true}
\end{align}
One solution to this problem can be given as follows:
\[\boldsymbol{\Phi}^* = \begin{pmatrix} 1 & 0 & 0 & \dots & 0\\0 & 1 & 0 & \dots & 0\end{pmatrix}^T, \quad \rmW^* = \begin{pmatrix} 1 & \varepsilon\\1 & -\varepsilon\end{pmatrix},
\]
where $\boldsymbol{\Phi}^*$ projects the data generated by $\mu_i$ to a two-dimensional space by discarding its $d-2$ last dimensions and the linear predictors satisfy the data generating process from Eq.~\ref{eq:data_gen_model} with $\varepsilon=0$. One can verify that in this case $\rmW^*$ have singular values equal to $\sqrt{2}$ and $\sqrt{2}\varepsilon$, and $\kappa(\rmW^*) = \frac{1}{\varepsilon}$. When $\varepsilon \rightarrow 0$, the optimal predictors make the ratio arbitrary large thus violating Assumption 1.

Let us now consider a different problem where we want to solve Eq. \ref{eq:meta_train_true} with constraints that force linear predictors to satisfy both assumptions:
\begin{align}
    \begin{split}
    \widehat{\phi}, \widehat{\rmW} &= \argmin_{\phi \in \Phi, \rmW \in \sR^{2\times 2}}\sum_{i=1}^2\underset{(\rvx,y)\sim\mu_i}{\E} \ell(y,\langle \rvw_i, \phi(\rvx)\rangle), \\
    &\quad \text{s.t.}\ \kappa(\rmW) \approx 1 \quad \text{ and } \quad \forall i, \quad \|\rvw_i\| \approx 1.
    \end{split}
    \label{eq:meta_train_hat}
\end{align}
Its solution is different and is given by  
\[\widehat{\boldsymbol{\Phi}} = \begin{pmatrix} 0 & 1 & 0 & \dots & 0\\0 & 0 & 1 & \dots & 0\end{pmatrix}^T, \quad \widehat{\rmW} = \begin{pmatrix} 0 & 1 \\1 & -\varepsilon\end{pmatrix}.
\]
Similarly to $\boldsymbol{\Phi}^*$, $\widehat{\boldsymbol{\Phi}}$ projects to a two-dimensional space by discarding the first and last $d-3$ dimensions of the data generated by $\mu_i$. The learned predictors in this case also satisfy Eq. \ref{eq:data_gen_model} with $\varepsilon=0$, but contrary to $\rmW^*$, $\kappa(\widehat{\rmW}) = \sqrt{ \frac{2 + \varepsilon^2 + \varepsilon \sqrt{\varepsilon^2 + 4}}{2 + \varepsilon^2 - \varepsilon \sqrt{ \varepsilon^2 + 4 }} }$ tends to 1 when $\varepsilon \rightarrow 0$.
The construction used in this proof is illustrated in Figure \ref{fig:intuition}

\end{proof}

\subsection{More Details on the regularization terms}\label{ax:details_reg}

By adding $\|\rmW\|_F^2$ in the loss, we force the model to have a low norm on the weights. Since it cannot be put to 0 or below, the model will keep the norm relatively constant instead of increasing it. The second regularizer term is a softer way to apply the constraint on the norm rather than considering normalized weights as in Eq.~\ref{eq:meta_train_hat}.

According to Theorem 7.1 from \cite{lewisNonsmoothAnalysisSingular2005}, subgradients of singular values function are well-defined for absolutely symmetric functions. In our case, we are computing in practice the squared singular values $\sigma^2(\rmW)$ and we retrieve the singular values by taking the square root, as explained in Section 3.2 of the paper. This means that effectively, we are computing $\kappa(\rmW) = \max(|\sigma(\rmW)|) / \min(|\sigma(\rmW)|)$, which is an absolutely symmetric function. Consequently, subgradients of the spectral regularization term $\kappa(\rmW)$ are well-defined and can be optimized efficiently when used in the objective function.

\subsection{Detailed Experimental Setup}\label{ax:expe_setup}
We consider the few-shot image classification problem on three benchmark datasets, namely:
\begin{enumerate}
    \item \textbf{Omniglot} \cite{lakeHumanlevelConceptLearning2015} is a dataset of 20 instances of 1623 characters from 50 different alphabets. Each image was hand-drawn by different people. The images are resized to $28 \times 28$ pixels and the classes are augmented with rotations by multiples of $90$ degrees.
    \item \textbf{miniImageNet} \cite{raviOPTIMIZATIONMODELFEWSHOT2017} is a dataset made from randomly chosen classes and images taken from the ILSVRC-12 dataset \cite{russakovskyImageNetLargeScale2015}. The dataset consists of $100$ classes and $600$ images for each class. The images are resized to $84 \times 84$ pixels and normalized.
    \item \textbf{tieredImageNet} \cite{renMetaLearningSemiSupervisedFewShot2018} is also a subset of ILSVRC-12 dataset. However, unlike miniImageNet, training classes are semantically unrelated to testing classes. The dataset consists of $779,165$ images divided into $608$ classes. Here again, the images are resized to $84 \times 84$ pixels and normalized.
\end{enumerate}
For each dataset, we follow a common experimental protocol used in \cite{finnModelAgnosticMetaLearningFast2017,chenCLOSERLOOKFEWSHOT2019} and use a four-layer convolution backbone (Conv-4) with 64 filters as done by \cite{chenCLOSERLOOKFEWSHOT2019} optimized with Adam~\cite{kingma2015adam} and a learning rate of $0.001$. On \emph{miniImageNet} and \emph{tieredImageNet}, models are trained on $60000$ 5-way 1-shot or 5-shot episodes and on $30000$ 20-way 1-shot or 5-shot episodes for \emph{Omniglot}. We use a batch size of $4$ and evaluate on the validation set every $1000$ episodes. We keep the best performing model on the validation set to evaluate on the test set. We measure the performance using the top-1 accuracy with $95\%$ confidence intervals, reproduce the experiments with 4 different random seeds using a single NVIDIA V100 GPU, and average the results over $2400$ test tasks. The seeds used for all experiments are $1$, $10$, $100$ and $1000$. For \Maml\ and \MC, we use an inner learning rate of $0.01$ for \emph{miniImageNet} and \emph{tieredImageNet}, and $0.1$ for \emph{Omniglot}. During training, we perform $5$ inner gradient step and $10$ step during testing. For all FSC experiments, unless explicitly stated, we use the regularization parameters $\lambda_1 = \lambda_2 = 1$.

We also provide experiments with the ResNet-12 architecture \cite{leeMetaLearningDifferentiableConvex2019}. In this case, we follow the recent practice and initialize the models with the weights pretrained on the entire meta-training set \cite{ye2020fewshot, rusu2018meta, qiao2018few}. Like in their protocol, this initialization is updated by meta-training with \Proto\ or \Maml\ on at most $20 000$ episodes, grouping every $100$ episodes into an \emph{epoch}. Then, the best performing model on the validation set, evaluated every epoch, is kept and the performance on $10 000$ test tasks is measured. For all experiments with the ResNet-12 architecture, the SGD optimizer with a weight decay of $0.0005$ and momentum of $0.9$ and a batch of episodes of size $1$ are used.
For \Proto, following the protocol of Ye et al. \cite{ye2020fewshot}, an initial learning rate of $0.0002$, decayed by a factor $0.5$ every $40$ epochs, is used. For \Maml, following Ye et Chao \cite{yeHowTrainYour2021a}, the initial learning rate is set to $0.001$, decayed by a factor $0.1$ every 20 epochs. The number of inner loop updates are respectively set to 15 and 20 with a step size of $0.05$ and $0.1$ for 1-shot and 5-shot episodes on the miniImageNet dataset, and respectively 20 and 15 with a step size of 0.001 and 0.05 on the tieredImageNet dataset.

\subsection{Detailed performance comparisons} \label{ax:performancecomparisons}

\begin{figure*}[!ht]
    \includegraphics[width=0.30\textwidth]{images/proto_norm_cond_1shot.pdf}
    \includegraphics[width=0.30\textwidth]{images/proto_norm_norm_1shot.pdf}
    \includegraphics[width = .30\linewidth]{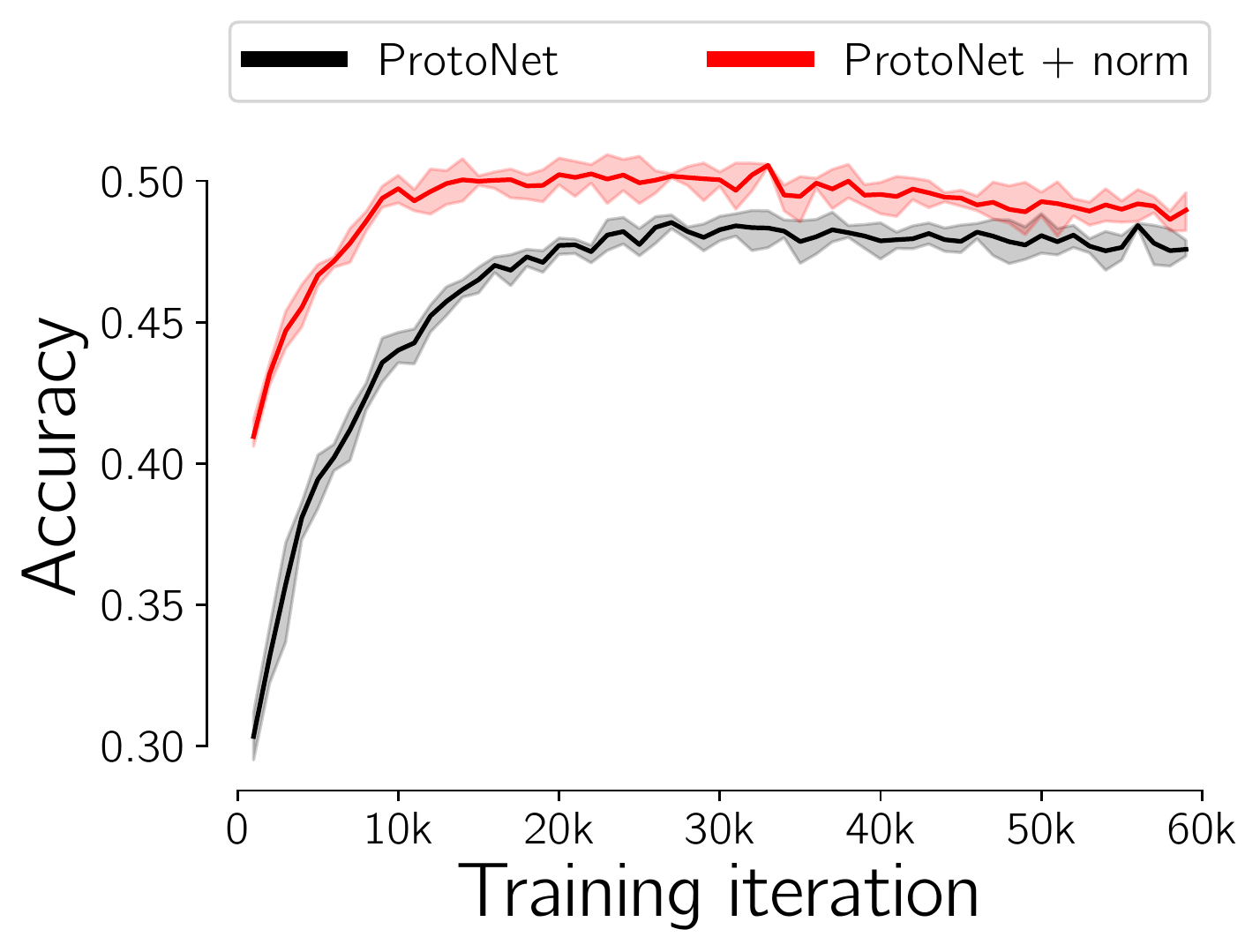}
    
    \includegraphics[width=0.30\linewidth]{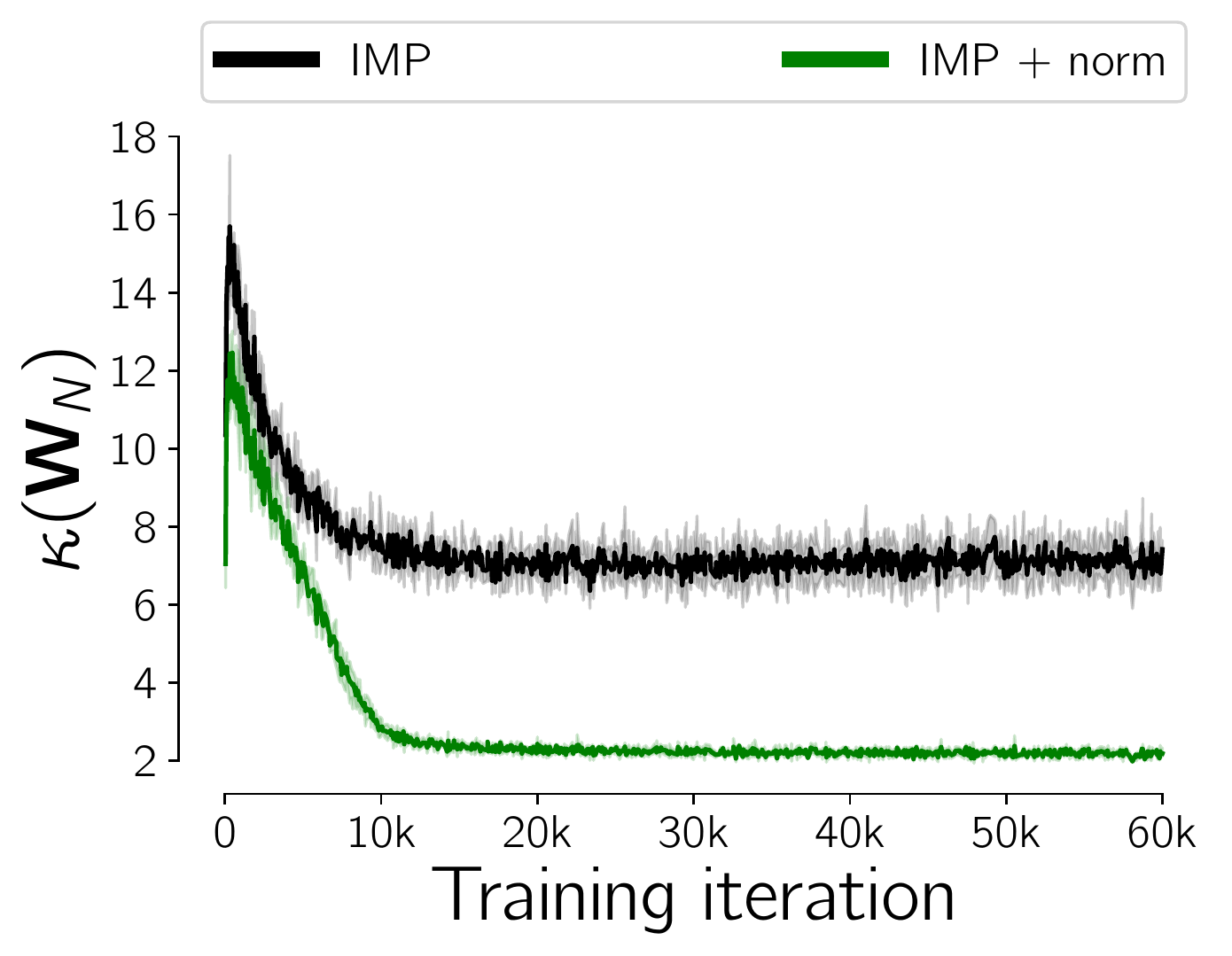}
    \includegraphics[width=0.30\linewidth]{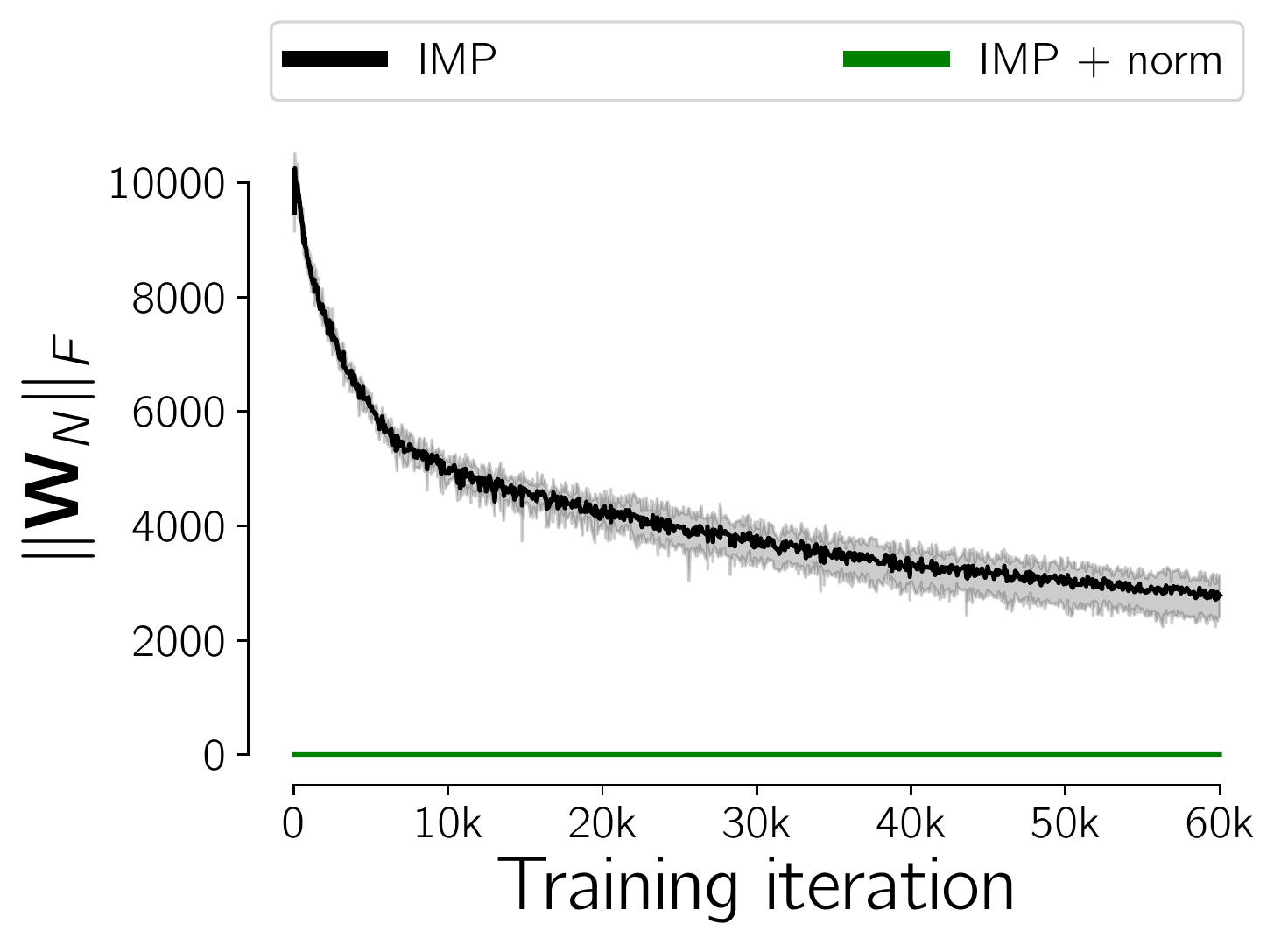}
    \includegraphics[width = .30\linewidth]{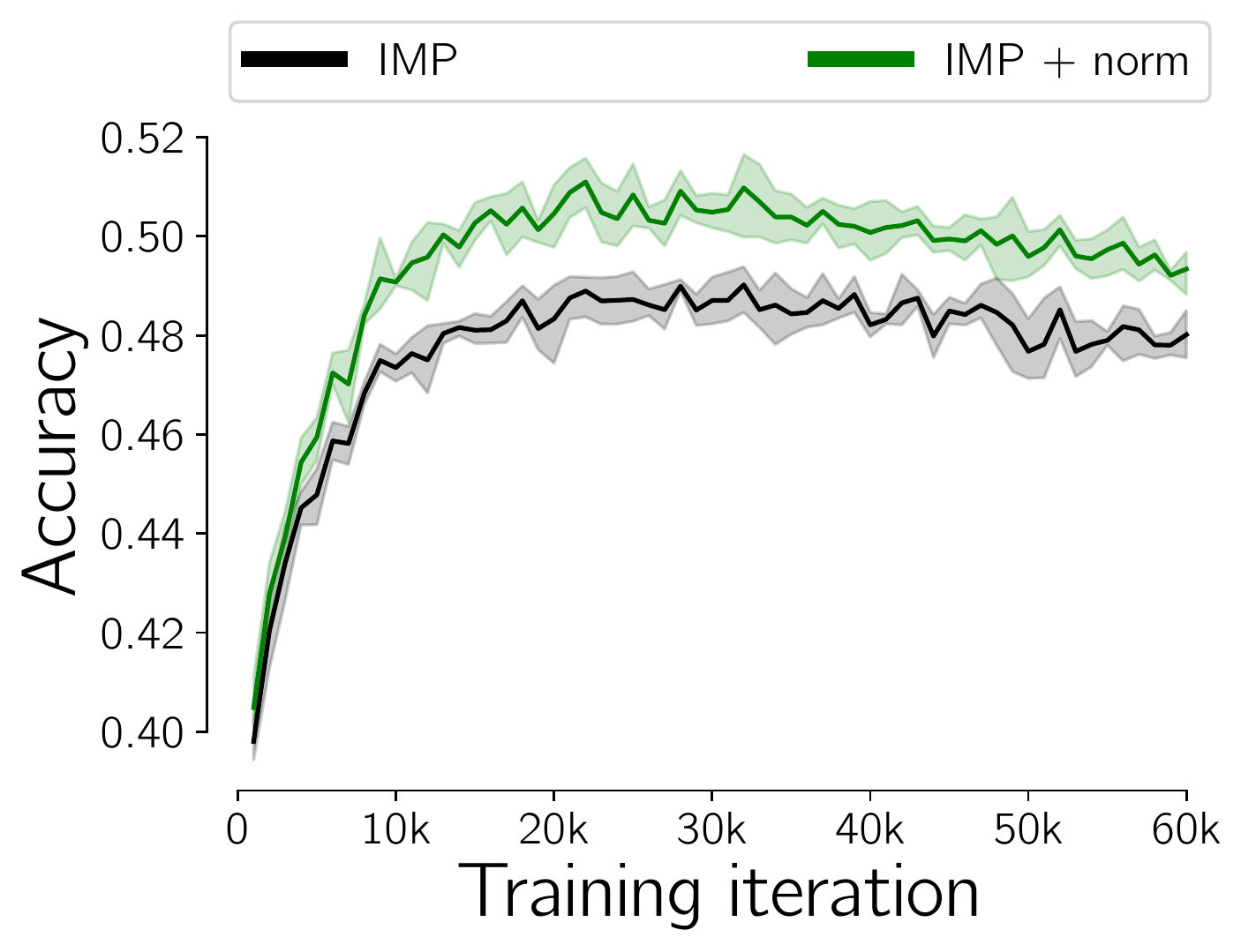} 
    
    \includegraphics[width=0.30\textwidth]{images/maml_reg_cond.pdf}
    \includegraphics[width=0.30\textwidth]{images/maml_reg_norm.pdf}
    \includegraphics[width = .30\linewidth]{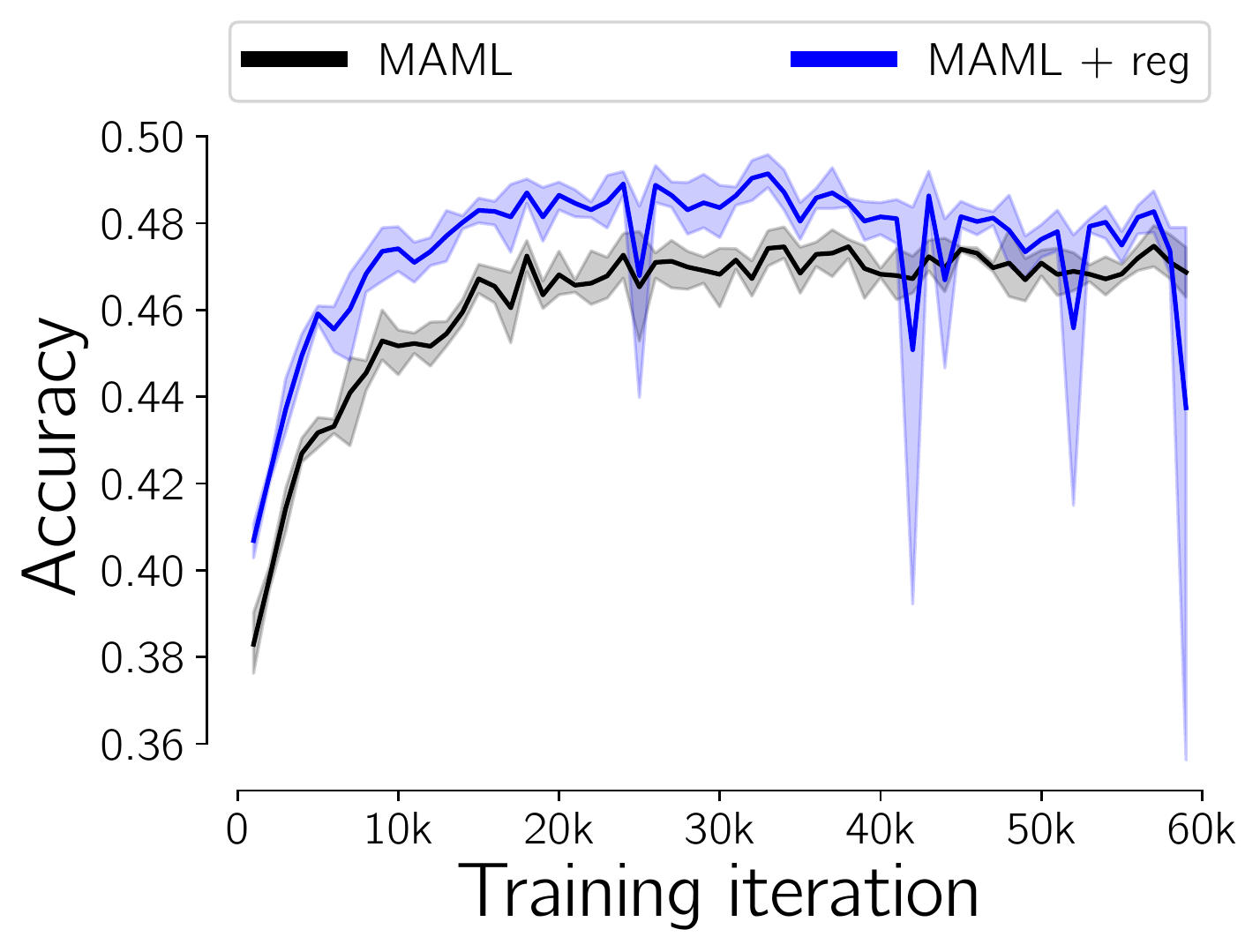}
    
    \includegraphics[width=0.30\linewidth]{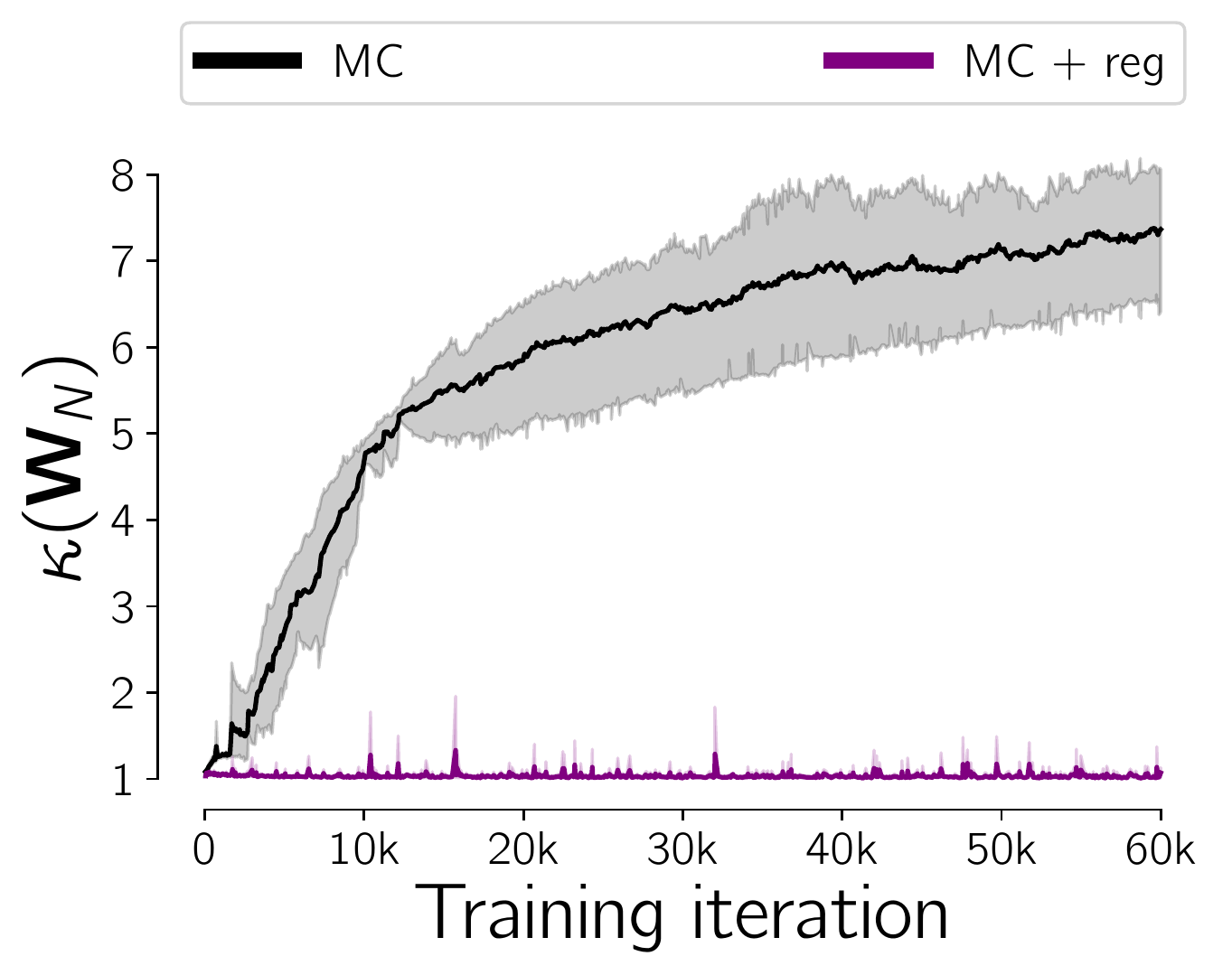}
    \includegraphics[width=0.30\linewidth]{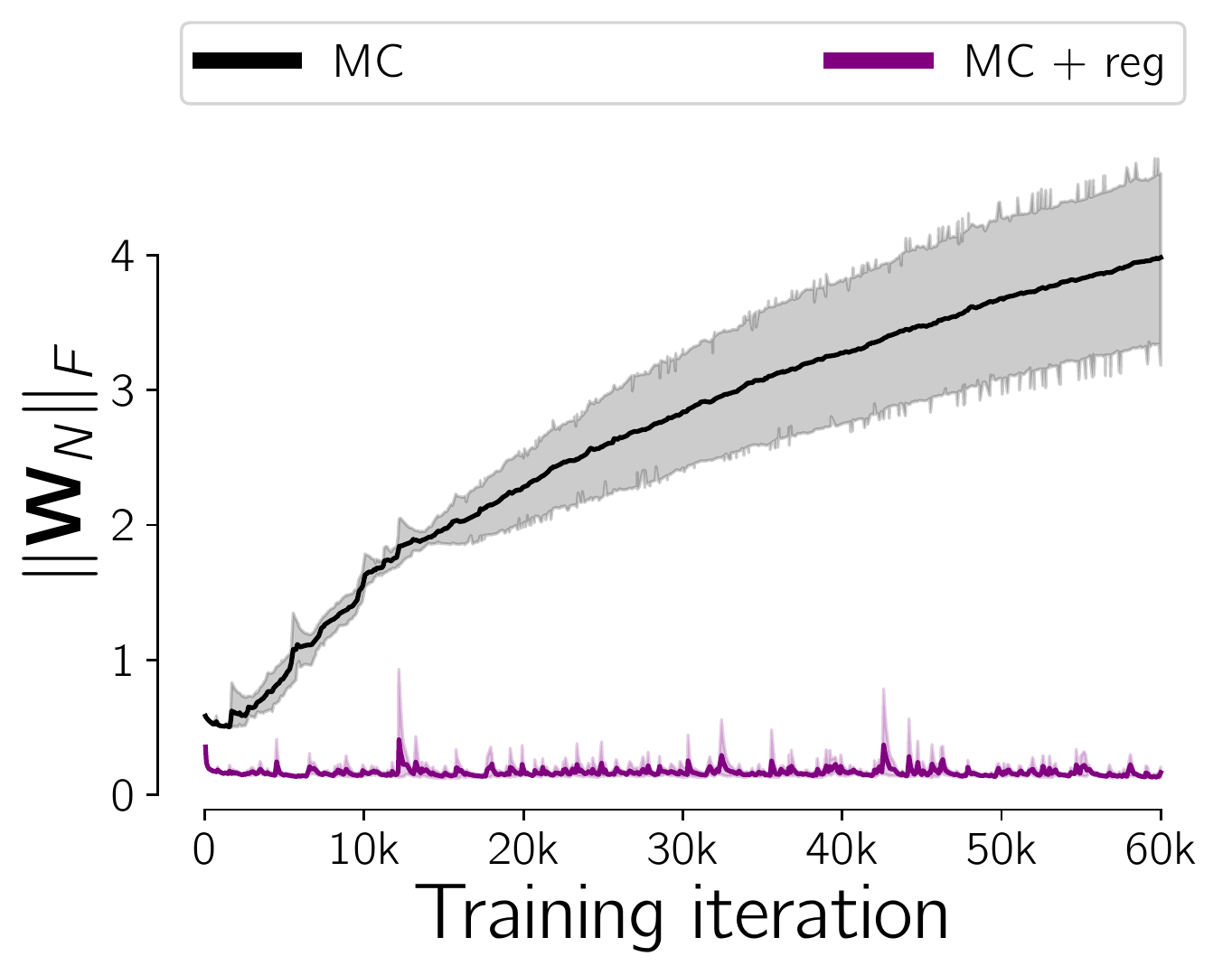}
    \includegraphics[width = .30\linewidth]{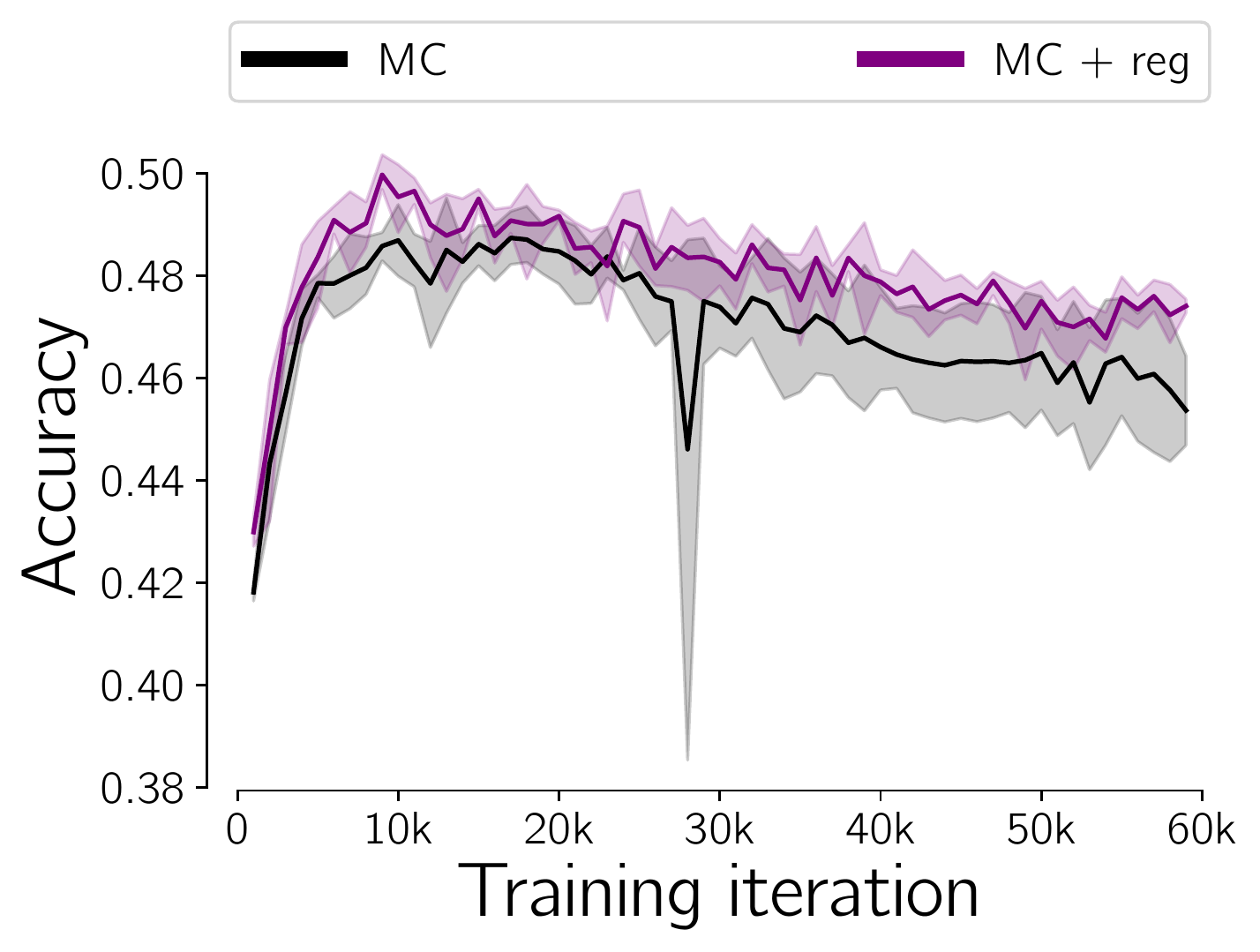}
   \captionof{figure}{
    Evolution of $\kappa(\rmW_N)$ (\emph{left}), $\|\rmW_N\|_F$ (\emph{middle}) and the accuracy (\emph{right}) on 5-way 1-shot episodes from \emph{miniImageNet},
    for \Proto, \IMP, \Maml, \MC\ (\emph{from top to bottom respectively.}) and their regularized or normalized counterparts. All results are averaged over 4 different random seeds. The shaded areas show $95\%$ confidence intervals.
    }
    \label{fig:acc_curves}
\end{figure*}

\begin{figure*}[!htb]
    \centering
    \includegraphics[width=0.32\linewidth]{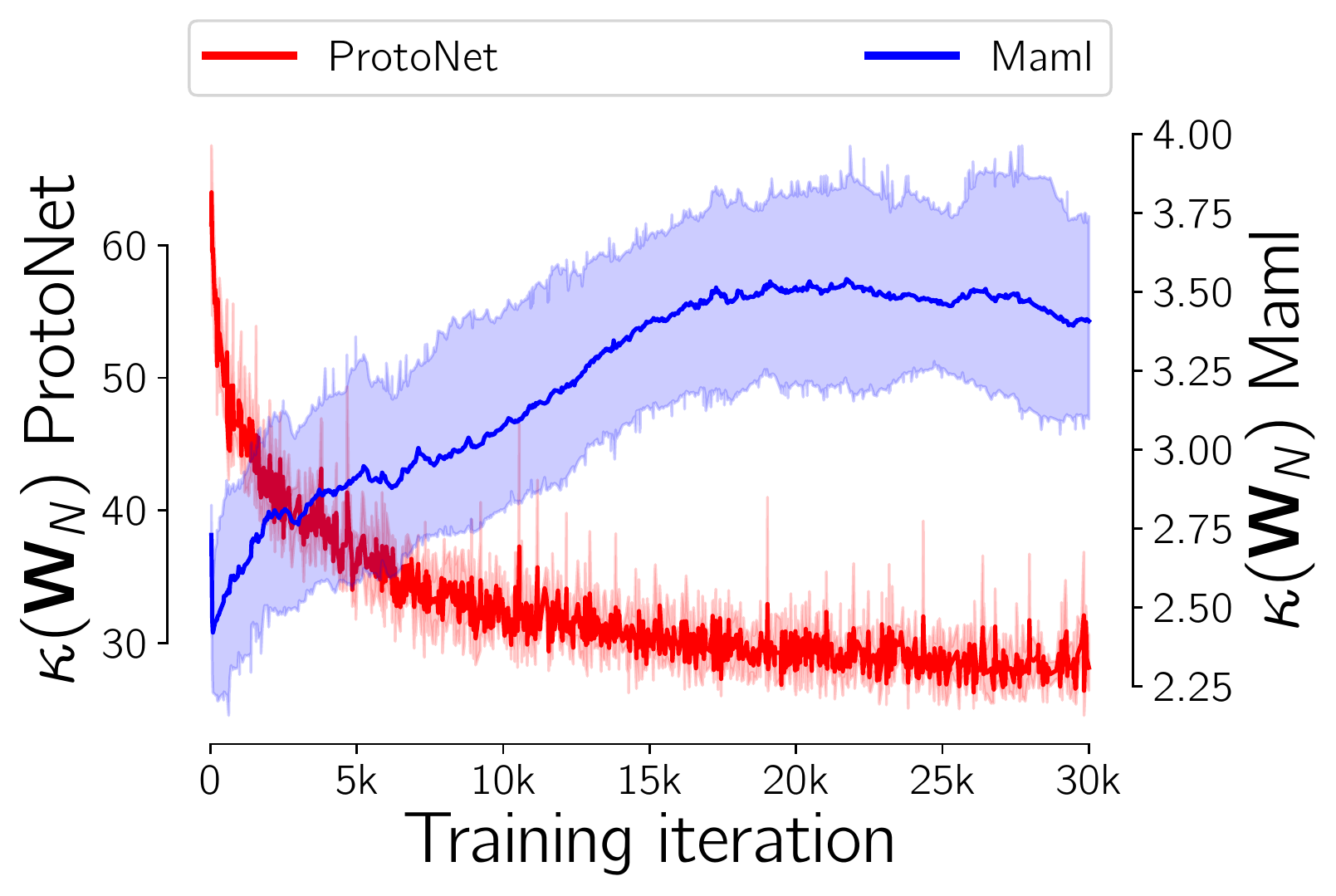}
    \hspace{0.9em}
    \includegraphics[width=0.32\linewidth]{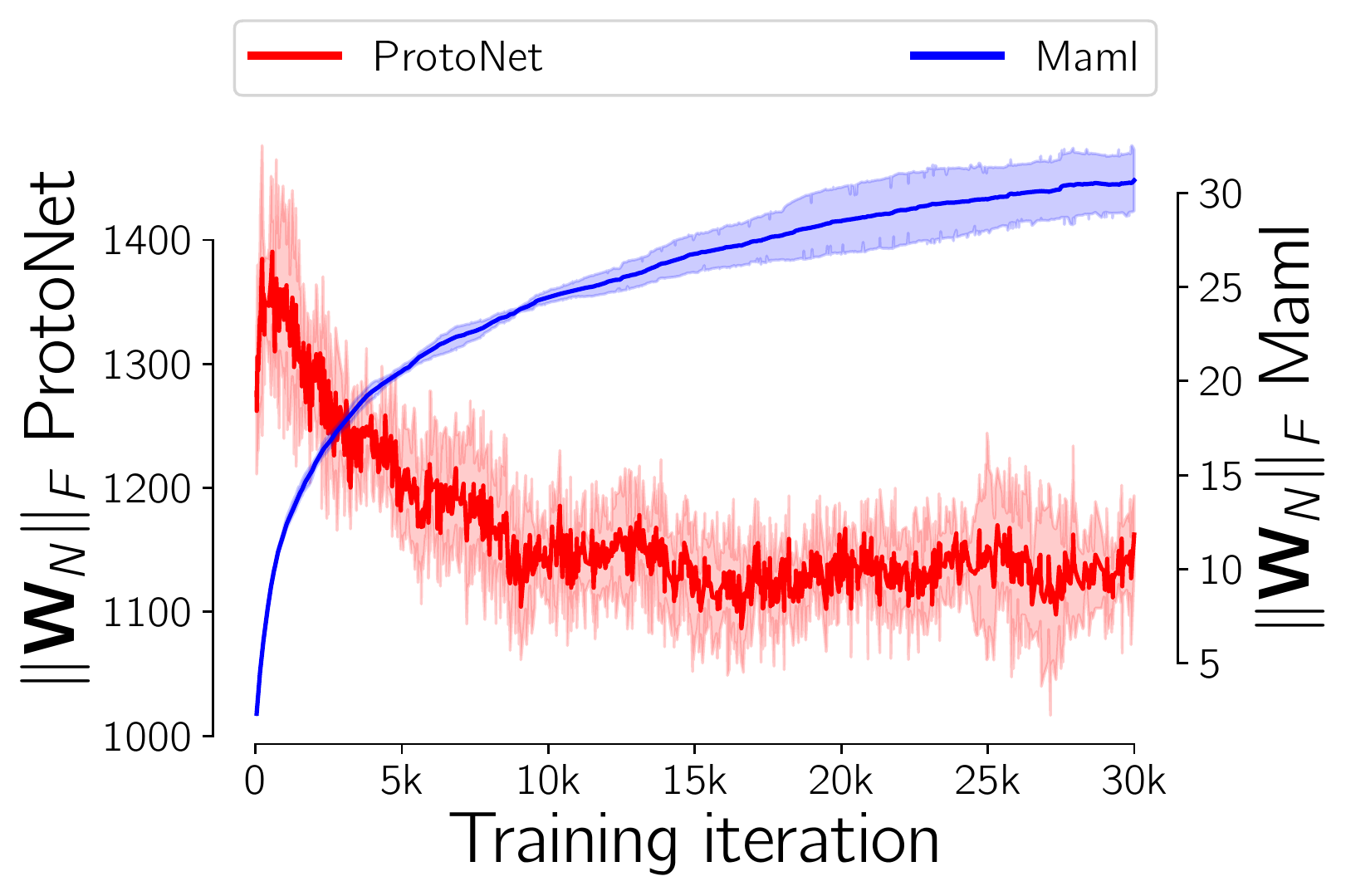}
    \hspace{0.9em}
    \includegraphics[width=0.28\linewidth]{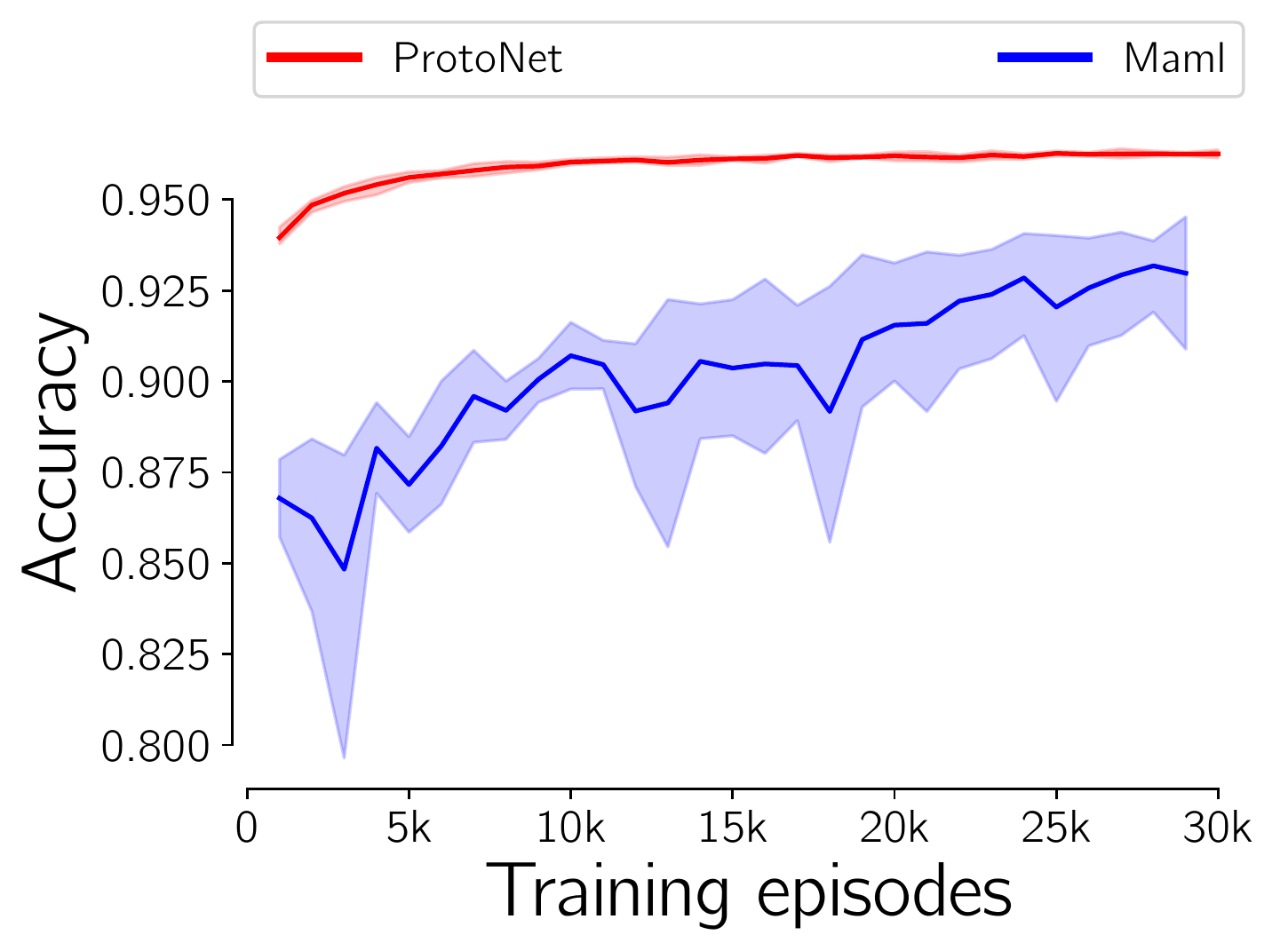}
    \caption{Evolution of $\kappa(\rmW_N)$, $\|\rmW_N\|_F$ and $\kappa(\rmW)$ (\emph{in log scale}) during the training of \Proto\ (\emph{red, left axes}) and \Maml\ (\emph{blue, right axes}) on Omniglot with 5-way 1-shot episodes. 
    }
    \label{fig:training_curves}
\end{figure*}

\begin{figure}
    \centering
    \includegraphics[width=0.7\linewidth]{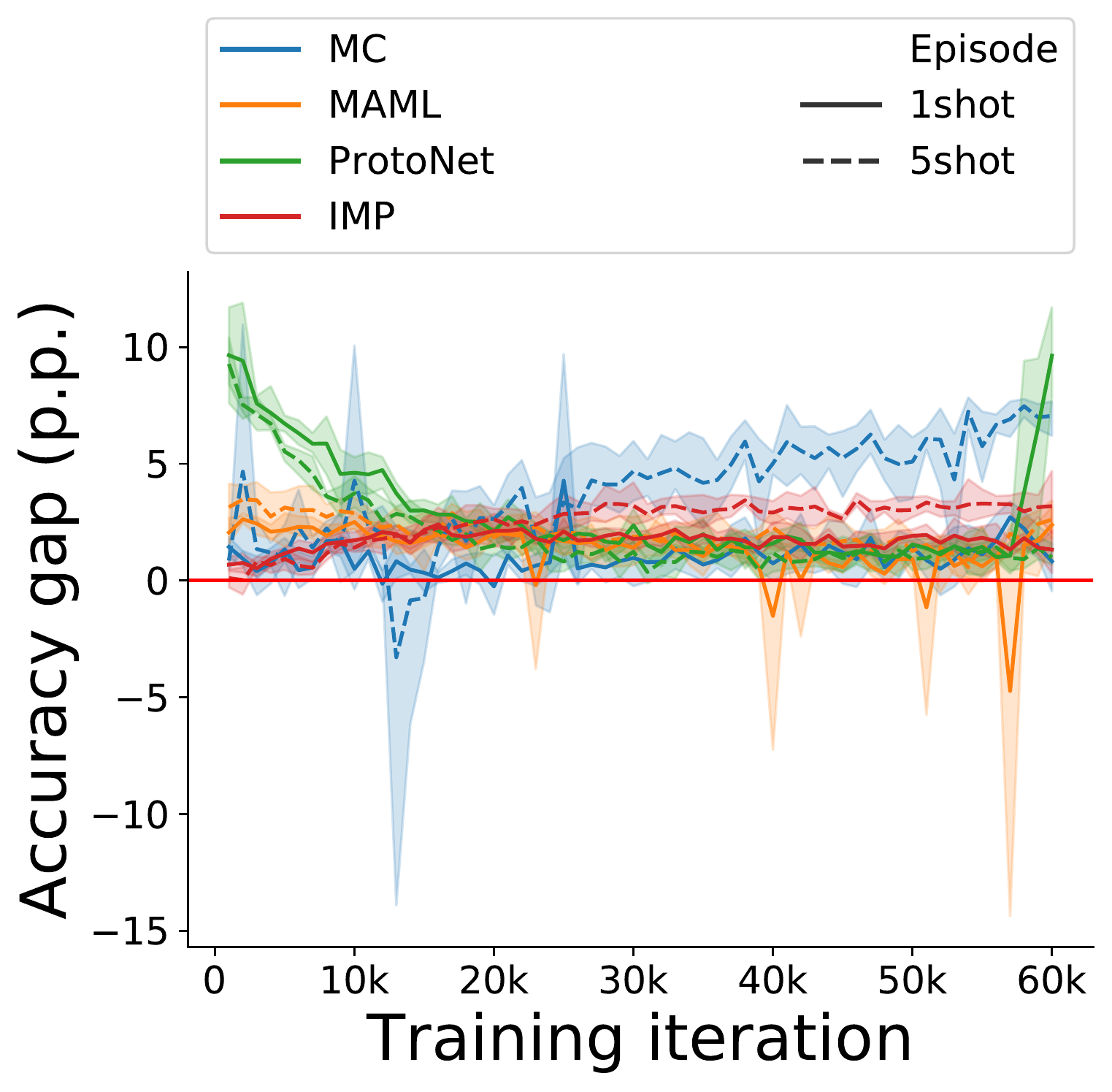}
    \caption{Performance gap (in p.p.) when applying regularization for gradient-based and normalization for metric-based methods throughout the training process on 5-way 1-shot and 5-shot episodes on miniImageNet (\emph{better viewed in color}). Each data point is averaged over 2400 validation episodes and 4 different seeds and shaded areas report $95\%$ confidence interval. We can see that the gap is globally positive throughout training and generally higher at the beginning of training. The increase in the gap at the end of training is linked to a lower overfitting.}
    \label{fig:gap_curves}
\end{figure}

The plots showing the behavior of \Proto\ and \Maml\ on Omniglot are shown in Figure~\ref{fig:training_curves}. The detailed training curves of the regularized and normalized versions of \Proto, \IMP, \Maml\ and \MC\ can be found in Figure~\ref{fig:acc_curves}. The performance gap (difference of accuracy in p.p.) throughout training for all methods is shown in Figure~\ref{fig:gap_curves}.
Table~\ref{tab:perf} provides the detailed performance of our reproduced methods with and without our regularization or normalization and Figure~\ref{fig:gap_curves} shows the performance gap throughout training for all methods on miniImageNet. 
From them, we note that the gap in performance due to our regularization is globally positive throughout the whole training, which shows the increased generalization capabilities from enforcing the assumptions. There is also generally a high gap at the beginning of training suggesting faster learning.
The best performance with the proposed regularization is achieved after training on a significantly reduced amount of training data. 
These results are also summarized in Table~1 of our paper and discussions about them can be found in Section~4.2 and 4.3.

\begin{table}[ht]
    \caption{Performance of several meta-learning algorithms without and with our regularization (or normalization in the case of \Proto\ and \IMP) to enforce the theoretical assumptions. All accuracy results (in \%) are averaged over 2400 test episodes and 4 different seeds and are reported with $95\%$ confidence interval. Episodes are 20-way classification for Omniglot and 5-way classification for miniImageNet and tieredImageNet.}
    \label{tab:perf}
    \begin{center}
    \begin{tabular}{@{}lllll@{}} \toprule
    Method & Dataset & Episodes & without Reg./Norm. & with Reg./Norm. \\
    \midrule
    
    \multirow{7}{*}{\Proto} & \multirow{2}{*}{Omniglot} & 1-shot & $95.56 \pm 0.10\%$ & $\mathbf{95.89 \pm 0.10\%}$ \\
     & & 5-shot & $\mathbf{98.80 \pm 0.04\%}$ & $\mathbf{98.80 \pm 0.04\%}$ \\
    \cmidrule{2-5}
     & \multirow{2}{*}{miniImageNet} & 1-shot & $49.53 \pm 0.41\%$ & $\mathbf{50.29 \pm 0.41\%}$ \\
     & & 5-shot & $65.10 \pm 0.35\%$ & $\mathbf{67.13 \pm 0.34\%}$ \\
    \cmidrule{2-5}
     & \multirow{2}{*}{tieredImageNet} & 1-shot & $51.95 \pm 0.45\%$ & $\mathbf{54.05 \pm 0.45\%}$ \\
     & & 5-shot & $\mathbf{71.61 \pm 0.38\%}$ & $\mathbf{71.84 \pm 0.38\%}$ \\
     
    \midrule
    
    \multirow{7}{*}{\IMP} & \multirow{2}{*}{Omniglot} & 1-shot & $\mathbf{95.77 \pm 0.20\%}$ & $\mathbf{95.85 \pm 0.20\%}$ \\
     & & 5-shot & $\mathbf{98.77 \pm 0.08\%}$ & $\mathbf{98.83 \pm 0.07\%}$ \\
    \cmidrule{2-5}
     & \multirow{2}{*}{miniImageNet} & 1-shot & $48.85 \pm 0.81\%$ & $\mathbf{50.69 \pm 0.80\%}$ \\
     & & 5-shot & $66.43 \pm 0.71\%$ & $\mathbf{67.29 \pm 0.68\%}$ \\
    \cmidrule{2-5}
     & \multirow{2}{*}{tieredImageNet} & 1-shot & $52.16 \pm 0.89\%$ & $\mathbf{53.46 \pm 0.89\%}$ \\
     & & 5-shot & $\mathbf{71.79 \pm 0.75\%}$ & $\mathbf{72.38 \pm 0.75\%}$ \\
     
    \midrule
    
    \multirow{7}{*}{\Maml} & \multirow{2}{*}{Omniglot} & 1-shot & $91.72 \pm 0.29\%$ & $\mathbf{95.67 \pm 0.20 \%}$ \\
    &  & 5-shot & $97.07 \pm 0.14\%$ & $\mathbf{98.24 \pm 0.10\%}$ \\
    \cmidrule{2-5}
    & \multirow{2}{*}{miniImageNet} & 1-shot & $47.93 \pm 0.83\%$ & $\mathbf{49.16 \pm 0.85\%}$ \\
    & & 5-shot & $64.47 \pm 0.69\%$ & $\mathbf{66.43 \pm 0.69\%}$ \\
    \cmidrule{2-5}
    & \multirow{2}{*}{tieredImageNet} & 1-shot & $50.08 \pm 0.91\%$ & $\mathbf{51.5 \pm 0.90\%}$ \\
    & & 5-shot & $67.5 \pm 0.79\%$ & $\mathbf{70.16 \pm 0.76\%}$ \\
    
    \midrule
    
    \multirow{7}{*}{\MC} & \multirow{2}{*}{Omniglot} & 1-shot & $\mathbf{96.56 \pm 0.18\%}$ & $95.95 \pm  0.20\%$ \\
    &  & 5-shot & $\mathbf{98.88 \pm 0.08\%}$ & $98.78 \pm 0.08\%$ \\
    \cmidrule{2-5}
    & \multirow{2}{*}{miniImageNet} & 1-shot & $\mathbf{49.28 \pm 0.83\%}$ & $\mathbf{49.64 \pm 0.83\%}$ \\
    & & 5-shot & $63.74 \pm 0.69\%$ & $\mathbf{65.67 \pm 0.70\%}$ \\
    \cmidrule{2-5}
    & \multirow{2}{*}{tieredImageNet} & 1-shot & $55.16 \pm 0.94\%$ & $\mathbf{55.85 \pm 0.94\%}$ \\
    & & 5-shot & $71.95 \pm 0.77\%$ & $\mathbf{73.34 \pm 0.76\%}$ \\

    
     
    
    
    \bottomrule
    \end{tabular}
    \end{center}
\end{table}

\subsection{Further enforcing a low condition number on Metric-based methods}\label{ax:entropy}

To guide the model into learning an encoder with the lowest condition number, we consider adding $\kappa(\rmW_N)$ as a regularization term when training a normalized \Proto. In addition to the normalization of the prototypes, this should further enforce the assumption on the condition number.
Unfortunately, this latter strategy hinders the convergence of the network and leads to numerical instabilities. It is most likely explained by prototypes being computed from image features which suffer from rapid changes across batches, making the smallest singular value $\sigma_N(\rmW_N)$ close to $0$.
Consequently, we propose to replace the condition number as a regularization term by the \emph{negative entropy of the vector of singular values} as follows:
\begin{align*}
    & H_\sigma(\rmW_N):= \sum_{i=1}^N \softmax(\mathbf{\sigma}(\rmW_N))_i \cdot \log \softmax(\mathbf{\sigma}(\rmW_N))_i,
\end{align*}
where $\softmax(\cdot)_i$ is the $i^{th}$ output of the \emph{softmax} function. Since uniform distribution has the highest entropy, regularizing with $\kappa(\rmW_N)$ or $H_\sigma(\rmW_N)$ leads to a better coverage of $\sR^k$ by ensuring a nearly identical importance regardless of the direction.


\noindent
We obtain the following regularized optimization problem:
\begin{align}
    \widehat{\phi}, \widehat{\rmW} &= \argmin_{\phi \in \Phi, \rmW \in \sR^{T\times k}} \frac{1}{Tn_1}\sum_{t=1}^T\sum_{i=1}^{n_1} \ell(y_{t,i},\langle \tilde{\rvw}_t, \phi(\rvx_{t,i})\rangle) + \lambda_1 H_\sigma(\rmW),
    \label{eq:meta_train_reg_metric}
\end{align}
where $\tilde{\rvw} = \frac{\rvw}{\|\rvw\|}$ are the normalized prototypes.

In Table~\ref{tab:proto_perf}, we report the performance of \Proto\ without normalization, with normalization and with both normalization and regularization on the entropy. Finally, we can see that further enforcing a regularization on the singular values through the entropy does not help the training since \Proto\ naturally learns to minimize the singular values of the prototypes. In Table~\ref{tab:hparam_proto}, we show that reducing the \emph{strength} of the regularization with the entropy can help improve the performance.



\begin{table}[!ht]
    \begin{center}
    \caption{Performance of \Proto\ with and without our regularization on the entropy and/or normalization. All accuracy results (in \%) are averaged over 2400 test episodes and 4 different random seeds and are reported with $95\%$ confidence interval. Further enforcing regularization on the singular values can be detrimental to performance.}
    \label{tab:proto_perf}
    \begin{tabular}{@{}lllll@{}} \toprule
    \multirow{2}{*}{Dataset} & \multirow{2}{*}{Episodes} & without Norm., & with Norm.,  & with Norm.,  \\
     & & $\lambda_1 = 0$ & $\lambda_1 = 0$ & $\lambda_1 = 1$ \\
    \midrule
    \multirow{2}{*}{Omniglot} & 20-way 1-shot & $95.56 \pm 0.10\%$ & $\mathbf{95.89 \pm 0.10\%}$ & $91.90 \pm 0.14\%$ \\
     & 20-way 5-shot & $\mathbf{98.80 \pm 0.04\%}$ & $\mathbf{98.80 \pm 0.04\%}$ & $96.40 \pm 0.07\%$ \\
    \midrule
    \multirow{2}{*}{miniImageNet} & 5-way 1-shot & $49.53 \pm 0.41\%$ & $\mathbf{50.29 \pm 0.41\%}$ & $49.43 \pm 0.40\%$ \\
     & 5-way 5-shot & $65.10 \pm 0.35\%$ & $\mathbf{67.13 \pm 0.34\%}$ & $65.71 \pm 0.35\%$ \\
    \midrule
    \multirow{2}{*}{tieredImageNet} & 5-way 1-shot & $51.95 \pm 0.45\%$ & $\mathbf{54.05 \pm 0.45\%}$ & $53.54 \pm 0.44\%$ \\
     & 5-way 5-shot & $\mathbf{71.61 \pm 0.38\%}$ & $\mathbf{71.84 \pm 0.38\%}$ & $70.30 \pm 0.40\%$ \\
    \bottomrule
    \end{tabular}
    \end{center}
\end{table}

\begin{table}[ht]
    \caption{Ablative study on the strength of the regularization with normalized \Proto. All accuracy results (in \%) are averaged over 2400 test episodes and 4 random seeds and are reported with $95\%$ confidence interval.}
    \label{tab:hparam_proto}
    \begin{center}
    \resizebox{0.95\textwidth}{!}{
    \begin{tabular}{@{}lllllllll@{}} \toprule
    Dataset & Episodes & Original & $\lambda_1 = 0$ & $\lambda_1 = 1$ & $\lambda_1 = 0.1$ & $\lambda_1 = 0.01$ & $\lambda_1 = 0.001$ & $\lambda_1 = 0.0001$ \\
    \midrule
    \multirow{2}{*}{miniImageNet} & 5-way 1-shot & $49.53 \pm 0.41\%$ & $\mathbf{50.29 \pm 0.41\%}$ & $49.43 \pm 0.40\%$ & $\mathbf{50.19 \pm 0.41\%}$ & $\mathbf{50.44 \pm 0.42\%}$ & $\mathbf{50.46 \pm 0.42\%}$ & $\mathbf{50.45 \pm 0.42\%}$ \\
    & 5-way 5-shot & $65.10 \pm 0.35\%$ & $\mathbf{67.13 \pm 0.34\%}$ & $65.71 \pm 0.35\%$ & $66.69 \pm 0.36\%$ & $66.69 \pm 0.34\%$ & $\mathbf{67.2 \pm 0.35\%}$ & $\mathbf{67.12 \pm 0.35\%}$ \\
    \midrule
    \multirow{2}{*}{Omniglot} & 20-way 1-shot & $95.56 \pm 0.10\%$ & $\mathbf{95.89 \pm 0.10\%}$ & $91.90 \pm 0.14\%$ & $94.38 \pm 0.12\%$  & $95.60 \pm 0.10\%$ & $95.7 \pm 0.10\%$ & $95.77 \pm 0.10\%$ \\
     & 20-way 5-shot & $98.80 \pm 0.04\%$ & $98.80 \pm 0.04\%$ & $96.40 \pm 0.07\%$ & $97.93 \pm 0.05\%$ & $98.62 \pm 0.04\%$ & $98.76 \pm 0.04\%$ & $\mathbf{98.91 \pm 0.03\%}$ \\
    \bottomrule
    \end{tabular}
    }
    \end{center}
\end{table}

\subsection{Ablation studies}\label{ax:ablation}

In this Section, we present a study on the effect of each term in the proposed regularization for \Maml\ and MTL.
In Table~\ref{tab:maml_ablative}, we compare the performance of \Maml\ without regularization ($\lambda_1 = \lambda_2 = 0$), with a regularization on the condition number $\kappa(\rmW_N)$ ($\lambda_1 = 1$ and $\lambda_2 = 0$), on the norm of the linear predictors ($\lambda_1 = 0$ and $\lambda_2 = 1$), and with both regularization terms ($\lambda_1 = \lambda_2 = 1$) on Omniglot and miniImageNet. We can see that both regularization terms are important in the training and that using only a single term can be detrimental to the performance.
Table~\ref{tab:hparam_mtl} presents the effect of varying independently either parameter $\lambda_1$ or $\lambda_2$ in the regularization, the other being fixed to 1. From these results, we can see that performance is much more impacted by the condition number regularization (parameter $\lambda_1$) than by the normalization (parameter $\lambda_2$). Indeed, varying the regularization weight can lead from the lowest accuracy (74.64\%, for $\lambda_1=0$) to one of the highest accuracies (76.15\% for $\lambda_1=0.2$).

\begin{table}[!ht]
    \caption{Performance of MTL~\cite{pmlr-v139-wang21ad} when varying either $\lambda_1$ or $\lambda_2$, the other being fixed to 1, on the miniImageNet 5-way 5-shot benchmark. All accuracy results (in \%) are averaged over 2000 test episodes on a single random seed.}
    \label{tab:hparam_mtl}
    \begin{center}
    \begin{tabular}{@{}llllllllll@{}} \toprule
    $\lambda_1$ & 1 & 0.8 & 0.6 & 0.4 & 0.2 & 0.1 & 0.05 & 0.01 & 0 \\
    Accuracy & \multirow{2}{*}{75.84} & \multirow{2}{*}{75.85} & \multirow{2}{*}{76.02} & \multirow{2}{*}{76.11} & \multirow{2}{*}{76.15} & \multirow{2}{*}{75.99} & \multirow{2}{*}{75.65} & \multirow{2}{*}{75.08} & \multirow{2}{*}{74.64} \\
    ($\lambda_2 = 1$) & & & & & & & & & \\
    \midrule
    $\lambda_2$ & 1 & 0.8 & 0.6 & 0.4 & 0.2 & 0.1 & 0.05 & 0.01 & 0 \\
    Accuracy & \multirow{2}{*}{75.84} & \multirow{2}{*}{76.09} & \multirow{2}{*}{75.81} & \multirow{2}{*}{76.28} & \multirow{2}{*}{76.23} & \multirow{2}{*}{76.1} & \multirow{2}{*}{76.25} & \multirow{2}{*}{\textbf{76.42}} & \multirow{2}{*}{76.06} \\
    ($\lambda_1 = 1$) & & & & & & & & & \\
    \bottomrule
    \end{tabular}
    \end{center}
\end{table}

\begin{table}[!ht]
    \caption{
    Ablative study of the regularization parameter for \Maml, on Omniglot (\emph{left}) with 20-way 1-shot (\emph{top values}) and 20-way 5-shot (\emph{bottom values}) episodes, and miniImageNet (\emph{right}) with 5-way 1-shot (\emph{top values}) and 5-way 5-shot (\emph{bottom values}) episodes. All accuracy results (in \%) are averaged over 2400 test episodes and 4 different random seeds and are reported with $95\%$ confidence interval. We can see that in all cases, using both regularization terms is important.}
    \label{tab:maml_ablative}
    \begin{subtable}{0.45\textwidth}
    \centering
    \subcaption{Omniglot (20-way 1-shot / 5-shot)}
    \begin{tabular}{@{}lll@{}} \toprule
     & $\lambda_1 = 0$ & $\lambda_1 = 1$ \\
    \midrule
    \multirow{2}{*}{$\lambda_2 = 0$} & $91.72 \pm 0.29\%$ & $89.86 \pm 0.31 \%$ \\
    & $97.07 \pm 0.14\%$ & $72.47 \pm 0.17 \%$ \\
    \midrule
    \multirow{2}{*}{$\lambda_2 = 1$} & $92.80 \pm 0.26 \%$ & $\mathbf{95.67 \pm 0.20 \%}$ \\
    & $96.99 \pm 0.14 \%$ & $\mathbf{98.24 \pm 0.10\%}$ \\
    \bottomrule
    \end{tabular}
    \end{subtable}
    \hfill
    \begin{subtable}{0.45\textwidth}
    \centering
    \subcaption{miniImageNet (5-way 1-shot / 5-shot)}
    \begin{tabular}{@{}lll@{}} \toprule
     & $\lambda_1 = 0$ & $\lambda_1 = 1$ \\
    \midrule
    \multirow{2}{*}{$\lambda_2 = 0$} & $47.93 \pm 0.83\%$ & $47.76 \pm 0.84\%$ \\
    & $64.47 \pm 0.69\%$ & $64.44 \pm 0.68\%$ \\
    \midrule
    \multirow{2}{*}{$\lambda_2 = 1$} & $48.27 \pm 0.81\%$ & $\mathbf{49.16 \pm 0.85\%}$ \\
    & $64.16 \pm 0.72\%$ & $\mathbf{66.43 \pm 0.69\%}$ \\
    \bottomrule
    \end{tabular}
    \end{subtable}
\end{table}

\subsection{Out-of-Domain Analysis}\label{ax:cross-domain}
\begin{figure}
\begin{center}
    \includegraphics[width = .24\linewidth]{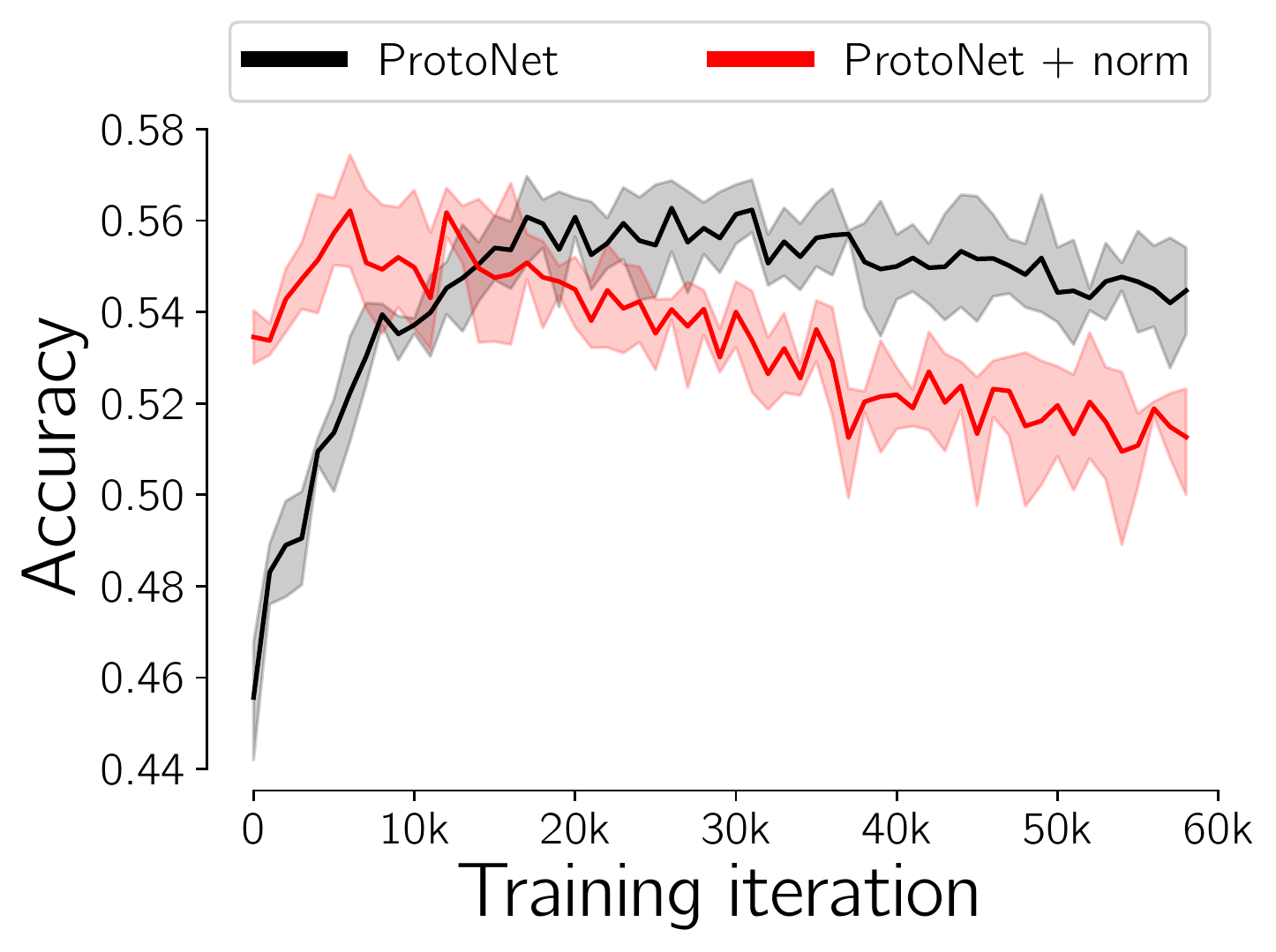}
    \includegraphics[width = .24\linewidth]{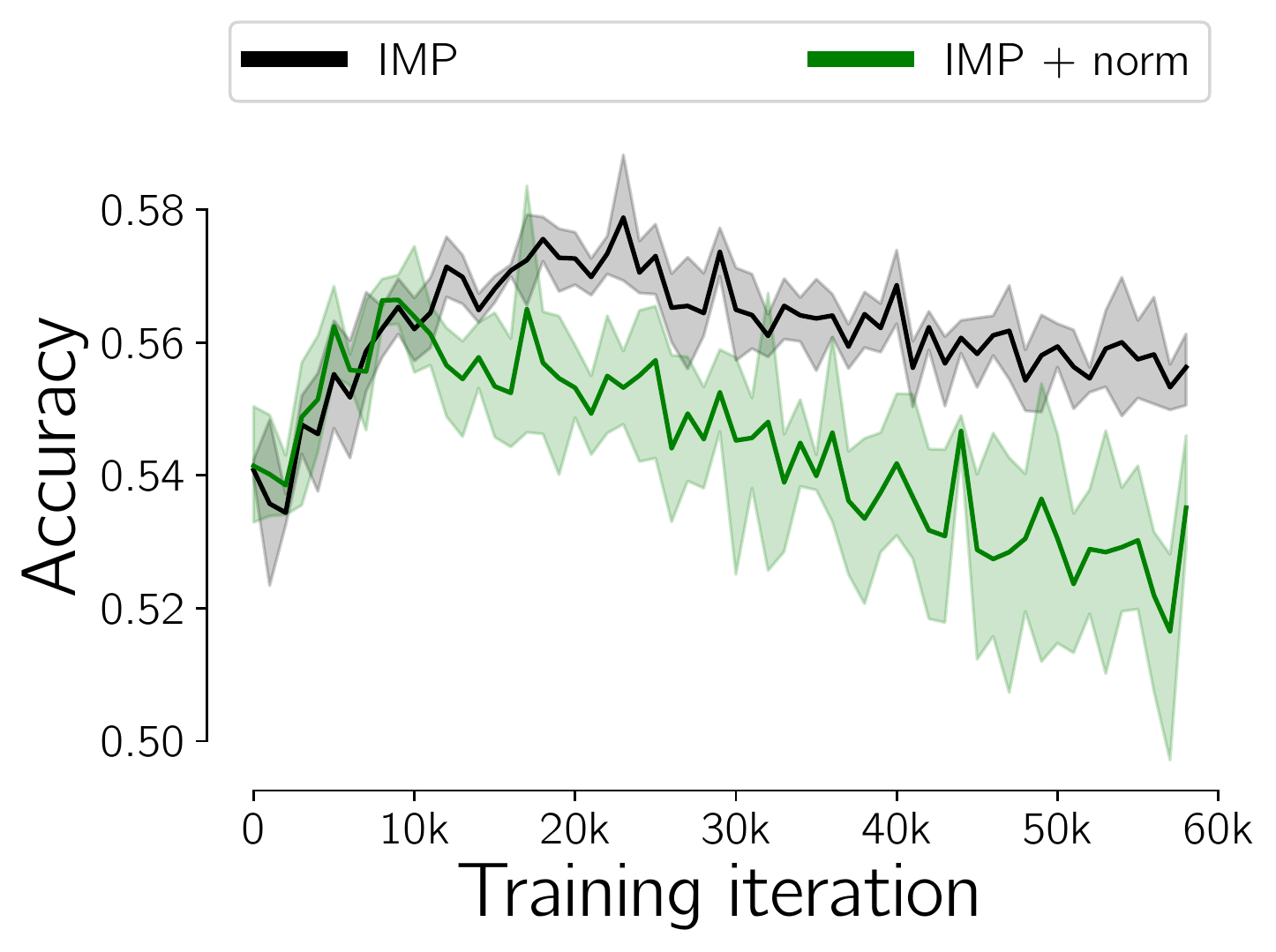}
    \includegraphics[width = .24\linewidth]{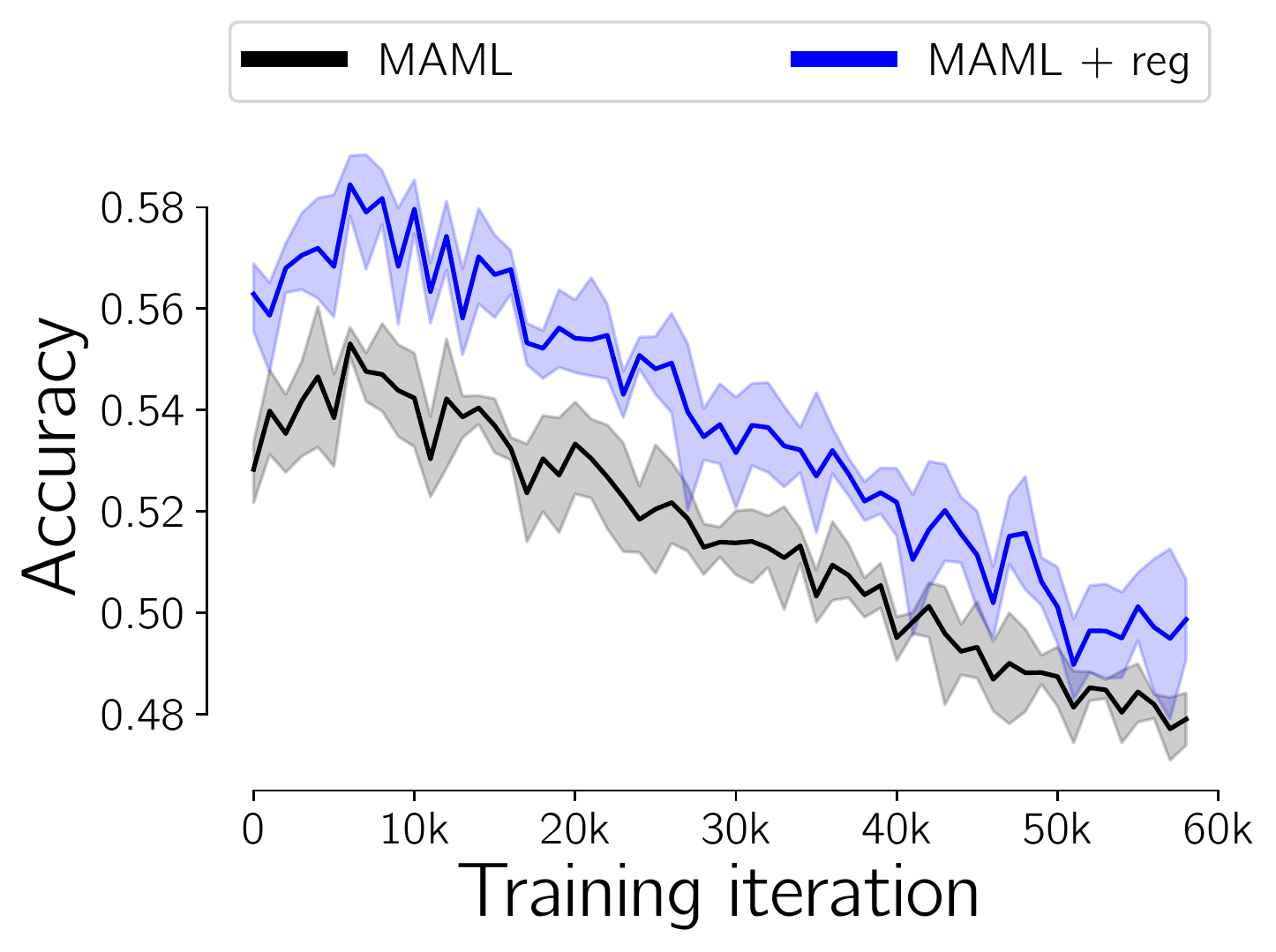}
    \includegraphics[width = .24\linewidth]{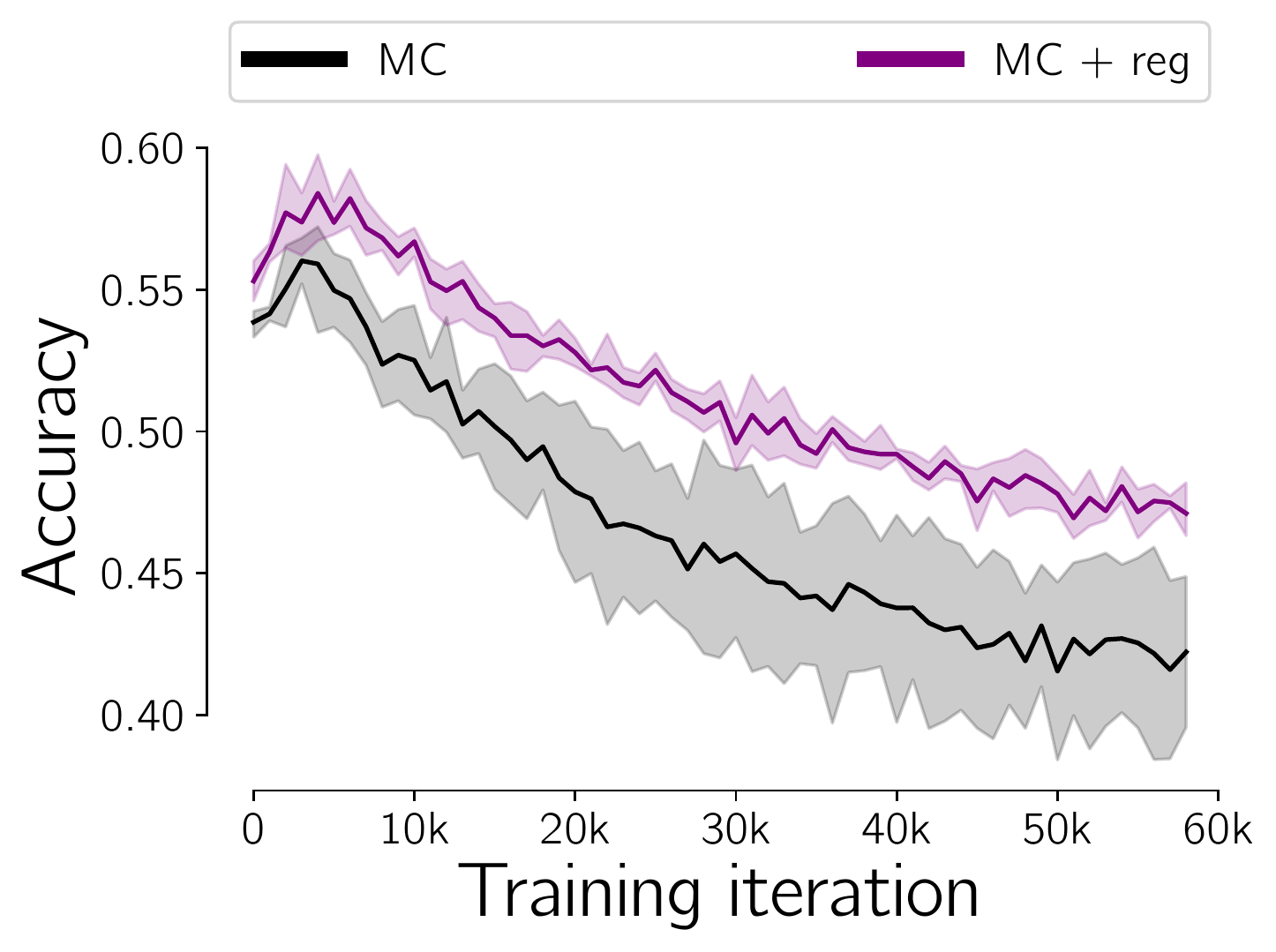}
    
    \includegraphics[width = .24\linewidth]{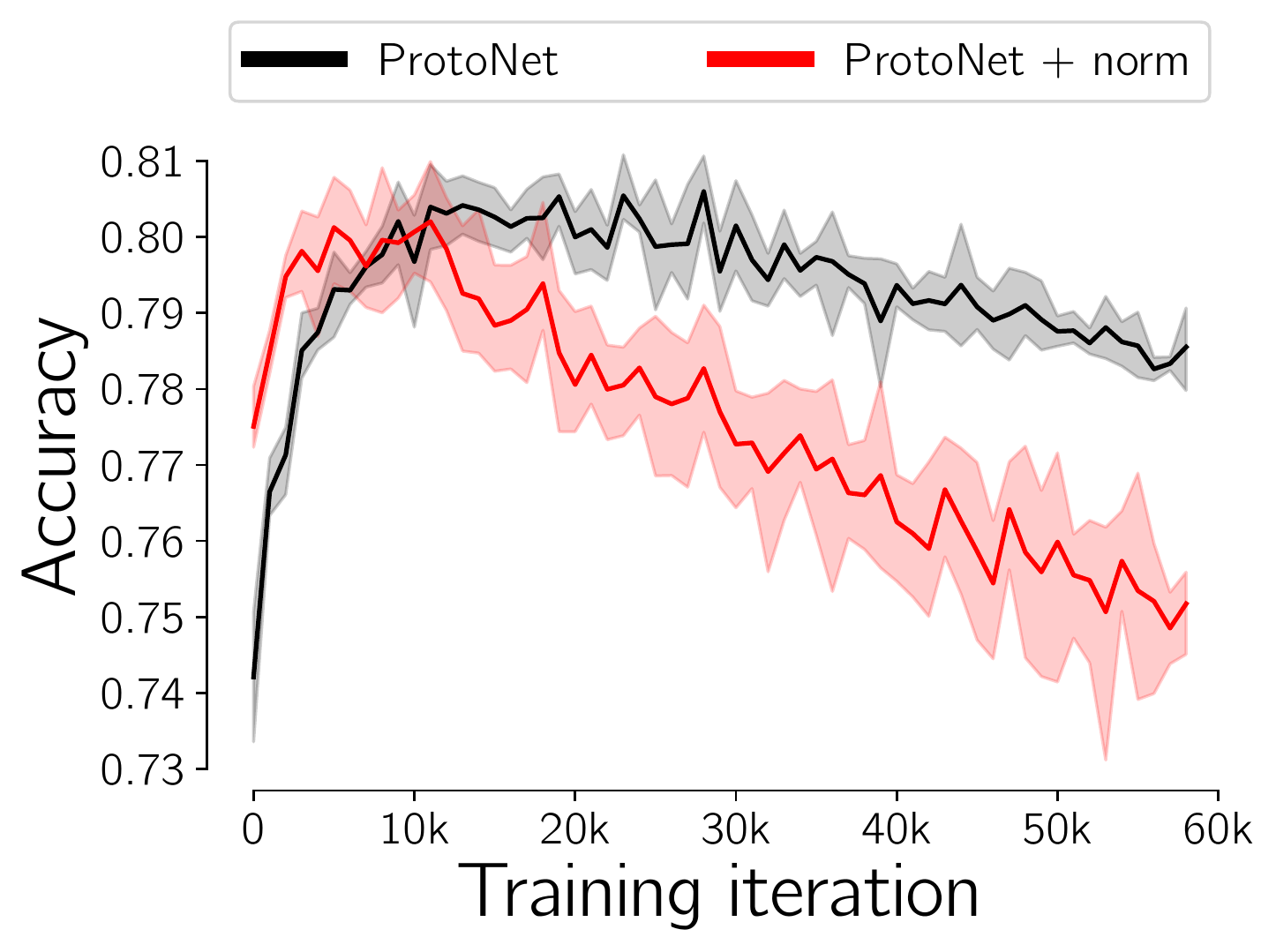}
    \includegraphics[width = .24\linewidth]{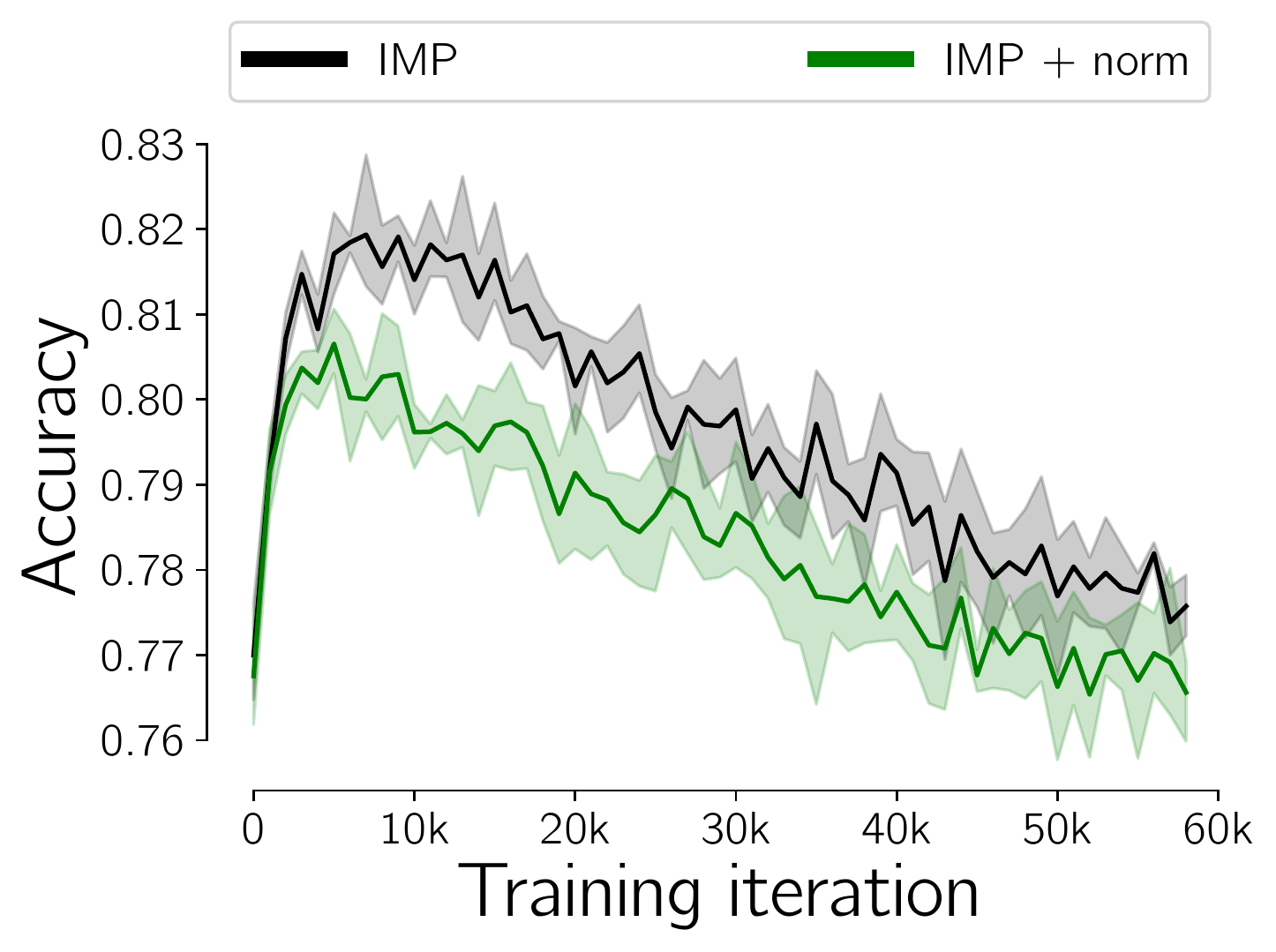}
    \includegraphics[width = .24\linewidth]{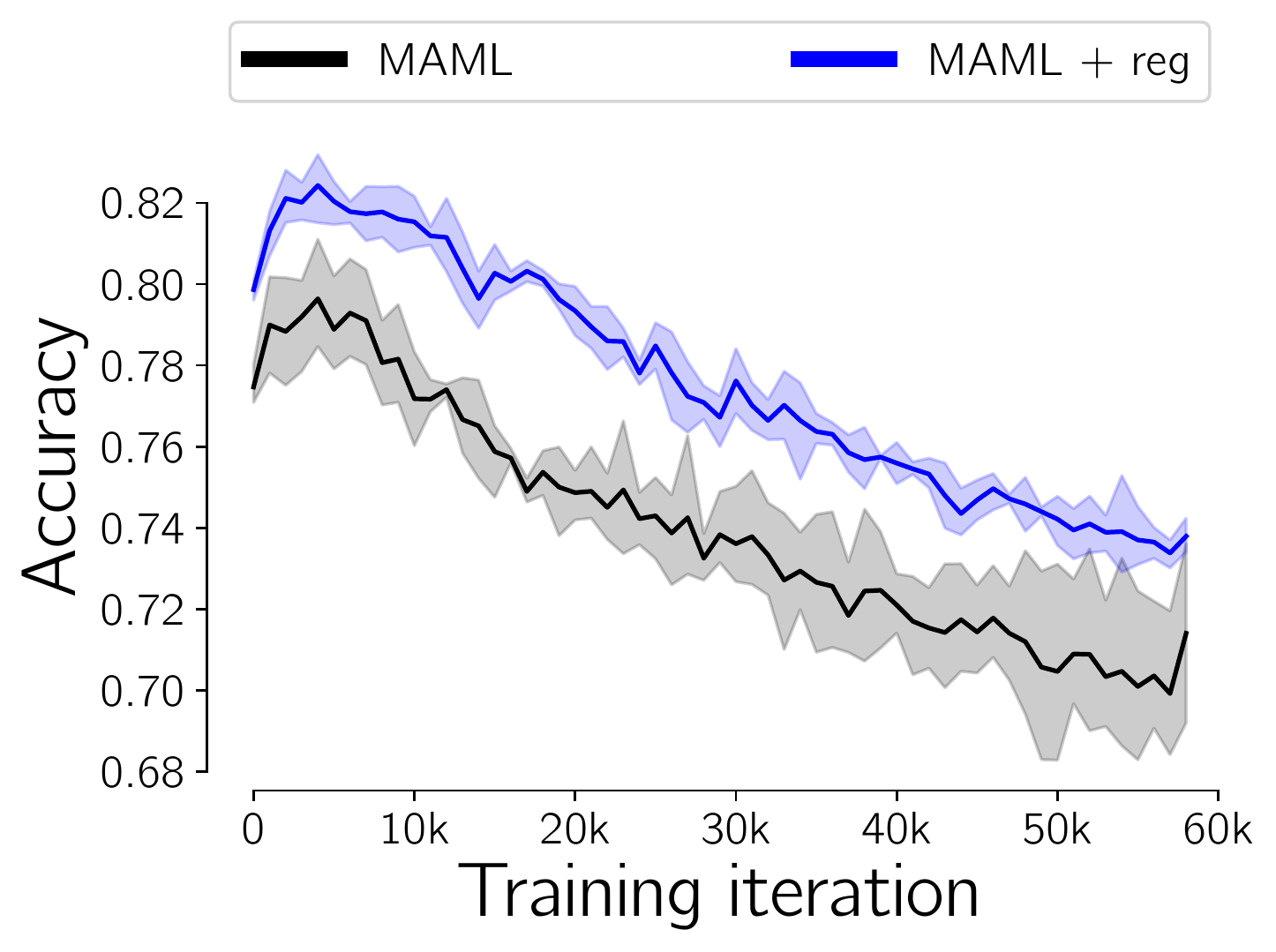}
    \includegraphics[width = .24\linewidth]{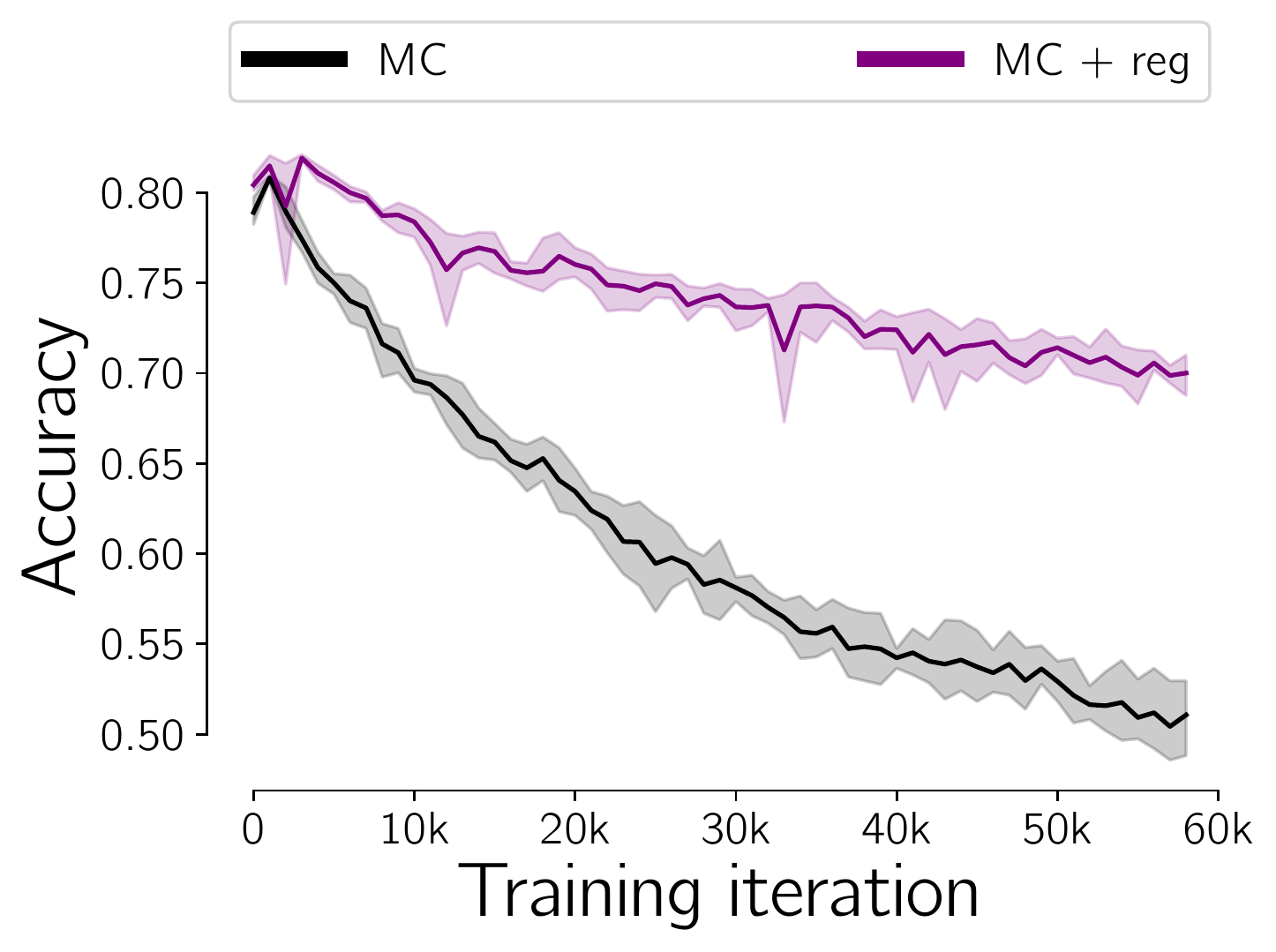}

    \captionof{figure}{
    Evolution of accuracy of 5-way 1-shot (\emph{top}, resp. 5-way 5-shot, \emph{bottom}) meta-test episodes from \emph{CropDisease} during meta-training on 5-way 1-shot (\emph{top}, resp. 5-way 5-shot, \emph{bottom}) episodes from \emph{miniImageNet},
    for \Proto, \IMP, \Maml, \MC\ (\emph{from left to right}) and their regularized or normalized counterparts (\emph{in red, green, blue and purple, respectively}).
    All results are averaged over 4 different random seeds. The shaded areas show $95\%$ confidence intervals.}
    \label{fig:acc_curves_ood}
    \end{center}
\end{figure}

In the theoretical MTR framework, one additional critical assumption made is that 
the task data marginal distributions 
are similar (see Appendix~\ref{ax:MTR} and assumption A2.5 for more information), which does not hold in a \emph{cross-domain} setting, where we evaluate a model on a 
dataset 
different from the training dataset. In this setting, we do not have the same guarantees that our regularization or normalization schemes will be as effective as in \emph{same-domain}.
To verify this, we measure out-of-domain performance on the \textbf{CropDiseases} dataset~\cite{mohantyUsingDeepLearning2016} adopted by \cite{guoBroaderStudyCrossDomain2020a}. Following their protocol, this dataset is used only for testing purposes. In this specific experiment, evaluated models are trained on \emph{miniImageNet}.

On the one hand, for metric-based methods, the improvement in the \emph{same-domain} setting does not translate to the \emph{cross-domain} setting. From Figure~\ref{fig:acc_curves_ood}, we can see that even though the low condition number in the beginning of training leads to improved early generalization capabilities of \Proto, this is not the case for \IMP. We attribute this discrepancy between \Proto\ and \IMP\ to a difference 
in cluster radius parameters of
\IMP\ and normalized \IMP, making the encoder less adapted to \emph{out-of-domain} features. 
On the other hand, we found that gradient-based models keep their accuracy gains when evaluated 
in cross-domain setting with improved generalization capabilities due to our regularization. This can be seen on Figure~\ref{fig:acc_curves_ood}, where we achieve an improvement of about 2 p.p. for both \Maml\ and \MC models on both 1-shot and 5-shot settings.

These results confirm that minimizing the norm and condition number of the linear predictors learned improves the generalization capabilities of meta-learning models. As opposed to metric-based methods which are already implicitly doing so, the addition of the regularization terms for gradient-based methods leads to a more significant improvement of performance in \emph{cross-domain}.

\end{document}